\documentclass[conference]{IEEEtran}
\IEEEoverridecommandlockouts
\usepackage{dblfloatfix}
\usepackage{cite}
\usepackage[hidelinks]{hyperref}
\usepackage{amsmath,amssymb,amsfonts}
\usepackage{algorithmic}
\usepackage{graphicx}
\usepackage{textcomp}
\usepackage{caption}
\usepackage{subcaption}
\usepackage[table]{xcolor}
\usepackage{url}
\usepackage{csvsimple}
\usepackage{booktabs}
\usepackage{adjustbox}
\usepackage{array}
\usepackage{makecell}
\usepackage{multirow}
\usepackage[T1]{fontenc}
\usepackage{float}

\usepackage{booktabs}
\usepackage{array}

\usepackage{bm}
\def\BibTeX{{\rm B\kern-.05em{\sc i\kern-.025em b}\kern-.08em
    T\kern-.1667em\lower.7ex\hbox{E}\kern-.125emX}}
\begin{document}

\title{Survey on Hand Gesture Recognition from Visual Input\\
\thanks{The research leading to these results received funding from the European Commission under Grant Agreement No. 101168042 (TRIFFID) and No. 101189557 (TORNADO).}
}

\author{\IEEEauthorblockN{Manousos Linardakis}
\IEEEauthorblockA{\textit{Department of Informatics and} \\ \textit{Telematics,} \\
\textit{Harokopio University of Athens}\\
Athens, Greece \\
manouslinard@gmail.com}
\and
\IEEEauthorblockN{Iraklis Varlamis}
\IEEEauthorblockA{\textit{Department of Informatics and} \\ \textit{Telematics,} \\
\textit{Harokopio University of Athens}\\
Athens, Greece \\
varlamis@hua.gr}
\and
\IEEEauthorblockN{Georgios Th. Papadopoulos}
\IEEEauthorblockA{\textit{Department of Informatics and} \\ \textit{Telematics,} \\
\textit{Harokopio University of Athens}\\
Athens, Greece \\
g.th.papadopoulos@hua.gr}
}

\maketitle

\begin{abstract}
Hand gesture recognition has become an important research area, driven by the growing demand for human-computer interaction in fields such as sign language recognition, virtual and augmented reality, and robotics. Despite the rapid growth of the field, there are few surveys that comprehensively cover recent research developments, available solutions, and benchmark datasets. This survey addresses this gap by examining the latest advancements in hand gesture and 3D hand pose recognition from various types of camera input data including RGB images, depth images, and videos from monocular or multiview cameras, examining the differing methodological requirements of each approach. Furthermore, an overview of widely used datasets is provided, detailing their main characteristics and application domains. Finally, open challenges such as achieving robust recognition in real-world environments, handling occlusions, ensuring generalization across diverse users, and addressing computational efficiency for real-time applications are highlighted to guide future research directions. By synthesizing the objectives, methodologies, and applications of recent studies, this survey offers valuable insights into current trends, challenges, and opportunities for future research in human hand gesture recognition.
\end{abstract}

\begin{IEEEkeywords}
Gesture classification, Gesture estimation, Hand gesture recognition, Sign language recognition.
\end{IEEEkeywords}

\maketitle

\section{Introduction}
\label{sec:introduction}
\begin{figure*}[htbp]
    \centering
    \begin{subfigure}[b]{0.32\textwidth}
            \centering
            \includegraphics[width=\textwidth]{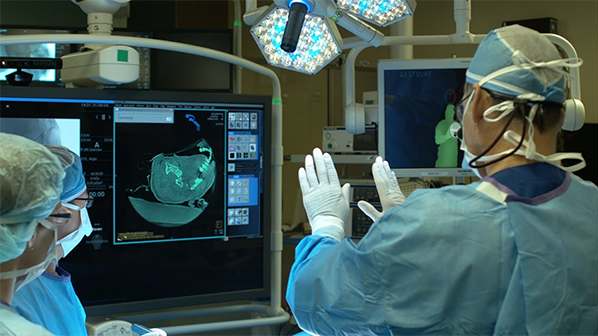}
            \caption{Medical Assistance.}
    \end{subfigure}
    \begin{subfigure}[b]{0.32\textwidth}
            \centering
            \includegraphics[width=\textwidth]{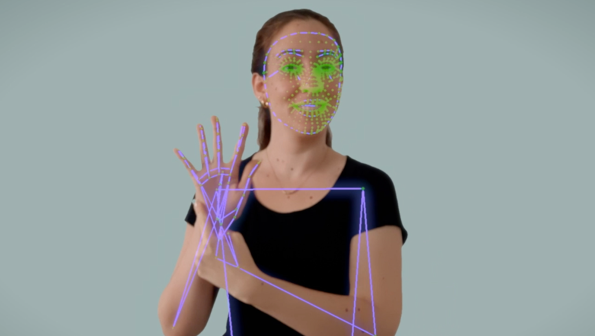}
            \caption{Sign Language Interpretation.}
    \end{subfigure}
    \begin{subfigure}[b]{0.32\textwidth}
            \centering
            \includegraphics[width=\textwidth]{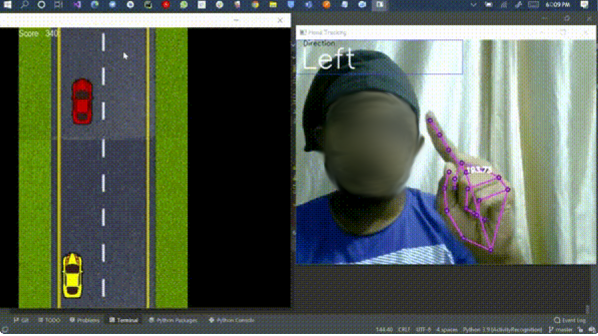}
            \caption{Gaming and Virtual Environments.}
    \end{subfigure}
    \caption{Indicative applications of hand gesture recognition.}
    \label{fig:apps}
\end{figure*}

Hand gesture recognition has captivated researchers' interest for decades, as it offers a natural and intuitive form of human-computer interaction. The ability to accurately detect and interpret hand gestures has numerous applications, from enhancing communication for the hearing impaired through sign language recognition \cite{9622242, sign_language_recognition_survey, papad_hgr} to enabling more efficient human-robot interaction in industrial and medical settings \cite{8545095, BAMANI2024108443} (see Fig. \ref{fig:apps}). Human hands play a versatile role in non-verbal communication, frequently conveying messages that can replace spoken words when interactions need to be brief and unambiguous. For example, gestures such as a thumbs-up for approval or a wave to beckon someone closer serve as universally understood signals that transcend language barriers and are often more practical than spoken language in certain contexts \cite{GOLDINMEADOW1999419}. Additionally, hand gestures become especially valuable when communicating across physical distances where speech may not be audible, and enhance gaming by providing an intuitive, immersive control method that replaces traditional input devices.

\begin{figure*}[!htb]
\centering
\includegraphics[width=\textwidth]{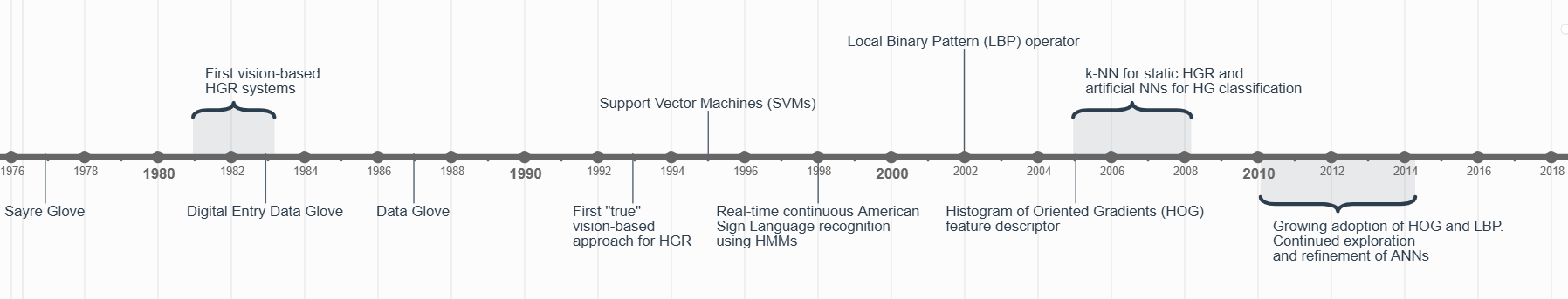}
\caption{\textcolor{black}{Timeline of the main advances in HGR for the last 40 years.}}
\label{fig:timeline}
\end{figure*}

\textcolor{black}{
The field of hand gesture recognition (HGR) has undergone a significant evolution over the years, transitioning from early reliance on sensor-based data gloves, exemplified by the Sayre Glove in 1977 and the Digital Entry Data Glove in 1983, to increasingly sophisticated vision-based approaches, with the first such systems appearing in the early 1980s  and a "true" vision-based approach reported by in 1993 \cite{premaratne2014historical}. Key advancements included the development and application of various feature extraction techniques such as image moments, convex hull analysis, Gabor filters, Histogram of Oriented Gradients \cite{dalal2005histograms}, and Local Binary Patterns \cite{ojala2002multiresolution}. Furthermore, the application of classical machine learning algorithms like Hidden Markov Models in the late 1990s, k-Nearest Neighbors with foundational early applications in HGR in the early 2000s, Naive Bayes and Support Vector Machines applied to HGR in the early 2000s, played a crucial role in enabling gesture classification. This rich history of foundational work provided the essential knowledge and techniques that paved the way for the more recent and rapid progress observed in the last five years. In this later period the transformative impact of deep learning methodologies, fueled by increased computational power and the availability of larger datasets, led to significant breakthroughs in the accuracy and robustness of hand gesture recognition systems during that time. Fig. \ref{fig:timeline} provides a timeline of the evolution of HGR over the years.   
}

As gesture recognition technology continues to advance, it holds the potential to transform both human-computer \cite{9903078, TERRERAN2023104523} and human-to-human \cite{article, EEI6059} interactions by enabling contactless control and enhancing accessibility for users with disabilities. Additionally, 3D hand reconstruction \cite{10030397, 9648765} is an important aspect of human gesture recognition, offering valuable digital insights into hand movements and supporting applications in fields like 3D animation. The growing importance of this field underscores the need for accurate, real-time recognition systems capable of both interpreting and reconstructing a wide range of gestures across diverse environments.

The type of data used for hand gesture recognition influences the accuracy, efficiency, and suitability of hand recognition systems across various application domains. Such data are available in various modalities, collected by infrared sensors \cite{thermal_camera, infrared_camera}, motion capture systems, cameras, and wearable devices \cite{guo2025rapid, wearable}. Methods relying on infrared sensors or motion capture technology for hand gesture classification and estimation typically involve complex setups that can limit their applicability in real-world scenarios. On the other side, RGB cameras are commonly found in smartphones, tablets, and laptops, making it easier for researchers to test methodologies and to collect large datasets without using specialized equipment. This democratization of technology enables researchers to develop and deploy gesture recognition systems that can be utilized in everyday settings. Based on the above, we focus on works that use RGB images, depth images, and video data collected by cameras, which are broadly available and accessible to researchers and are widely used in common user applications.

\textcolor{black}{
Existing surveys in the field, their scope, methodology, contribution, and novelties are summarized in Table \ref{tab:surveys}.
}

\begin{table*}[htbp]
\centering
\caption{\textcolor{black}{An overview of recent surveys in the field and their features.}}
\resizebox{\textwidth}{!}{%
\begin{tabular}{p{2.28cm}p{2cm}p{3cm}p{2.6cm}p{2.28cm}p{3cm}}
\toprule
\textbf{Article} & \textbf{Scope} & \textbf{Methodology} & \textbf{Contribution} & \textbf{Novelty} & \textcolor{black}{\textbf{Limitations}} \\
\midrule

A systematic review on hand gesture recognition techniques, challenges and applications (2019) \cite{yasen2019systematic} & Research on HGR. & Surveys 244 papers (49 excluded). Compares techniques, applications, and challenges. & Discusses gesture acquisition methods (image-based, glove-based, motion sensing). & Focuses on the latest research comparisons across multiple dimensions. & \textcolor{black}{Limited by its 2016-2018 timeframe, reliance solely on the IEEE Xplore database. Lacks current and visual-input-focused scope.} \\ \addlinespace

Hand Gesture Recognition Based on Computer Vision (2020) \cite{oudah2020hand} & Computer vision techniques for HGR. & Reviews literature on HG techniques. Tabulates performance, classification, datasets, and camera types. & Lists merits/limitations of methods. Comparative review of studies using computer vision. & First review to analyze camera types, distance limitations, and recognition rates. & \textcolor{black}{Published in 2020, it lacks latest advancements, a systematic methodology for paper selection, and focus on hand pose estimation.} \\ \addlinespace

Methods, databases and recent advancement of vision-based HGR for HCI systems (2021) \cite{sarma2021methods} & Vision-based methods for human-computer interaction, focusing on static/dynamic gestures and dataset characteristics. & Critical analysis of 85+ papers (2016-2021). Compares preprocessing techniques, feature descriptors, and classifiers across 10 benchmark datasets. & Detailed comparison of recognition accuracy across illumination conditions. Introduces HCI-specific evaluation metrics beyond pure accuracy. & First survey to explicitly link gesture recognition performance to HCI usability requirements and ergonomic factors. & \textcolor{black}{Primarily HCI-focused and published in 2021, it misses recent HGR advancements, datasets, and the latest systematic method classification.} \\ \addlinespace

A Structured and Methodological Review on Vision-Based Hand Gesture Recognition System (2022) \cite{al2022structured} & Vision-driven hand gesture recognition across various camera orientations. & Literature survey of 108 articles summarizes and compares similar methodologies. & Identifies limitations in image acquisition, segmentation, tracking, feature extraction, and classification. & Aims to identify improving areas and underdeveloped challenges. & \textcolor{black}{Covers up to 2022, missing recent works and the latest systematic dataset/methodology synthesis.} \\ \addlinespace

A Systematic Review of Hand Gesture Recognition (2024) \cite{hashi2024systematic} & Comprehensive analysis of HGR advancements from 2018-2024, including deep learning and sensor fusion approaches. & Systematic review using PRISMA framework. Analyzes 127 papers. Classifies by sensing modality and application domain. & Identifies emerging trends and persistent challenges (cross-cultural gestures, real-time deployment). & First review to systematically benchmark post-2020 deep learning architectures and their deployment constraints in HGR. & \textcolor{black}{Includes sensor-based methods, not solely visual input; less comprehensive on specific visual techniques classification (e.g., box-filter vs. skeleton).} \\ \addlinespace

A Decade of Progress in Human Motion Recognition (2024) \cite{noh2024decade} & HMR covering vision sensor-based (VSM) and wearable sensor-based methods (WSM). & Categorization of research into VSM and WSM, assessed based on sensors, classification algorithms, datasets, gesture types, target body parts, and performance. & Provides a comprehensive and systematic understanding of the latest developments in HMR techniques from 2010 to 2020. & Categorizes HMR Assesses HMR from multiple viewpoints to provide a comprehensive overview of research trends and technological advancements. & \textcolor{black}{Broader HMR scope (not just HGR), 2010-2020 timeframe, and inclusion of wearable sensors limits its HGR visual-input focus.}\\ \addlinespace

A Methodological and Structural Review of HGR (2024) \cite{shin2024methodological} & HGR techniques and data modalities. & Reviews HGR techniques and modalities (2014–2024). Assesses efficacy through recognition accuracy. & Comprehensive review of sensor tech. Identifies gap in continuous gesture recognition. & First to comprehensively review HGR data modalities within this timeframe. & \textcolor{black}{Covers diverse data modalities, not exclusively visual input, potentially weakening focus on visual-input HGR.} \\ \addlinespace

Survey on vision-based dynamic hand gesture recognition (2024) \cite{tripathi2024survey} & Vision-based dynamic HGR. & Summarizes techniques, merits/demerits, datasets, accuracy, and algorithms. Scrutinizes traditional vs. deep learning. & Examines performance of traditional and deep learning methods. & First to systematically compare traditional and deep learning for dynamic gestures. & \textcolor{black}{Focuses on dynamic HGR; less dedicated coverage of static gestures or 3D hand pose estimation.} \\ \addlinespace

Survey on Hand Gesture Recognition from Visual Input. (2025) [Current survey] & Recent advancements in hand gesture and 3D hand pose recognition from RGB, depth, and video data. & Systematic review of 137 papers from top computer vision venues using topic modeling. Classifies methods by task, input, capture, and ML techniques. & Comprehensive overview of recent research, solutions, and datasets. Organized presentation of research approaches. Review of current research trends. & Addresses a gap in comprehensive coverage of recent developments. More focused review concentrating on visual input. \\ \addlinespace

\bottomrule
\end{tabular}
} 
\label{tab:surveys}
\end{table*}

Compared to other recent surveys in the field, this survey offers a more comprehensive yet focused review of research in the field of Hand Gesture Recognition (HGR) using visual input. Unlike \cite{shin2024methodological} that explores various modalities beyond visual input (e.g., EEG, EMG, and audio) with a primary emphasis on Sign Language Recognition (SLR) and limited discussion on hand gesture estimation, our work specifically targets the visual input-based approaches. We categorize HGR into distinct tasks such as classification and estimation, as well as methodological distinctions like single-camera versus multi-camera setups, offering deeper insights into both general HGR and specialized topics like body reconstruction with an emphasis on HGR. Compared to \cite{noh2024decade} that categorizes Human Motion Recognition (HMR) methods broadly into vision sensor-based (VS) and wearable sensor-based (WS) ones, and further divide HGR into markerless and marker-based approaches, our survey focuses exclusively on visual input-based methods, ensuring a more detailed exploration of modern challenges, datasets, and trends. Moreover, we include research from 2018 to the present, incorporating all the latest advancements, challenges, and trends. This focus and up-to-date analysis allow our survey to present a more relevant and actionable synthesis of current and future directions in HGR research.

\subsection{Contributions of the Paper}
This survey provides a comprehensive overview of recent advances in hand gesture recognition, aiming to cover multiple aspects, from algorithms and input types to applications and datasets. The main contributions of this paper can be summarized in the following:

\begin{itemize}
\item An organized presentation of research approaches: We organize the surveyed studies based on input data types (e.g., RGB images, depth images, video), hand capture techniques, and recognition methods used, providing a clear framework for understanding current research directions.
\item A survey of applications and datasets: We analyze the applications and datasets employed in recent studies, discussing their main features, limitations, and suitability for different applications.
\item A review of current research trends: We examine the latest trends in human gesture recognition and estimation, highlighting emerging approaches and key focus areas within the field.
\end{itemize}

\subsection{Organization of the Paper}
Section \ref{sec:basics} presents the fundamental concepts and defines the terminology used in the field, and at the same time illustrates the various types of human gestures (e.g., static vs. dynamic, hand, body, and facial gestures). Section \ref{sec:review_method} outlines the review methodology used to select and categorize the surveyed papers. Section \ref{sec:methods} provides an overview of the hand gesture recognition methods and their classifications, whereas Section \ref{sec:algorithms} surveys the most popular algorithms employed in HGR and explains how they formulate and solve the task. Section \ref{sec:datasets} discusses datasets used across different applications in hand gesture recognition. Section \ref{sec:challenges} identifies trends and challenges highlighted in the literature. Finally, Section \ref{sec:conclusions} summarizes key findings and suggests future research directions for advancements in the field.


\section{Background and Fundamentals}\label{sec:basics}

\subsection{Definition of Gesture and Gesture Recognition} \label{subsec:def_gesture}
A \textit{gesture} can be defined as a structured, intentional movement or posture of the human body, often involving the hands, arms, or face, used to convey information and interact with devices, or control systems. Formally, let $G$ represent a gesture, which can be expressed as a temporal sequence of poses:

\begin{equation*}
G = \{P_1, P_2, \ldots, P_n\}, \quad P_i \in \mathbb{R}^d    
\end{equation*}

\noindent where $P_i$ denotes the $i$-th pose, represented in a $d$-dimensional feature space capturing spatial or spatio-temporal characteristics such as joint coordinates, velocity, or orientation.

In the context of this study, \textit{Gesture Recognition} is subdivided into two categories; \textit{Gesture Classification} and \textit{Gesture Estimation}.
\subsubsection{Gesture Classification} \label{subsubsec:gest_classif} Gesture classification refers to the computational process of detecting and interpreting gestures from input data. Given a time series of inputs $X = \{x_1, x_2, \ldots, x_t\}$ derived from sensors or imaging devices (e.g., RGB cameras, depth sensors), gesture recognition involves mapping $X$ to a predefined set of gesture labels $\mathcal{C} = \{C_1, C_2, \ldots, C_k\}$:

\begin{equation*}
f : X \rightarrow \mathcal{C},
\end{equation*}
\noindent where $f$ is the recognition function, trained to minimize a loss function $\mathcal{L}(y, f(X))$, $y$ being the ground truth label.

\subsubsection{Gesture Estimation} \label{subsubsec:gest_estim} Gesture estimation refers to the computational process of digitally representing a gesture's structure and configuration. Unlike gesture classification, which maps gestures to predefined labels, gesture estimation aims to extract key spatial or temporal features of the gesture, such as joint positions, angles, or trajectories. Given a time series of inputs $X = \{x_1, x_2, \ldots, x_t\}$ from sensors or imaging devices (e.g., RGB cameras, depth sensors), gesture estimation involves mapping $X$ to a set of gesture pose parameters $P = \{p_1, p_2, \ldots, p_n\}$ that define the gesture's digital representation:

\begin{equation*} g : X \rightarrow P,
\end{equation*}

\noindent where $g$ is the estimation function, trained to minimize a reconstruction loss $\mathcal{L}(\hat{P}, g(X))$, $\hat{P}$ representing the ground truth gesture pose parameters (e.g., 3D joint coordinates or gesture skeleton structure).

\subsection{Static and Dynamic Gestures}
Gestures can be classified into \textit{static} and \textit{dynamic} ones. A \textit{static gesture} is defined as a single pose \( P \) that remains unchanged over time:

\begin{equation*}
G_{\text{static}} = P, \quad P \in \mathbb{R}^d.
\end{equation*}

In contrast, a \textit{dynamic gesture} involves a sequence of poses over time, modeled as a trajectory:

\begin{equation*}
G_{\text{dynamic}} = \{P_t \mid t = 1, 2, \ldots, T\},
\end{equation*}

\noindent where $T$ represents the duration of the hand gesture. Dynamic gestures require temporal modeling to capture dependencies across the movement sequence.

\subsection{Multimodal Gestures}
Multimodal gestures refer to gestures that involve multiple body parts or the integration of multiple sensory modalities to enhance the richness and clarity of communication. These gestures combine various sources of information, such as hand movements, facial expressions, or even audio cues, to form a more comprehensive understanding of human intent.

Formally, a multimodal gesture $G_m$ can be expressed as the combination of modalities:
\begin{equation*}
G_m = \{M_1, M_2, \ldots, M_k\},
\end{equation*}
\noindent where $M_i$ represents the $i$-th modality, such as hand motion, facial expressions, or audio signals. Each modality $M_i$ can be represented in its own feature space $\mathbb{R}^{d_i}$, and the fusion of modalities $\phi(G_m)$ combines these features:

\begin{equation*}
\phi(G_m) = \phi_1(M_1) \oplus \phi_2(M_2) \oplus \ldots \oplus \phi_k(M_k),
\end{equation*}
\noindent where $\oplus$ denotes the fusion operation, which may involve concatenation, attention mechanisms, or other techniques to integrate information.

Examples of multimodal gestures include:
\begin{itemize}
    \item \textbf{Hand and Facial Expressions:} Gestures such as a thumbs-up combined with a smile to reinforce positive feedback.
    \item \textbf{Speech and Gestures:} Pointing while providing verbal instructions to emphasize spatial references.
    \item \textbf{Cross-Sensory Integration:} Clapping, snapping or vocal tone emphasis 
    to accompany hand gestures, providing auditory reinforcement.
\end{itemize}

\begin{figure}[htbp]
\centering
\includegraphics[width=\columnwidth]{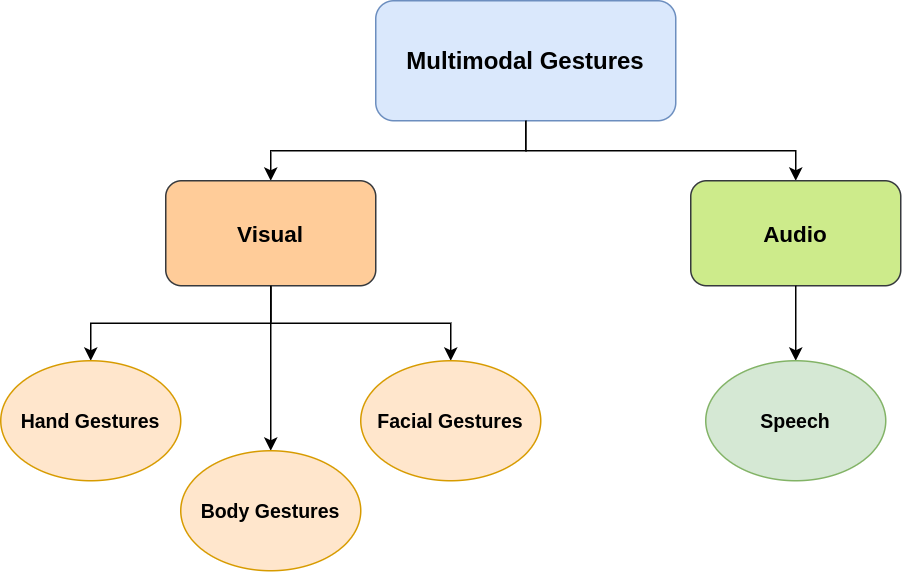}
\caption{Hierarchy of Multimodal Gestures.}
\label{fig:multimodal_cat}
\end{figure}

The integration of multimodal data often requires specialized frameworks to capture dependencies across modalities. These frameworks aim to improve recognition accuracy by leveraging complementary information and addressing ambiguities that may arise in unimodal systems.
Multimodal gestures are particularly relevant in applications such as virtual reality, where body movements and vocal commands are often combined for immersive interactions, or in sign language translation, where facial expressions are critical for conveying grammatical nuances. Fig. \ref{fig:multimodal_cat} explains what a multimodal gesture may comprise using a hierarchical structure, and categorizing the visual and audio modalities into distinct gesture types.

\subsection{Hand Gesture Recognition (HGR)} \label{subsec:hand_gest_reco}
Among the various types of gesture recognition, we focus on \textbf{hand gesture recognition} because it offers a unique combination of advantages. Hand gestures are highly expressive, enabling a wide range of commands, emotions, and intentions to be conveyed, making them versatile for applications such as human-computer interaction, virtual reality, and sign language interpretation. They also operate within a compact interaction zone, which is particularly practical for confined environments like vehicles, mobile devices, or personal workspaces. Additionally, hand gestures provide high precision and fine-grained control, essential for tasks like robotic manipulation, gaming, or medical surgery assistance. Many existing devices are already equipped with hardware optimized for capturing hand movements, reducing the need for specialized infrastructure. Furthermore, hand gestures are culturally and contextually universal, offering a natural and intuitive form of communication compared to other gesture types that may vary in interpretation. Lastly, the strong research foundation in hand recognition, supported by well-defined datasets and algorithms, allows for rapid development and deployment of systems, making it an ideal focus area.

\subsection{Unsupervised and Self-Supervised Learning in HGR}
\textcolor{black}{
Beyond traditional supervised approaches that rely on extensively labeled datasets, unsupervised and self-supervised learning paradigms are used in HGR, especially for scenarios with limited annotated data \cite{caron2021emergingpropertiesselfsupervisedvision}. These methods aim to learn meaningful representations or to discover patterns from visual input without manual labeling, thereby reducing annotation costs and enabling models to leverage vast amounts of readily available unlabeled hand gesture data.}

\textcolor{black}{Unsupervised learning in HGR focuses on identifying inherent structures within unlabeled gesture data. This can manifest as clustering visually similar static hand poses or dynamic gesture sequences, reducing the dimensionality of high-dimensional hand feature spaces (e.g., raw joint coordinates or image features) for more efficient processing, or detecting anomalous or out-of-distribution gestures. For instance, given a set of unlabeled gesture feature vectors $X = \{x_1, x_2, \dots, x_N\}$, clustering algorithms, such as K-Means, can partition the data into $K$ distinct groups, by minimizing an objective function like the sum of squared errors within each cluster:}

\begin{equation*} \label{eq:kmeans_hgr}
L_{\text{cluster}} = \sum_{k=1}^{K} \sum_{x_i \in C_k} ||x_i - \mu_k||^2
\end{equation*}
\textcolor{black}{where $C_k$ represents the $k$-th cluster and $\mu_k$ its centroid. Such techniques are valuable for initial data exploration, discovering novel gesture categories, or pre-training representations when labeled data is scarce or unavailable.}

\textcolor{black}{Self-supervised learning (SSL) is a specialized form of unsupervised learning where supervisory signals are automatically generated from the input data itself by defining "pretext" tasks. In HGR, SSL can learn powerful feature representations from large collections of unlabeled hand images or videos. Common pretext tasks include predicting the temporal order of shuffled video segments of a gesture, reconstructing masked-out patches of a hand image, predicting geometric transformations (e.g., rotation, scaling) applied to a hand image, or employing contrastive learning. In contrastive learning, the model is trained to maximize agreement between different augmented views of the same gesture instance (positive pairs), while minimizing agreement with views from different instances (negative pairs). A widely used objective for this is the InfoNCE loss \cite{oord2019representationlearningcontrastivepredictive}, which for an anchor sample $x_a$, its positive counterpart $x_p$, and a set of $M$ negative samples $\{x_n\}_{n=1}^M$, is often formulated as:}

\begin{equation*} \label{eq:infonce_hgr}
L = - \log \frac{e^{s_{a,p}/\tau}}{e^{s_{a,p}/\tau} + \sum_{n=1}^{M} e^{s_{a,n}/\tau}}
\end{equation*}
\textcolor{black}{
where $s_{i,j} = \text{sim}(f(x_i), f(x_j))$, $f(\cdot)$ is an encoder that maps inputs to latent feature vectors, $\text{sim}(\cdot, \cdot)$ denotes a similarity function (e.g., cosine similarity), and $\tau$ is a temperature scaling parameter. The objective encourages the model to assign higher similarity scores to positive pairs relative to negative ones. Representations learned via this mechanism can be effectively transferred and fine-tuned on downstream HGR tasks (e.g., gesture classification or estimation), often yielding notable performance improvements, especially in data-scarce scenarios.
}

\subsection{Key Takeaways of this section}
\textcolor{black}{
This section laid the groundwork by defining fundamental concepts and terminologies essential for understanding Hand Gesture Recognition. The main concepts covered include:
\begin{itemize}
    \item \textbf{Gesture Definition:} A gesture is a structured, intentional movement or posture, formally represented as a temporal sequence of poses.
    \item \textbf{Gesture Recognition Tasks:} Subdivided into \textit{Gesture Classification} (mapping input to predefined labels) and \textit{Gesture Estimation} (digitally representing a gesture's structure, e.g., pose parameters).
    \item \textbf{Gesture Types:} Differentiated into \textit{Static Gestures} (single, unchanging pose) and \textit{Dynamic Gestures} (sequence of poses over time).
    \item \textbf{Multimodal Gestures:} Involve multiple body parts or sensory modalities (e.g., hand movements with facial expressions or speech) to enhance communication.
    \item \textbf{HGR Focus:} This survey specifically targets Hand Gesture Recognition due to its expressiveness, compact interaction zone, precision, and strong research foundation.
    \item \textbf{Learning Paradigms:} Beyond supervised learning, \textit{Unsupervised Learning} (e.g., clustering) and \textit{Self-Supervised Learning (SSL)} (e.g., contrastive learning with InfoNCE loss) are also used in HGR to leverage unlabeled data.
\end{itemize}
}

\section{Review Methodology} \label{sec:review_method}

\subsection{Article Selection}

The selection of articles for this study targets all the recent papers in the computer vision field that specifically focus on hand gesture recognition from images and videos. To ensure a comprehensive review, we began by identifying the top-ranked venues in computer vision as listed under the "Top Publications - Computer Vision \& Pattern Recognition" section on Google Scholar\footnote{\url{https://scholar.google.com/citations?view_op=top_venues&hl=en&vq=eng_computervisionpatternrecognition}}. Using a crawler developed for this purpose\footnote{\href{https://github.com/manouslinard/gscholar\_scraper}{https://github.com/manouslinard/gscholar\_scraper}}, we retrieved 2,307 articles from these top venues. 
Subsequently, we conducted a systematic screening process in which the titles and abstracts of all articles were carefully reviewed based on predefined inclusion criteria detailed in Section \ref{subsec:inc_exc_crit}, which prioritized recent studies focusing on hand gesture recognition techniques, systems, or applications. This rigorous filtering ensured that only the most relevant to hand gesture recognition contributions from Google Scholar's top venues for computer vision were included in the survey, and resulted in 37 articles that were directly relevant to the task. Given the limited number of relevant studies (from Google Scholar), we widened the search base by conducting additional searches in all venues listed in Scopus using specific boolean search queries, which are listed in Table \ref{tab:queries}. The queries were carefully designed to contain the most relevant keywords to hand gesture recognition. To ensure consistency with the Google Scholar approach, we systematically reviewed the titles and abstracts of the matching papers from each query, assessing their alignment with the inclusion/exclusion criteria outlined in Section \ref{subsec:inc_exc_crit}. Articles were excluded if they lacked relevance to the hand gesture recognition task, its techniques, systems, or applications.

\begin{table}[h!]
\centering
\caption{\textcolor{black}{Overview of queries addressed to Scopus database and paper counts}}
\label{tab:queries}
\begin{tabular}{|p{3.2cm}|p{1cm}|p{1.1cm}|p{1cm}|}
\hline
\textbf{Query} & \textbf{Returned Papers} & \textbf{Candidate Papers} & \textbf{Selected Papers} \\
\hline
"RGB" AND ("hand" OR "gesture") AND ("recognition" OR 
"classification") & 1576 & 910 & 24 \\
\hline
"sign" AND "gesture" AND "recognition" & 2768 & 1442 & 32 \\
\hline
"Hand" AND ("gesture" OR "pose") AND ("recognition" OR "estimation") AND ("RGB" OR "video" OR "skeleton" OR "multi modal") & 2528 & 746 & 32 \\
\hline
"Multiview" AND "Camera" AND "Hand" AND "Gesture" & 5 & 1 & 1 \\
\hline
\end{tabular}
\end{table}

Table \ref{tab:queries} includes the following columns:
\begin{itemize}
    \item \textbf{Returned Papers:} The total number of papers retrieved from Scopus for each query.
    \item \textbf{Candidate Papers:} The number of papers remaining after duplicates were removed and topic modeling was applied.
    \item \textbf{Selected Papers:} The final number of papers chosen.
\end{itemize}

To determine the values in these columns, we employed a systematic methodology to curate the Scopus papers. The queries were executed sequentially, starting with the first query at the top of the table and progressing to the last query at the bottom. The process began with a broad query focused on terms such as "RGB," "hand," "gesture" and related recognition tasks, providing a foundation for further exploration. This was followed by an expansion into more specific domains, such as sign language recognition, addressed by the second query. To avoid redundancy, articles retrieved from the first query were excluded from the results of the second query. Similarly, the third query broadened the search to include tasks related to hand pose estimation and multimodal approaches, while ensuring duplicates from previous queries were removed. For each collection of retrieved papers, topic modeling using Non-Negative Matrix Factorization (NNMF) was performed in a similar way as detailed in Section \ref{subsec:topic_model}. Specifically, NNMF was applied to the titles, keywords, and abstracts of the papers to extract thematic topics. Articles with topics related to the research focus were retained, significantly narrowing down the number of candidate papers. Finally, a selection process was performed based on the inclusion/exclusion criteria outlined in Section \ref{subsec:inc_exc_crit}, resulting in the final set of papers used in this study. 

It is also worth noting that, despite sign language recognition generating a larger number of candidate papers, we chose not to select additional papers from this domain, in order to maintain a wider scope of hand gesture recognition from camera input, rather than focusing primarily on sign language. This reasoning also guided the continuation of queries related to hand gesture classification and estimation, as seen in the third query. This systematic approach ensured the curation of a high-quality collection of papers, specifically focusing on hand gesture recognition from camera input, for further analysis.

\subsection{Inclusion/Exclusion Criteria} \label{subsec:inc_exc_crit}
All selected articles have been rigorously reviewed and carefully analyzed based on the inclusion/exclusion criteria described in the following: 
\begin{itemize}
    \item Methodologies with human action recognition (entire body) that did not include hand estimation or classification were excluded, 
    \item Studies were included only if published between 2018 and 2025, 
    \item Only research publications available online (i.e., peer-reviewed conference proceedings, book chapters, and journal articles) were included, and
    \item Only studies written in English were considered.
\end{itemize}

All duplicate works that have been found in Scopus, while already in our list of Google Scholar results, were removed.

\subsection{Quantitative Analysis}

The search and filtering methodology described above helped us identify 125 studies (37 from Scholar and 88 from Scopus) that propose efficient methodologies for hand gesture recognition. To better understand how these studies are distributed over time and across different forums we performed a quantitative analysis, which is summarized in Table \ref{tab:quant_scholar} and Fig. \ref{fig:quant_topics}.
The visualizations provide a comprehensive overview of the research landscape of hand gesture recognition, revealing trends in publication frequency, preferred venues, and topics.

\subsubsection{Distribution of Articles to Venues} As shown in Table \ref{tab:quant_scholar}, the main forums where research in hand gesture recognition has been published include the major computer vision conferences such as CVPR, ICCV, ICCVW, ECCV, and the International Journal of Computer Vision. These forums list more than 50\% of the relevant works that contribute to advancements in this field.

\begin{table}[!htb]
\centering
\caption{\textcolor{black}{Distribution of Paper Venues.}}
\label{tab:quant_scholar}
\begin{adjustbox}{width=\columnwidth}
\begin{tabular}{|l|l|}
\hline
\textbf{Venue} & \textbf{Frequency} \\
\hline
\makecell[l]{IEEE/CVF Conference on \\ Computer Vision and Pattern Recognition} & 22.22\% \\\hline
\makecell[l]{IEEE/CVF International Conference on \\Computer Vision} & 11.11\% \\\hline
European Conference on Computer Vision & 8.33\% \\\hline
\makecell[l]{IEEE/CVF International Conference on\\ Computer Vision Workshops (ICCVW)} & 8.33\% \\\hline
International Journal of Computer Vision & 8.33\% \\\hline
\makecell[l]{IEEE International Conference on\\ Automatic Face \& Gesture Recognition} & 5.56\% \\\hline
\makecell[l]{IEEE/CVF Winter Conference on \\Applications of Computer Vision (WACV)} & 5.56\% \\\hline
International Conference on 3D Vision (3DV) & 5.56\% \\\hline
International Conference on Pattern Recognition & 5.56\% \\\hline
Asian Conference on Computer Vision (ACCV) & 2.78\% \\\hline
British Machine Vision Conference (BMVC) & 2.78\% \\\hline
Computer Vision and Image Understanding & 2.78\% \\\hline
\makecell[l]{IEEE Transactions on \\Pattern Analysis and Machine Intelligence (PAMI)} & 2.78\% \\\hline
\makecell[l]{IEEE/CVF Computer Society Conference \\on Computer Vision and \\Pattern Recognition Workshops (CVPRW)} & 2.78\% \\\hline
Journal of Visual Communication and Image Representation & 2.78\% \\\hline
Pattern Recognition & 2.78\% \\\hline
\end{tabular}
\end{adjustbox}
\end{table}

\subsubsection{Research Timeline by Topic} 
\begin{figure}[htbp]
\centering
\includegraphics[width=\columnwidth]{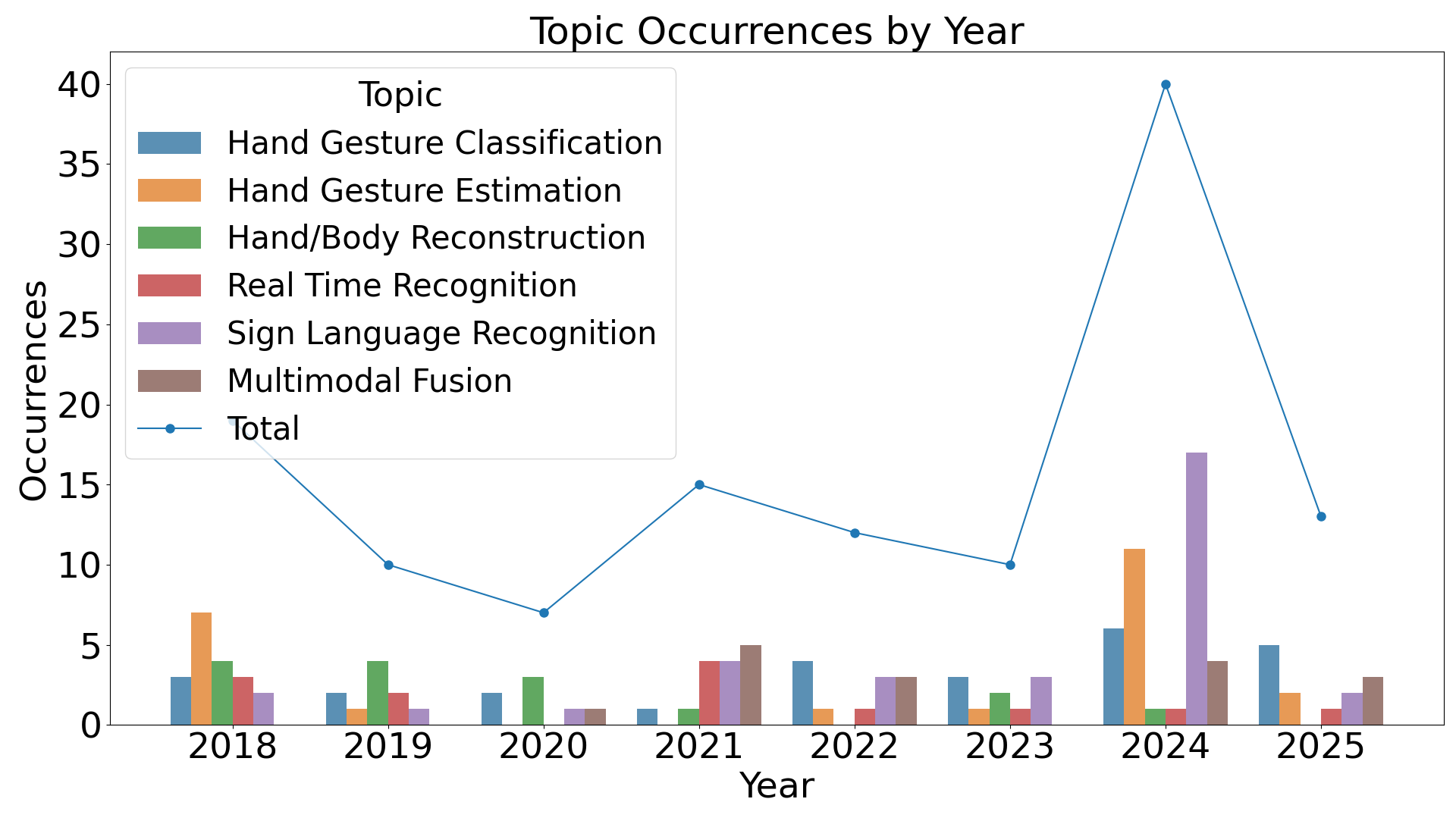}
\caption{Research Timeline by Topic.}
\label{fig:quant_topics}
\end{figure}

As shown in Fig. \ref{fig:quant_topics} the main topics discussed in the articles. The methodology for extracting the topics from the collection of articles is detailed in Section \ref{subsec:hand_gest_reco} that follows. The bar charts depict the annual occurrences of individual topics, while the line plot represents the total number of publications per year. The analysis reveals a substantial increase in publications in 2024, signaling growing research interest and advancements in the field of hand gesture recognition. The lower number of publications in 2025 is expected, as the year is still ongoing and many papers have yet to be published.

\subsection{Topic modeling} \label{subsec:topic_model}

\begin{figure}[!htbp]
\centering
\includegraphics[width=0.8\columnwidth]{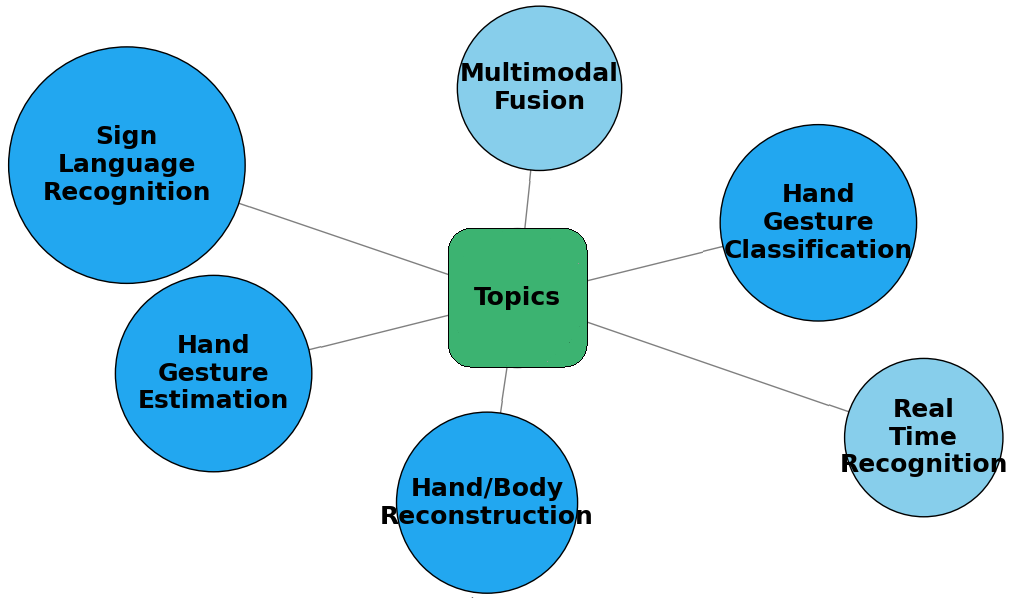}
\caption{Graph Representation of Hand Gesture Recognition Topics.}
\label{fig:topics_graph}
\end{figure}

In order to get a better understanding of the articles we retrieved, we decided to group them into topics, applying topic modeling \cite{paul2009topic}. The main topics are depicted in Fig. \ref{fig:topics_graph} and are described in the following.

\begin{figure*}[!b]
\centering
\includegraphics[width=0.8\textwidth]{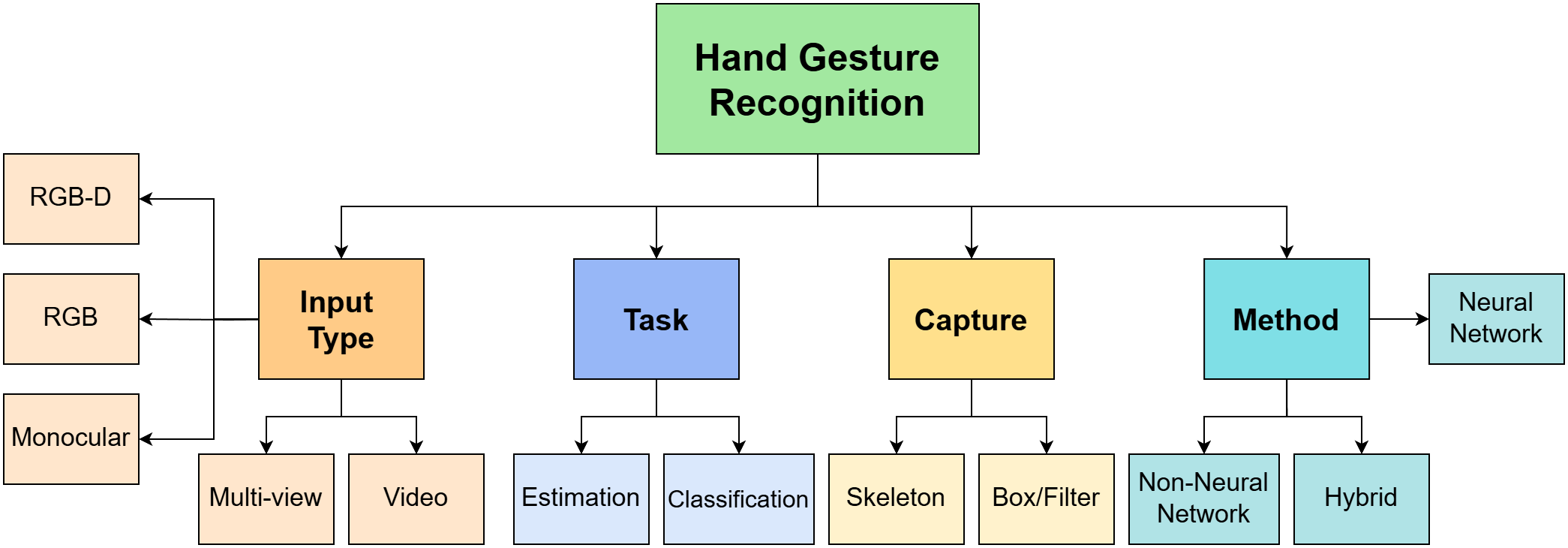}
\caption{\textcolor{black}{Categories of Hand Gesture Recognition.}}
\label{fig:hgr_cat}
\end{figure*}

Among the various methods available for topic modeling, such as Latent Dirichlet Allocation (LDA) \cite{lda_topic}, Latent Semantic Analysis (LSA), Non-Negative Matrix factorization (NNMF) and clustering algorithms like K-Means or DBSCAN \cite{cluster_topic}, we employed Non-Negative Matrix Factorization (NNMF) \cite{nmf_topic1, nmf_topic2} for this task. NNMF is a linear algebra technique that decomposes a high-dimensional dataset into two lower-dimensional matrices with non-negative entries, making it particularly suitable for extracting interpretable topics. NNMF has been applied to the vectorized representations of the titles and keywords from the collected papers. This allowed us to uncover latent semantic structures, effectively identifying recurring article themes. After experimenting with various configurations, we determined that using six components produced the most coherent topics. To facilitate presentation and ensure clarity, the resulting topics were manually refined into the following:

\begin{itemize}
\item \textbf{Hand Gesture Classification}: Focuses on categorizing different types of hand gestures into predefined classes.
\item \textbf{Hand Gesture Estimation}: Involves determining the 3D position and orientation of hand joints.
\item \textbf{Sign Language Recognition}: Translates hand gestures into a corresponding sign language vocabulary.
\item \textbf{Hand/Body Reconstruction}: Deals with accurate 3D model creation/estimation of hands or bodies.
\item \textbf{Multimodal Fusion}: Combines multiple models and integrating data from diverse sources (e.g., video, depth cameras) to improve accuracy and robustness.
\item \textbf{Real-Time Recognition}: Emphasizes systems capable of recognizing and processing gestures in real-time.
\end{itemize}

In Fig. \ref{fig:topics_graph}, the topics are represented as nodes, with larger nodes indicating a higher number of articles associated with the topic. The figure reveals that Hand Gesture Classification, Hand Gesture Estimation, and Sign Language Recognition are the most widely discussed topics in the related literature, likely because they correspond to broader tasks commonly addressed in the computer vision research domain. This further demonstrates that NNMF successfully captured the differentiation discussed in Section \ref{subsec:def_gesture}, particularly between gesture classification and estimation. Hand/Body Reconstruction, although having fewer articles, can also be considered a task as it represents a narrower subcategory of Hand Gesture Estimation. However, its focus on indirect hand gesture estimation naturally results in fewer articles being assigned compared to the broader parent category. The remaining topics—Multimodal Fusion and Real-Time Recognition—are more specific and often describe techniques or strategies employed to tackle the aforementioned tasks. These categories were appropriately identified by NNMF as distinct due to their notable contributions, underscoring their importance as complementary topics.

\subsection{Key Takeaways of this section}
\textcolor{black}{This section detailed the systematic approach undertaken to select, analyze, and categorize the research papers included in this survey. The core aspects of our methodology are:
\begin{itemize}
    \item \textbf{Article Selection:} A systematic process was employed, starting with top-ranked computer vision venues (Google Scholar) and expanding with targeted Scopus queries for papers published between 2018-2025.
    \item \textbf{Inclusion/Exclusion Criteria:} Focused on studies on hand gesture recognition/estimation from visual input (images/videos), excluding whole-body action recognition without specific hand analysis, and non-English or unavailable publications.
    \item \textbf{Quantitative Analysis:} 125 relevant studies were identified. Analysis revealed CVPR, ICCV, ECCV as major publication venues. A significant increase in HGR publications was noted, especially in 2024.
    \item \textbf{Topic Modeling:} Non-Negative Matrix Factorization (NNMF) was applied to titles and keywords to identify six primary research themes: Hand Gesture Classification, Hand Gesture Estimation, Sign Language Recognition, Hand/Body Reconstruction, Multimodal Fusion, and Real-Time Recognition.
\end{itemize}}

\section{Overview of HGR methods} \label{sec:methods}

Hand gesture recognition (HGR) has been the focus of extensive research, resulting in a wide range of methodologies. This section provides a comprehensive review of the methods found in the current literature, classified based on key factors such as the primary task objective (hand gesture classification or estimation), the type of input devices (e.g., RGB, RGB-D, or video, monocular or multi-view cameras), the type of hand gesture capture and representation (e.g., skeleton or box/filter), and the machine learning techniques employed in the recognition (e.g., SVMs, neural networks). These classifications are also summarized and visualized in Fig. \ref{fig:hgr_cat}.

\subsection{Classify by task objective}

As broadly described in Section \ref{sec:basics}, we can distinguish two main hand recognition tasks: hand gesture classification and hand gesture estimation. While this survey focuses primarily on hand gesture classification, the significance of the hand gesture estimation task cannot be overlooked, since it is supported by a crucial subset of general hand gesture recognition methods.
In what follows, we briefly present the main objectives of the two tasks and discuss their main applications.

\begin{figure*}[!htb]
    \centering
    \begin{subfigure}[b]{0.45\textwidth}
            \centering
            \includegraphics[width=\textwidth]{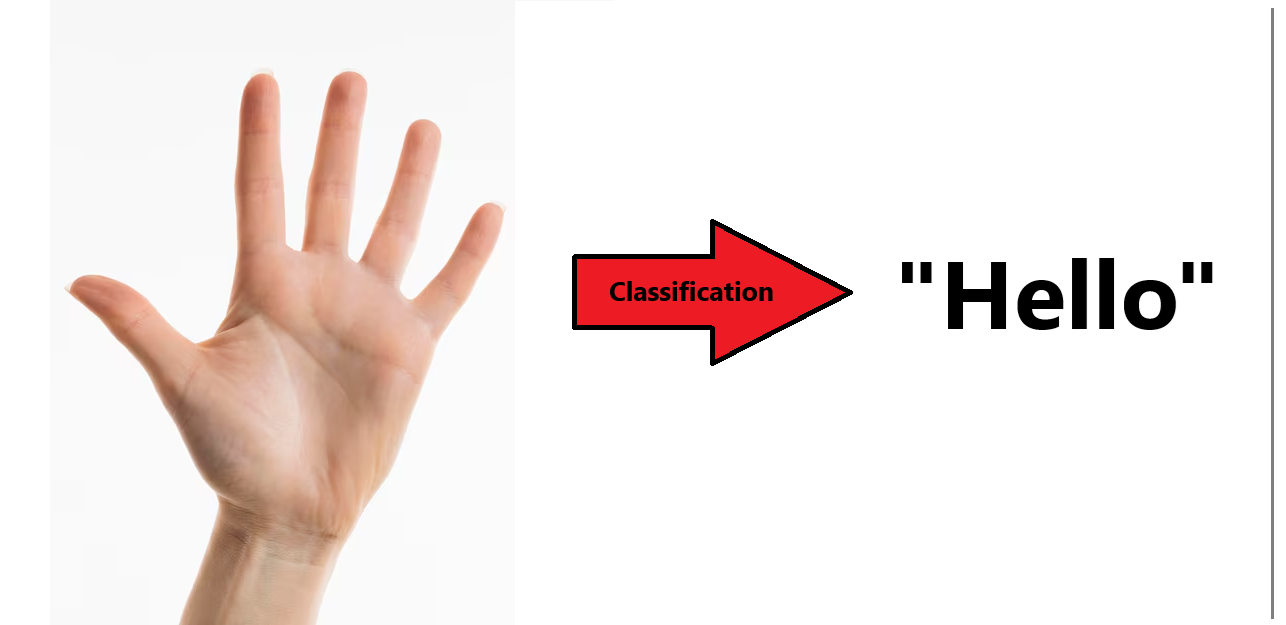}
            \caption{Hand Gesture Classification.}
            \label{sfig:hand_classification}
    \end{subfigure}
    \begin{subfigure}[b]{0.45\textwidth}
            \centering
            \includegraphics[width=\textwidth]{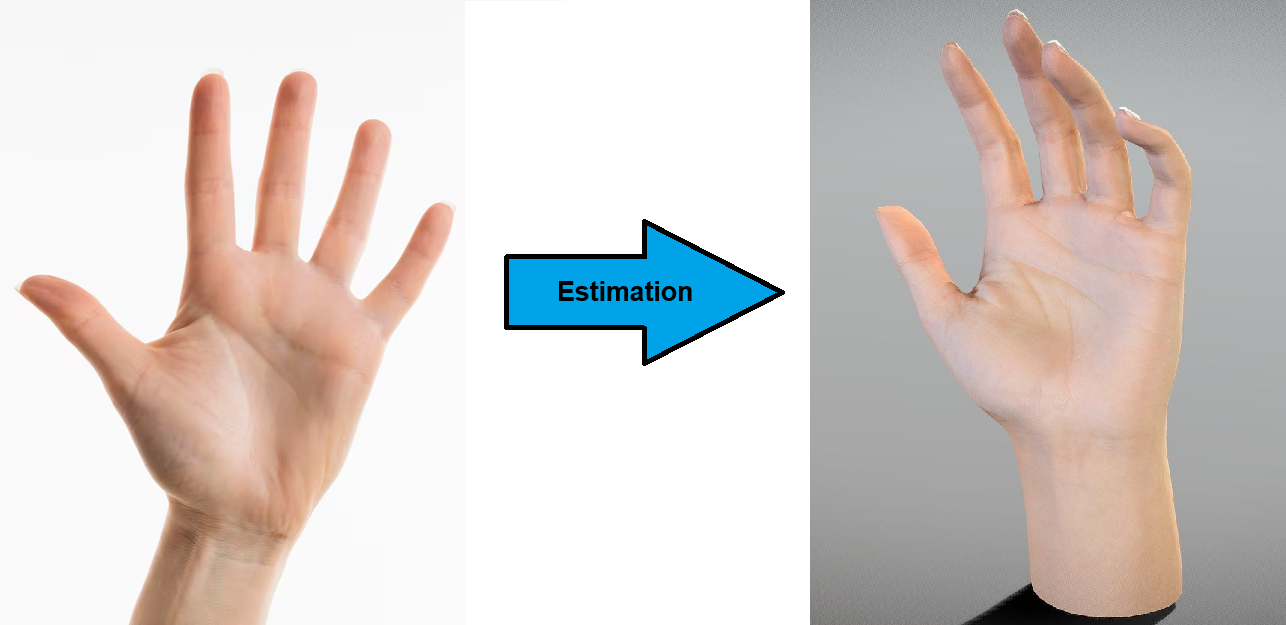}
            \caption{Hand Gesture Estimation.}
            \label{sfig:hand_estimation}
    \end{subfigure}%
    \caption{Hand Recognition Objectives.}
    \label{fig:hand_goal}
\end{figure*}

\subsubsection{Hand Gesture Classification}  
The primary goal of hand gesture classification is to identify and categorize a given gesture into predefined classes \cite{ABDULLAHI2024123258, 9505250, noreen_hamid_akram_malik_saleem_2021}. As a result, the respective methods focus on interpreting the semantic meaning of gestures rather than their capturing and predicting their spatial representation. Hand gesture classification methods are widely used in applications like sign language recognition \cite{article, SADEGHZADEH2024200384, 10560333} and human-robot communication \cite{9903078, BAMANI2024108443}. 

In sign language recognition, the classifier distinguishes between subtle variations in gestures to correctly identify words or phrases, as shown in Fig. \ref{fig:hand_goal}a. Similarly, in HCI, classification methods enable systems to interpret hand gestures to user commands that are then used to control devices or software applications. Hand classification approaches are often evaluated based on metrics such as accuracy, precision, recall, and F1-score, which reflect their ability to correctly classify gestures. The broad applicability of this task makes it an accessible and popular area of research, particularly with the growing availability of related image/video datasets for training and testing \cite{9622242, Materzynska_2019_ICCV, 9093512, joze2018ms}.

\subsubsection{Hand Gesture Estimation}  
Hand gesture estimation focuses on understanding and representing the topology or spatial structure of the hand \cite{SHANMUGAM2024123351, cai2018weakly, Iqbal_2018_ECCV}. Unlike classification, which assigns a gesture to a specific category, estimation methods aim to digitally recreate the hand's pose or movement, as shown in Fig. \ref{fig:hand_goal}b. This is typically achieved by generating a 2D or 3D representation, such as a skeleton with joints \cite{10466755, 10030397} or a detailed mesh of the hand's surface \cite{Pavlakos_2019_CVPR, Rong_2021_ICCV}.

The applications of hand gesture estimation are mainly in areas that require detailed spatial information, such as augmented reality (AR), virtual reality (VR), gaming, and animation. For example, in a VR environment, hand estimation methods allow users to interact with virtual objects through accurate hand tracking \cite{SWAMY2024104073, SHUANG2024104129}.

In some cases, hand gesture recognition systems combine both classification and estimation objectives to achieve a comprehensive understanding of gestures \cite{10233392, 10582035, AVOLA2022108762}. These hybrid approaches usually estimate the hand’s pose or topology (hand estimation), and then use this representation to classify the gesture (hand classification). This two-step process is particularly useful in scenarios where both precise hand tracking and semantic understanding are required.

\subsection{Classify by Input Data type}

The type of input data is an important factor in defining the effectiveness and scope of a hand gesture recognition method. This survey focuses on methods utilizing static images or video, which are widely used input data types due to the widespread availability of cameras and their accessibility to users. One classification is 
based on the format of the input data, which can be either RGB or RGB-D images, or video. A different classification is based on the type of camera, namely monocular or multi-view. In the following, we provide the main features of each of these input categories. 

\subsubsection{RGB Methods}  
RGB methods rely on static RGB images without considering sequential video frames, so the works of this category analyze a single frame to recognize/estimate static hand gestures \cite{KWOLEK2021586, 8354158, ZHOU2022102226}. Although these methods are limited to processing one frame at a time, they can be easily extended to handle multiple frames for video-based hand gesture recognition. However, such extensions are usually less efficient than methods specifically designed for video gesture recognition.

\subsubsection{RGB-D Methods}  
RGB-D methods utilize depth image data, which combines traditional RGB information with depth data to enhance hand gesture recognition \cite{WANG2018404, supanvcivc2018depth, 9760658}. These methods typically work on a single RGB-D frame at a time to capture gesture details, providing richer spatial information compared to RGB-only approaches.

\subsubsection{Video Methods}  
Video-based methods process sequences of RGB frames \cite{mueller2018ganerated, Cai_2019_ICCV, 10050006}, or even RGB-D frame sequences \cite{TERRERAN2023104523, 9523142, SWAMY2024104073, chen_li_fang_xin_lu_miao_2022}. In the case of RGB-D video sequences, depth data accompanies each frame to provide a more comprehensive understanding of hand gestures. The respective methods in this category are designed for analyzing dynamic hand movements in tasks such as sign language recognition or complex gesture analysis.

\begin{figure*}[htbp]
    \centering
    \begin{subfigure}[b]{0.5\textwidth}
            \centering
            \includegraphics[width=\textwidth]{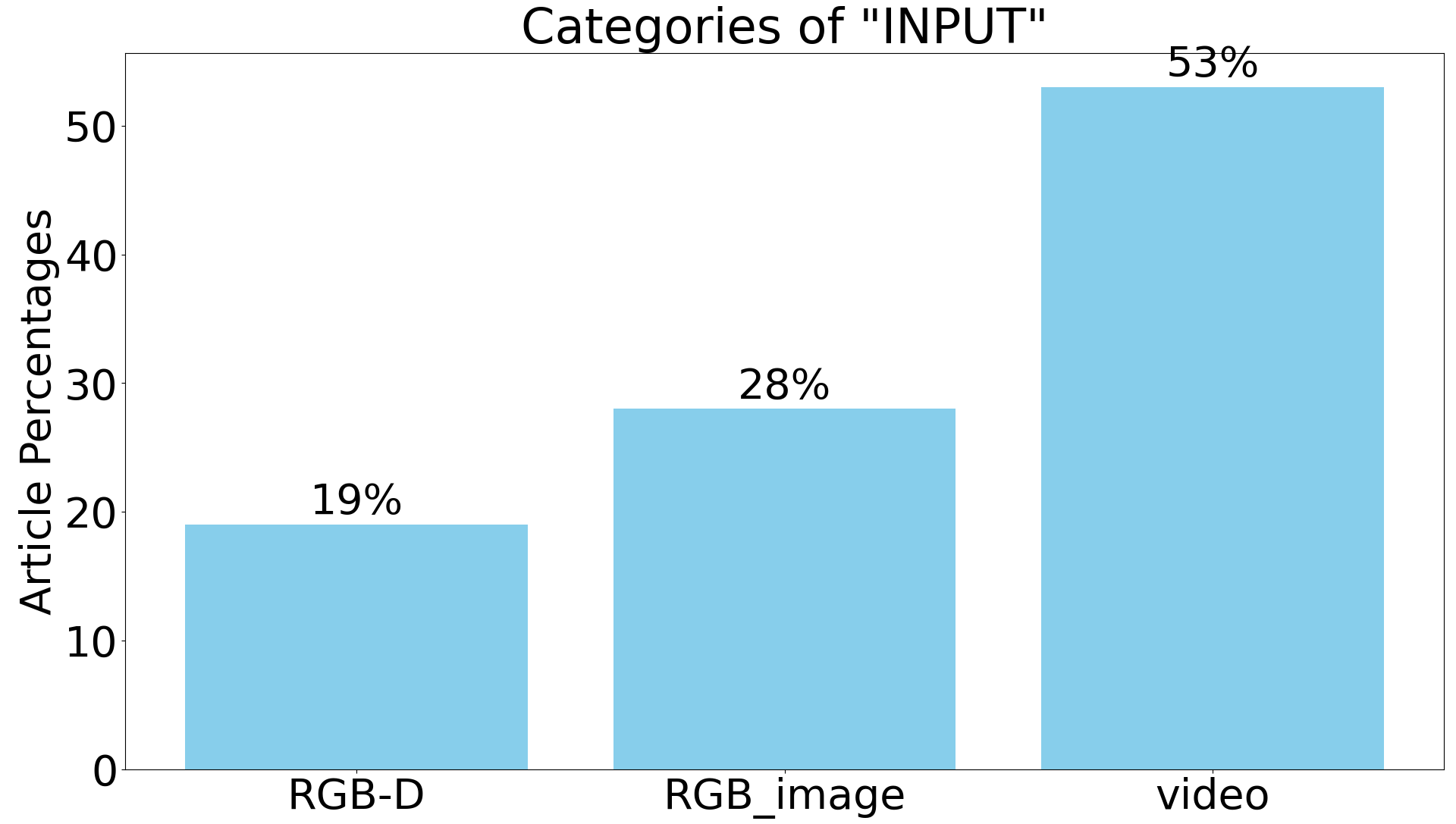}
            \caption{Number of articles for RGB, RGB-D and video input.}
            \label{sfig:input}
    \end{subfigure}%
    \begin{subfigure}[b]{0.5\textwidth}
            \centering
            \includegraphics[width=\textwidth]{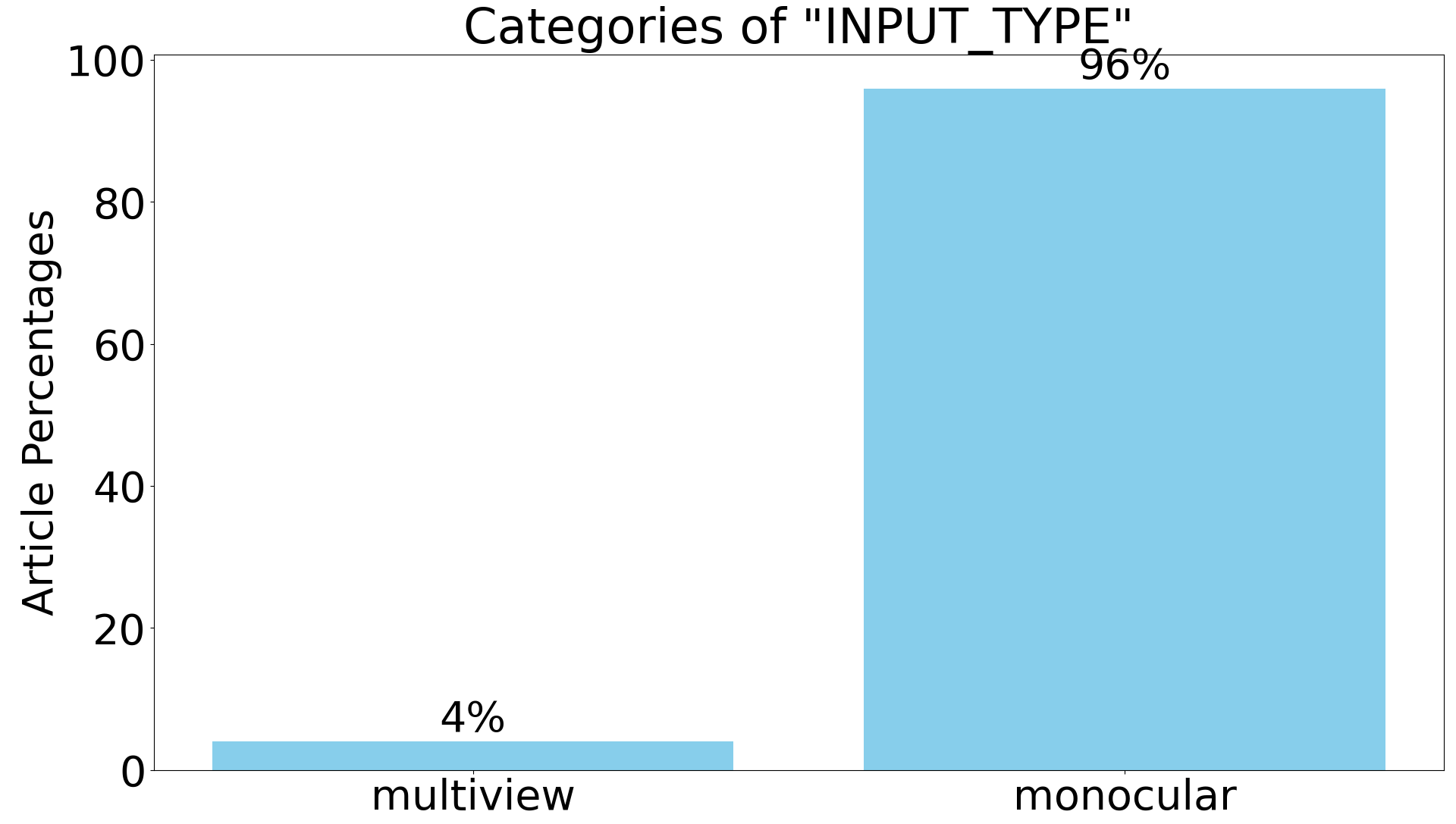}
            \caption{Number of articles for monocular and multiview camera input.}
            \label{sfig:input_type}
    \end{subfigure}
    \caption{The distribution of articles per input data (RGB, RGB-D and video) and input device type (monocular and multi-view cameras).}
    \label{fig:hand_input}
\end{figure*}

As shown in Fig. \ref{sfig:input}, video-based methods represent a significant portion of the studies reviewed in this survey, accounting for 53\% of the total number of articles. This prevalence can be attributed to the inherent versatility of video data, which not only incorporates static RGB frames but also effectively captures the dynamic aspects of hand motion. These qualities make video-based approaches particularly well-suited for a wide range of hand recognition tasks. Following video methods, static RGB images are used in 28\% of the articles, while RGB-D images account for 19\%.

\subsubsection{Monocular vs. Multi-View Input}  
The type of camera setup—monocular or multi-view—further influences the methodology of hand gesture recognition. Monocular methods rely on a single camera to capture hand gestures, whereas multi-view methods use multiple cameras from different angles to provide a more complete representation of the hand. 

Monocular methods dominate the field due to their simplicity, cost-effectiveness, and ease of deployment in real-life scenarios \cite{DECASTRO2023119394, 10049842, app14198937}. In contrast, multi-view methods, while offering improved accuracy and robustness in capturing complex gestures, require intricate setups involving multiple cameras, which limits their practicality for everyday use.

Despite their limited prevalence, multi-view setups are worth mentioning, particularly for applications in hand gesture estimation where capturing detailed 3D spatial information is critical \cite{joo2018total, corona2022lisa, 8600529}. However, as the survey progressed, the distinction between monocular and multi-view methods became less significant, with most articles opting for monocular setups (as also depicted in Fig. \ref{sfig:input_type}) due to their practicality and widespread applicability.

\subsection{Classify by Hand Gesture capture and representation}

Hand capture methods play an important role in hand gesture recognition systems as they determine how hand gestures are acquired for further processing. These methods can be broadly classified into skeleton-based captures and box/filter-based captures, each offering unique advantages and challenges.


\subsubsection{Skeleton-Based Captures}
\begin{figure}[htbp]
\centering
\includegraphics[width=\columnwidth]{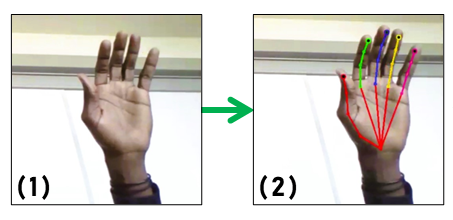}
\caption{\textcolor{black}{Example of skeleton-based hand capture. Source \cite{10216066}.}}
\label{fig:hand_skeleton_cap}
\end{figure}

Skeleton-based methods represent the hand using a skeletal structure that consists of key joints and connections \cite{Shi_2020_ACCV, DENG2024127194, 10456765}. These captures typically utilize graphs, where nodes represent specific hand joints or segments such as fingers, and edges denote the relationships or connections between them. This graph-based representation enables the system to focus on the underlying structure and movement dynamics of the hand. By abstracting the hand into a simplified skeleton, these methods are particularly effective in scenarios where the precise articulation of fingers and joints is important. \textcolor{black}{An example of an approach using skeleton based capture is shown in Fig. \ref{fig:hand_skeleton_cap}.}

\subsubsection{Box/Filter-Based Captures}
\begin{figure}[htbp]
\centering
\includegraphics[width=\columnwidth]{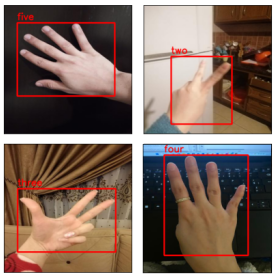}
\caption{\textcolor{black}{Example of box/filter-based hand capture. Source \cite{Sharma_2023_Hand-Gesture-Recognition-YOLOv8-OAK-D}.}}
\label{fig:hand_box_cap}
\end{figure}

Box/filter-based methods follow a different approach, focusing on capturing the hand's region using bounding boxes or regions of interest \cite{alaftekin2024real, 10560333, PigouODHD15}. These methods involve detecting the hand from the image using a bounding box and once the hand is isolated, applying additional filters or algorithms to extract meaningful features related to the gesture, such as shape, texture, or movement patterns. \textcolor{black}{An example of an approach using box/filter based capture is shown in Fig. \ref{fig:hand_box_cap}.}

\textcolor{black}{
The fundamental difference between these approaches lies in how they represent and process hand information. Skeleton-based captures excel in scenarios that require precise motion analysis and are robust against background noise because they directly model the hand's articulated structure as a set of interconnected joints and segments. This detailed skeletal representation is then typically processed using graph-based algorithms or fed into models designed to understand structural relationships. However, this level of detail often requires more computational resources, both for initially extracting the skeleton and for its subsequent graph-based representation and analysis, making these methods more suitable for applications where accuracy and nuanced articulation are prioritized over raw speed or implementation simplicity.
}

\textcolor{black}{
In contrast, box/filter-based methods adopt a two-stage approach: first, they focus on localizing the hand as a whole within a broader image, typically using a bounding box or region of interest. Once this region is identified, various features (e.g., shape descriptors, texture patterns) are extracted from the pixel data within that box. This makes them well-suited for working with 2D image data and readily adaptable across diverse environments. These methods are generally more user-friendly to implement and are widely applied in hand gesture classification tasks (as described in Section \ref{subsec:bayes_groups}, where Bayes' Theorem demonstrates that box/filter methods are predominantly used in classification tasks). Here, the primary goal is to categorize the overall gesture's appearance or motion pattern, rather than to perform detailed estimation tasks which necessitate precise joint locations and orientations. Despite their advantages in simplicity and directness, box/filter-based methods can be more sensitive to external factors such as inconsistent lighting conditions, cluttered backgrounds that interfere with hand segmentation, and variations in hand appearance, all of which can negatively impact recognition performance by corrupting the features extracted from the hand region.
}

\begin{figure}[htbp]
\centering
\includegraphics[width=\columnwidth]{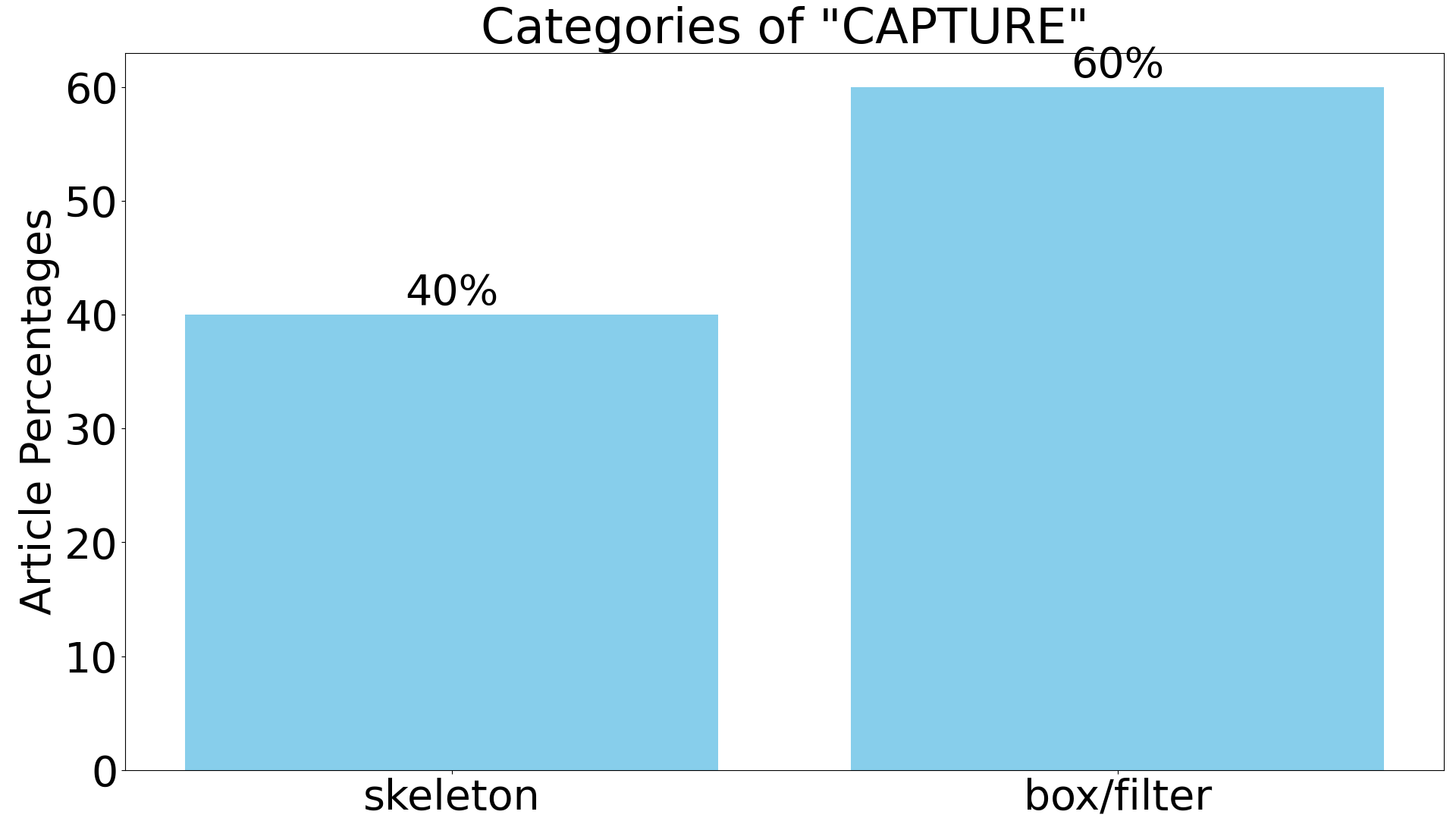}
\caption{Number of articles per hand capture method.}
\label{fig:hand_cap_stats}
\end{figure}

It is worth noting that box/filter-based approaches have been studied more extensively in the literature compared to skeleton-based methods (as shown in Fig. \ref{fig:hand_cap_stats}). The greater prevalence of box/filter-based methods can likely be attributed to their relative simplicity, ease of implementation, and broader applicability across various scenarios, making them a preferred choice for researchers in the field.

\subsection{Classify by Recognition Techniques}

Hand gesture recognition employs a variety of techniques to extract gesture-specific information from RGB data. In recent years, the combination of convolutional neural networks (CNNs) with other neural network-based methods has emerged as the dominant approach, leveraging the strengths of multiple models. This trend is evident in the findings of this survey, where 68\% of the selected articles employ such hybrid methods for hand gesture recognition. Nonetheless, standalone neural networks or traditional machine learning techniques, such as Support Vector Machines (SVMs) and Random Forests, continue to play a role in specific applications. Based on the methodologies used, the articles can be categorized into three groups: hybrid approaches, neural network-based approaches, and non-neural network approaches as discussed in the following.

\subsubsection{Neural Network Approaches}
Neural networks, particularly CNNs \cite{9491897, singh2024impact, 10506333}, have been extensively applied to hand gesture recognition due to their ability to learn spatial hierarchies and extract complex features directly from RGB data. Recent advancements, such as transformers \cite{10484197}, have further enriched this field by enabling models to capture temporal and contextual information. This capability is especially valuable for applications like sign language recognition and human-computer interaction (HCI), where understanding the context of previous gestures is crucial.

\subsubsection{Non-Neural Network Approaches}

Traditional machine learning methods, such as SVMs, remain relevant for specific hand gesture recognition tasks \cite{10.1371/journal.pone.0298699, s24030826, BAHUGUNA2024106203}. These methods typically involve pre-processing input images or video frames to extract features, which are then fed into machine learning models for classification or regression. While less common in recent studies, these approaches are still employed in scenarios where computational efficiency or simplicity is a priority.

\subsubsection{Hybrid Approaches}

Hybrid approaches represent the predominant methodology in hand gesture recognition, combining the strengths of multiple techniques to achieve higher accuracy and robustness. For instance, methods integrating CNNs with LSTMs leverage CNNs for spatial feature extraction and LSTMs for temporal pattern learning \cite{electronics11152427, koller2019weakly, NUNEZ201880}. Other hybrid systems blend neural networks with traditional machine learning techniques in ensemble models or integrated frameworks \cite{9760658, 9432163}. These approaches are particularly effective in tasks requiring adaptability and precision, as they exploit the complementary strengths of their components, offering a versatile solution for complex hand gesture recognition challenges.

\begin{figure}[htbp]
\centering
\includegraphics[width=\columnwidth]{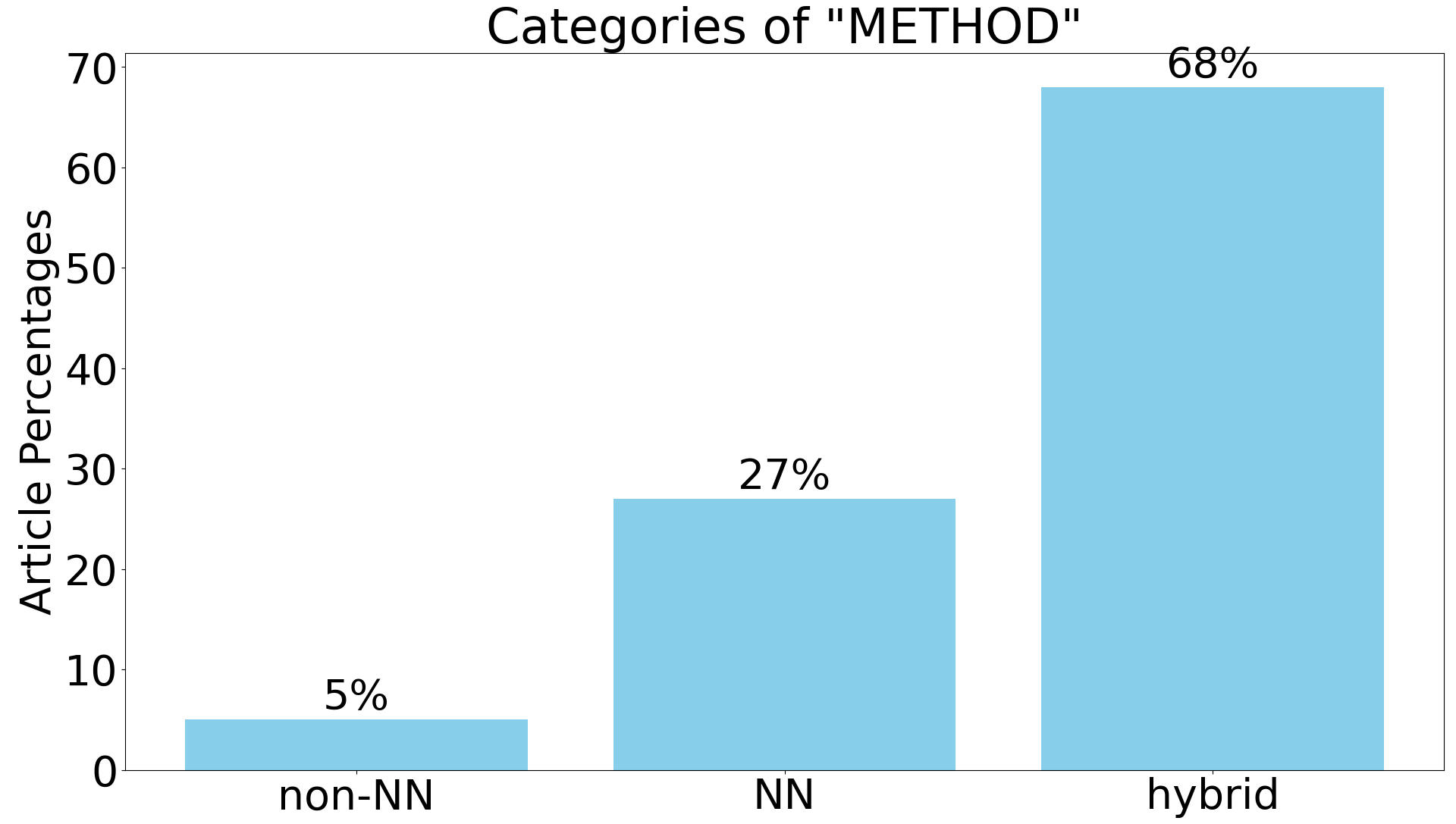}
\caption{Number of articles per recognition method.}
\label{fig:hand_methods}
\end{figure}

The distribution of articles among these recognition methods is summarized in Fig. \ref{fig:hand_methods}, providing a quantitative overview of their prevalence in the field.

\subsection{Summary of Methodology Overview}

To provide a clear understanding of the groupings, Table \ref{tab:excel} summarizes the classification of the articles based on the criteria described in this section. Specifically, these criteria include:

\begin{table*}[!htb]
\centering
\caption{\textcolor{black}{Summary of HGR Methods based on classification criteria of Section \ref{sec:methods}.}}
\label{tab:excel}
\begin{adjustbox}{width=\textwidth}
\begin{tabular}{lllllp{5cm}}
\specialrule{1.5pt}{0pt}{0pt} 
\textbf{Input} & \textbf{Input type} & \textbf{Capture} & \textbf{Task} & \textbf{Method} & \textbf{Article} \\
\specialrule{1.5pt}{0pt}{0pt} 
\multirow{2}{*}{\textbf{RGB-D}} & \multirow{2}{*}{monocular} & box/filter & classification & NN & \cite{noreen_hamid_akram_malik_saleem_2021, FANG2025111653, 10.1145/3689644} \\ \cline{3-6}
 &  & skeleton & estimation & hybrid & \cite{WANG2018404, 8490961} \\\hline
\multirow{4}{*}{\textbf{RGB-D+RGB\_image}} & \multirow{4}{*}{monocular} & \multirow{3}{*}{box/filter} & classification & NN & \cite{9505250} \\
 & & & classification & hybrid & \cite{9760658, 9432163} \\
 & & & estimation & hybrid & \cite{10637398} \\\cline{3-6}
 & & skeleton & estimation & hybrid & \cite{cai2018weakly, liu_ren_gao_wang_sun_qi_zhuang_liao_2024} \\\hline
\multirow{8}{*}{\textbf{RGB-D+video}} & \multirow{8}{*}{monocular} & \multirow{4}{*}{box/filter} & classification & NN & \cite{9796020} \\
 &  &  & classification & hybrid & \cite{Kopuklu_2018_CVPR_Workshops, chen_li_fang_xin_lu_miao_2022, 9903078, zhou2021adaptive} \\
 &  &  & estimation & non-NN & \cite{supanvcivc2018depth} \\
 &  &  & estimation & hybrid & \cite{SHANMUGAM2024123351, SWAMY2024104073} \\\cline{3-6}
 &  & \multirow{4}{*}{skeleton} & classification & NN & \cite{Shi_2020_ACCV, 10444226, LIU2025111343} \\
 &  & & classification & hybrid & \cite{8545718, BALAJI2024104019, TERRERAN2023104523, 9523142, 9525136} \\
 &  & & classification & non-NN & \cite{DESMEDT201960} \\
 &  & & classification+estimation & NN & \cite{garcia2018first} \\\hline
\multirow{11}{*}{\textbf{RGB\_image}} & \multirow{10}{*}{monocular} & \multirow{5}{*}{box/filter} & classification & NN & \cite{article, alaftekin2024real, 10560333, SHARMA2021115657, 9491897, technologies13040164} \\
 &  & & classification & hybrid & \cite{BAMANI2024108443, 10452793, SADEGHZADEH2024200384, ZHOU2022102226, s22165959, KWOLEK2021586, bahuguna2025hybrid} \\
 &  & & classification & non-NN & \cite{BAHUGUNA2024106203, 10.1371/journal.pone.0298699} \\
 &  & & estimation & NN & \cite{9320438, Rong_2021_ICCV} \\
 &  & & estimation & hybrid & \cite{Pavlakos_2018_CVPR, Saito_2020_CVPR, Kolotouros_2019_CVPR, Kolotouros_2019_ICCV, 10404038, JIAO2024155} \\ \cline{3-6}
 &  & \multirow{5}{*}{skeleton} & classification & NN & \cite{singh2024impact} \\
 &  &  & classification & hybrid & \cite{wang2024mobilenet, alabduallah2025innovative} \\
 &  &  & classification+estimation & hybrid & \cite{AVOLA2022108762} \\
 &  &  & estimation & NN & \cite{Iqbal_2018_ECCV, 8354158, Varol_2018_ECCV, 10484197, LI2025122003, LIU2025107221, SORLI2025104200} \\
 &  &  & estimation & hybrid & \cite{Pavlakos_2019_CVPR, 10466755, SHUANG2024104129} \\\cline{2-6}
 & multiview & skeleton & estimation & hybrid & \cite{Zimmermann_2019_ICCV} \\\hline
\multirow{14}{*}{\textbf{Video}} & \multirow{10}{*}{monocular} & \multirow{7}{*}{box/filter} & classification & NN & \cite{app11094164, 9093512, Oguntimilehin2024158, 9984249, TALAAT2024109475, FENG2025107469, WU202577} \\
 &  &  & classification & hybrid & \cite{RASTGOO2024123349, koller2019weakly, koller2018deep, kopuklu2019real, Materzynska_2019_ICCV, 8545095, Jiang_2019_ICCV, Munro_2020_CVPR, EEI6059, electronics13071229, 10577129, KALANDYK2024107879, https://doi.org/10.1049/cvi2.12240, DECASTRO2023119394, 9713865, 9533707, 10.1145/3595916.3626411, ali2023snapture, app14198937, 9786744, electronics11152427, 9590493, 9523108, 9483704} \\
 &  &  & estimation & NN & \cite{10506333} \\
 &  &  & estimation & hybrid & \cite{Kocabas_2020_CVPR, 10483600, 10268564} \\\cline{3-6}
 &  & \multirow{7}{*}{skeleton} & classification & NN & \cite{devineau2018deep, DENG2024127194, 10049842, 10452001, GUAN2025111602} \\
 &  &  & classification & hybrid & \cite{ABDULLAHI2024123258, NUNEZ201880, joze2018ms, tripathi_verma_2024, 10007834, 10581950, 10456765, 9601572} \\
 &  &  & classification+estimation & hybrid & \cite{10233392, 10582035} \\
 &  &  & classification+estimation & non-NN & \cite{s24030826} \\
 &  &  & estimation & NN & \cite{mueller2018ganerated, Cai_2019_ICCV} \\
 &  &  & estimation & hybrid & \cite{10050006} \\\cline{2-6}
 & \multirow{4}{*}{multiview} & box/filter & classification & non-NN & \cite{8600529} \\\cline{3-6}
 &  & \multirow{3}{*}{skeleton} & classification & hybrid & \cite{9205781} \\
 &  & & estimation & NN & \cite{corona2022lisa, guan2025multi} \\
 &  & & estimation & hybrid & \cite{joo2018total} \\
\specialrule{1.5pt}{0pt}{0pt} 

\end{tabular}
\end{adjustbox}
\end{table*}

\begin{itemize}
    \item Type of input data: video, RGB-D, (static) RGB\_image
    \item Input configuration: monocular or multiview
    \item Hand capture method: box/filter or skeleton
    \item Goal of hand recognition: estimation or classification
    \item Method utilized: Neural Network (NN), non-Neural Network (non-NN), or hybrid
\end{itemize}

Combinations of values within a specific group/column are indicated using the "+" character. For instance, an entry of RGB-D+video in the "Input Type" column signifies that the referenced articles utilize a combination of RGB-D and video data for hand gesture recognition. This notation allows for a concise representation of articles employing multiple values of a group.

\subsection{Association between input, input type, capture and hand recognition task} \label{subsec:bayes_groups}
To better understand the relationships between the hand gesture categories discussed in Section \ref{sec:methods} and the hand recognition task (classification or estimation), we utilized Bayes' theorem. Bayes' theorem allows us to determine the probability of Y occurring given that X is true. The formula is as follows:

\begin{equation*} P(Y|X) = \frac{P(X|Y) \cdot P(Y)}{P(X)} \end{equation*}

where:
\begin{itemize}
    \item $P(Y|X)$ is the conditional probability of Y given X,
    \item $P(X|Y)$ is the likelihood of X given Y,
    \item $P(Y)$ is the prior probability of Y, and
    \item $P(X)$ is the marginal probability of X.
\end{itemize}

In our analysis, X represents the groups: \textit{input}, \textit{input\_type}, \textit{capture}, and \textit{method}, and Y represents the \textit{goal}, which can be either \textit{estimation} or \textit{classification}. For the groups \textit{input} and \textit{method}, we extended Bayes' theorem to consider joint probabilities, as these categories include three distinct values each. To simplify the analysis and derive meaningful insights, we merged the values within these categories as follows:

\begin{itemize}
    \item \textbf{Input groups $\rightarrow$ RGB and Video vs. RGB-D:} Static RGB images and video data were combined into a single category to compare with RGB-D inputs. This grouping reflects the fact that RGB and video methods both rely solely on color information, whereas RGB-D methods include depth information, making them inherently different in terms of hardware requirements and data richness.
    \item \textbf{Method groups $\rightarrow$ Non-NN and Hybrid Methods vs. NN Methods:} Non-neural network (Non-NN) methods and hybrid methods were grouped together to contrast with purely neural network (NN) approaches. This grouping emphasizes the distinction between traditional machine learning approaches (Non-NN methods) and neural networks (NN). Hybrid methods were included with Non-NN methods as they sometimes incorporate Non-NN methods as submodels.
\end{itemize}

\begin{figure}[htb]
     \centering
     \includegraphics[width=0.9\columnwidth]{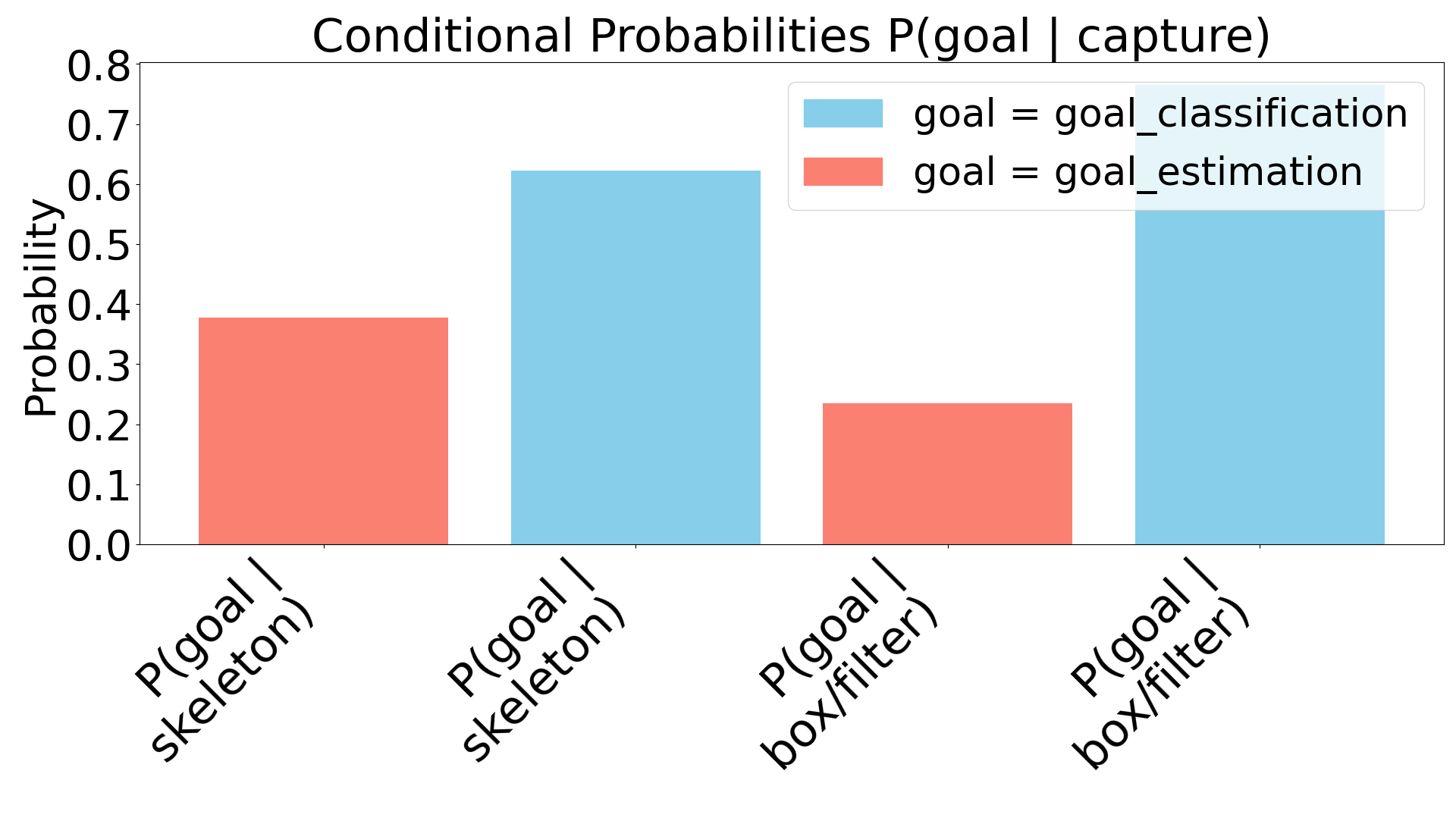}
     \caption{\textcolor{black}{Conditional Probability of goal given capture category.}}
     \label{sfig:bayes1}
\end{figure}
\begin{figure}[htb]
    \centering
    \includegraphics[width=0.9\columnwidth]{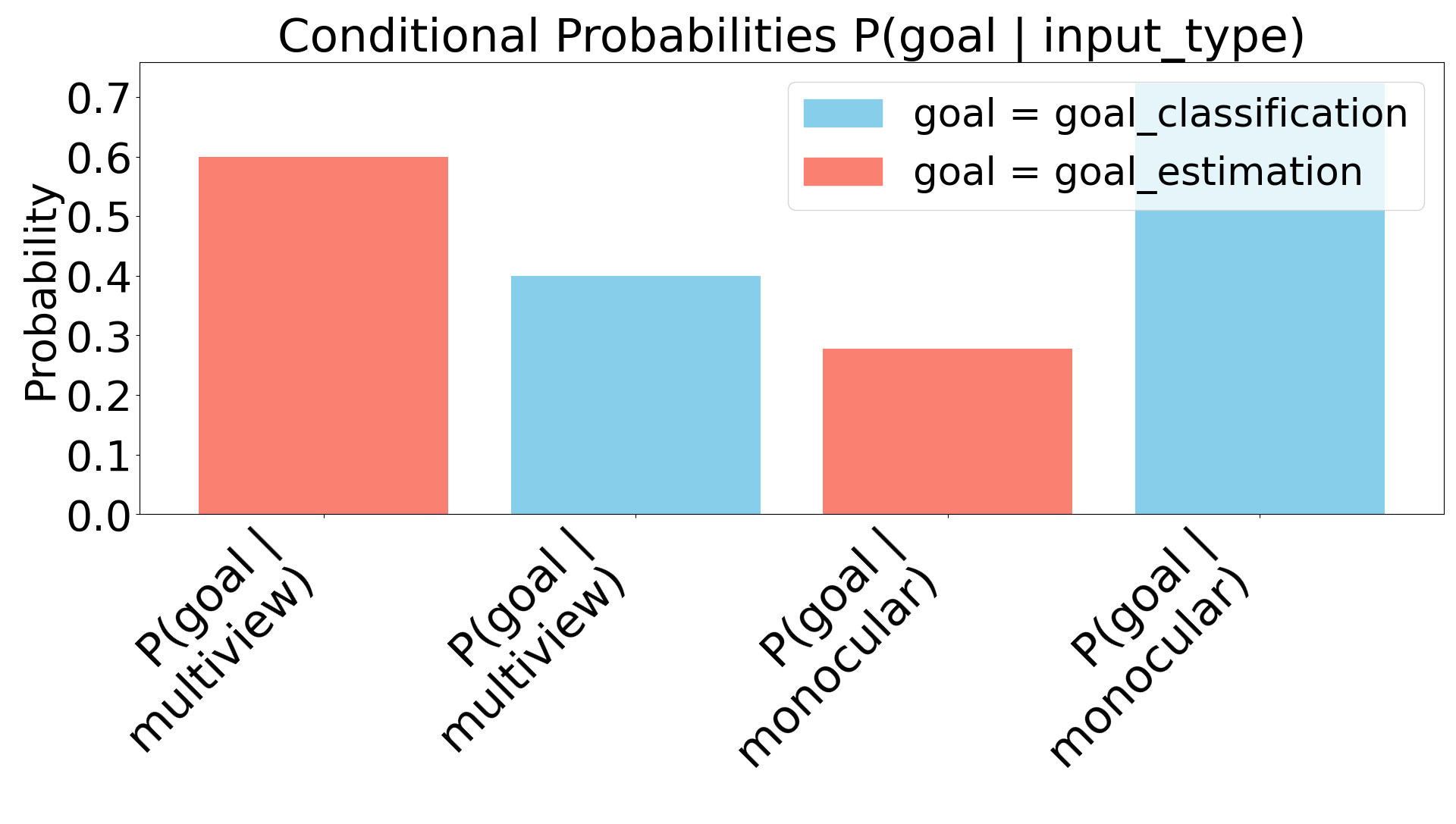}
    \caption{\textcolor{black}{Conditional Probability of goal given number of cameras.}}
    \label{sfig:bayes2}
\end{figure}
\begin{figure}[!htb]
    \centering
    \includegraphics[width=0.9\columnwidth]{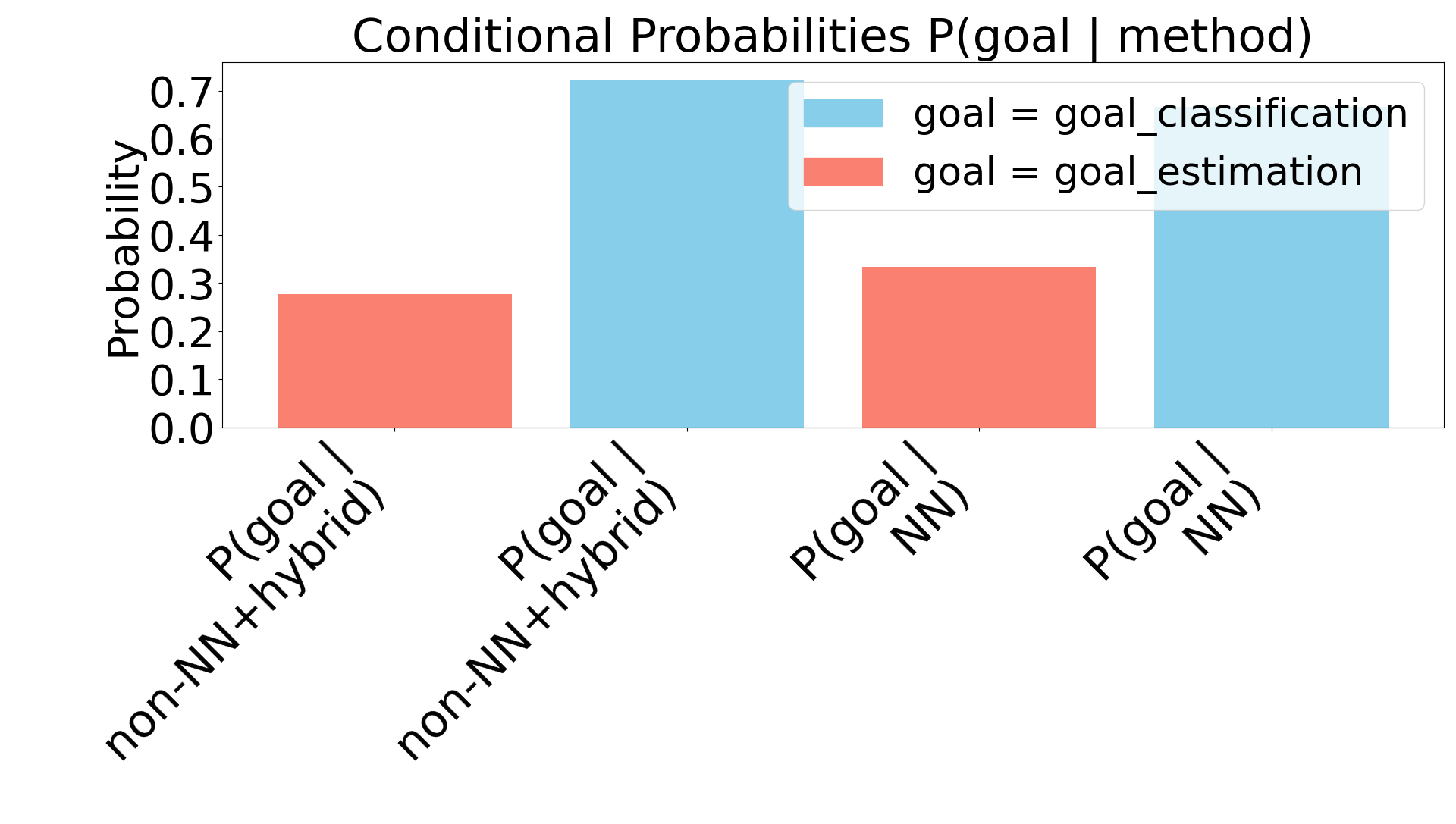}
    \caption{\textcolor{black}{Conditional Probability of goal given implementation method.}}
    \label{sfig:bayes3}
\end{figure}
\begin{figure}[!htb]
    \centering
    \includegraphics[width=0.9\columnwidth]{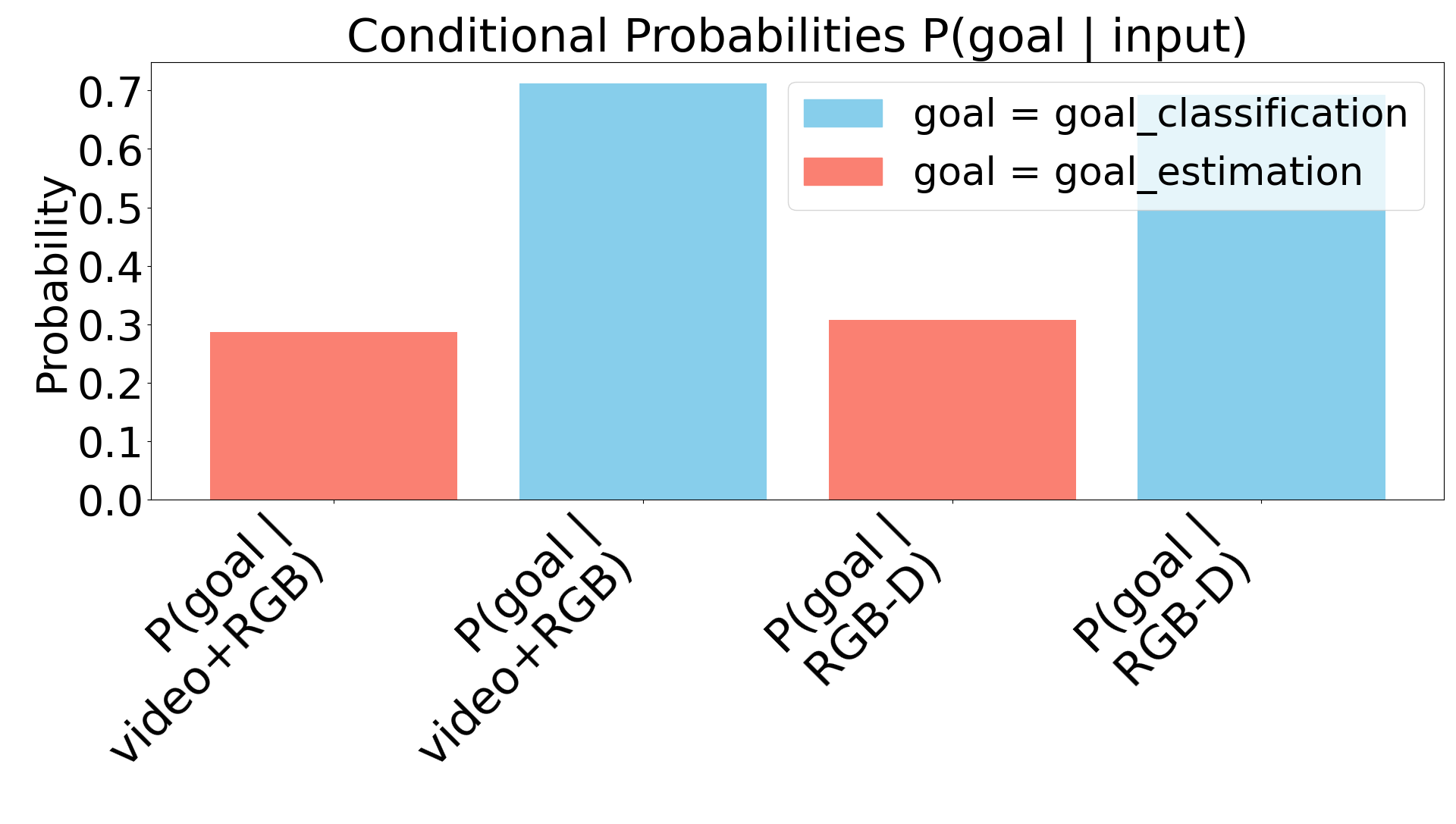}
     \caption{\textcolor{black}{Conditional Probability of goal given input modality.}}
     \label{sfig:bayes4}
\end{figure}

The results are visualized in figures \ref{sfig:bayes1} to \ref{sfig:bayes4}, where the conditional probabilities $P(Y|X)$ are shown for each combination of X and Y. Each subplot corresponds to a specific group, with blue bars representing the classification goal and red bars representing the estimation goal. Like so, the plots provide valuable insights into the relationship between different groups and the probability of achieving classification or estimation goals.

Fig. \ref{sfig:bayes1} clearly shows that when using a box/filter capture methodology, the likelihood of targeting a classification task is almost twice as high as that for an estimation task. In Fig. \ref{sfig:bayes2}, it is clear that multi-view camera setups are more likely to be used in hand estimation tasks than in classification. 

As illustrated in Fig. \ref{sfig:bayes3}, all method types --whether hybrid, non-NN, or NN-- exhibit a higher probability of being utilized for classification tasks rather than estimation tasks. Notably, hybrid and non-NN methods favor classification nearly twice as much as estimation, emphasizing their inclination toward hand gesture classification. Similarly, in the \textit{input} group (Fig. \ref{sfig:bayes4}), all input types (e.g., video, RGB, and RGB-D) demonstrate a consistent association with classification tasks rather than estimation tasks. These observations indicate a general tendency toward classification across the analyzed groups, which aligns with the survey's emphasis on hand gesture classification and reflects the broader trend in research papers that prioritize classification over hand representation

\subsection{Key Takeaways of this section}
\textcolor{black}{This section provided a comprehensive classification of Hand Gesture Recognition methods based on several key factors. The primary ways HGR methods are categorized and their prevalence are:
\begin{itemize}
    \item \textbf{Task Objective:} HGR methods are primarily categorized by their goal: \textit{Hand Gesture Classification} (identifying semantic meaning) or \textit{Hand Gesture Estimation} (representing hand topology/structure).
    \item \textbf{Input Data Type:} Methods utilize different visual inputs including RGB, RGB-D, and Video, with video-based methods being most prevalent.
    \item \textbf{Camera Setup:} \textit{Monocular} camera setups dominate due to their simplicity and cost-effectiveness over \textit{multi-view} setups.
    \item \textbf{Hand Gesture Capture:} Techniques are broadly classified into \textit{Skeleton-Based} captures (representing the hand as key joints and connections) and \textit{Box/Filter-Based} captures (detecting the hand regions).
    \item \textbf{Recognition Techniques:} Predominantly \textit{Hybrid Approaches}, combining strengths of multiple techniques (e.g., CNN+LSTM), followed by standalone Neural Network approaches and fewer Non-Neural Network methods.
    \item \textbf{Observed Associations:} Analysis showed tendencies such as box/filter methods being more frequently used for classification tasks, while multi-view setups are more common for estimation tasks. Overall, classification is the more common goal across most analyzed categories.
\end{itemize}}

\section{Overview of algorithms used in HGR tasks} \label{sec:algorithms}

The variety of techniques employed in HGR reflects the complexity of the task, ranging from spatial modeling to capturing temporal dynamics. This section explores the prominent algorithms utilized in HGR, emphasizing their unique strengths and applications in diverse real-world scenarios.

\subsection{Support Vector Machines (SVMs)}
Support Vector Machines are a robust classification tool in HGR, particularly effective for distinguishing complex hand poses or subtle variations in gesture patterns. 
The works in the field employ different feature extraction techniques and representations to capture information from images. Typical examples are Histogram of Oriented Gradients \cite{huu2021hand}, Hu Moment Invariants extracted by the hand contour \cite{zhang2023human}, or positional and trajectory features of the hand joints.

\begin{figure}[htbp]
    \centering
    \includegraphics[width=\columnwidth]{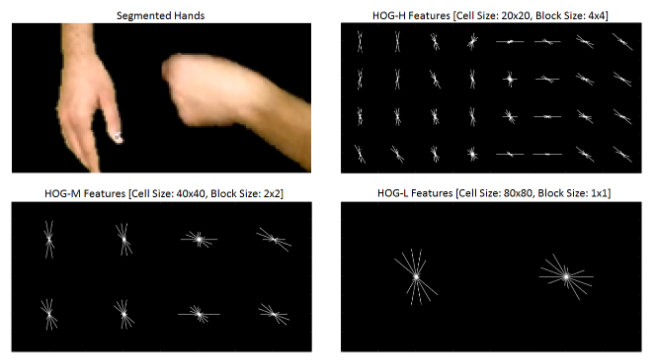}
    \caption{An example of HOG features extracted from a hands' image. Source \cite{camgoz2016sign}.}
    \label{fig:hogfeatures}
\end{figure}

Following the notation described in Section \ref{sec:basics}, from a set of N input images the potential gestures are represented as $\mathcal{X} = \{ \mathbf{x}_1, \mathbf{x}_2, \ldots, \mathbf{x}_N \}$ where each vector $x_i \in \mathbb{R}^d$ is a d-dimensional feature vector extracted using HOG, skeletal tracking, etc. information. The set of corresponding class labels for each image is denoted as $\mathcal{Y} = \{ y_1, y_2, \ldots, y_N \}$ where $y_i \in \{ 1,2,\ldots,C\}$ represents the class of gesture $i$ and $C$ is the total number of distinct gesture classes. The SVM aims to find a hyperplane that separates the classes in the feature space. The SVM optimization problem can be expressed as:
\[
\min_{\mathbf{w}, b} \frac{1}{2} ||\mathbf{w}||^2 + \gamma \sum_{i=1}^{N} \xi_i
\]
subject to:

\[
y_i (\mathbf{w}^T \mathbf{x}_i + b) \geq 1 - \xi_i, \quad i = 1, 2, ..., N
\]

\noindent where
\( \mathbf{w} \) is the weight vector defining the hyperplane,
\( b \) is the bias term,
\( \gamma > 0 \) is a regularization parameter controlling the trade-off between maximizing the margin and minimizing classification error,
\( \xi_i \) are slack variables that allow for misclassification.

Once trained, the SVM model predicts the class label for a new input vector \( \mathbf{x}_{new} \):
\[
f(\mathbf{x}_{new}) = sign(\mathbf{w}^T\phi(\mathbf{x}_{new}) + b)
\]
This function outputs \( f(\mathbf{x}_{new}) = y_j \), where \( j = 1, 2, ..., C\), indicating which class the new gesture belongs to.

The integration of SVMs with handcrafted feature descriptors, such as Local Extrema Min-Max Pattern (LEMMP), has led to remarkable recognition accuracies, reaching up to 99\% on benchmark datasets \cite{BAHUGUNA2024106203}. Kernel-based SVMs have proven successful in dynamic gesture recognition systems, achieving over 98\% accuracy in tasks like Japanese Sign Language classification \cite{s24030826}. Ensemble methods, where SVMs are part of multi-camera systems, further enhance recognition accuracy by fusing results from multiple classifiers \cite{8600529}. These attributes make SVMs a versatile choice for both static and dynamic gesture recognition.

\subsection{Hidden Markov Models (HMMs)}
Hidden Markov Models are widely employed for modeling sequential dependencies in HGR tasks \cite{9984249, koller2019weakly}. These probabilistic models excel at predicting gesture sequences by representing hidden processes through observable outputs. 

The process of gesture recognition using HMMs can be formulated with the following steps:
\begin{itemize}
    \item \textbf{Feature Extraction}: Initially, features are extracted from the video input. Techniques such as skin color segmentation may be employed to isolate the hand from the background, followed by tracking methods to monitor hand movements over time. Preprocessing techniques, such as grayscale conversion and edge detection, facilitate the extraction of key gesture regions for HMM-based sequence learning. 
    \item \textbf{State Representation}: Each gesture is represented by a sequence of hidden states. For example, in a sign language recognition system, each state might correspond to a different phase of a gesture (e.g., starting position, mid-gesture, ending position).
    \item \textbf{Observation Sequence}: Observations are derived from features extracted from video frames that capture relevant aspects of the gesture. Common features include the position of key points (such as fingers and hands) and angles between joints. The continuous observation sequences—such as angles or positions—are quantized into discrete states suitable for HMM modeling. This quantization process involves mapping continuous values into discrete categories that can be effectively handled by the HMM framework.
    \item \textbf{Transition Probabilities}: To model the transitions between these hidden states, transition probabilities are defined. For example, let \( P(s_t = j | s_{t-1} = i) = a_{ij} \) denote the probability of transitioning from state \( i \) at time \( t-1 \) to state \( j \) at time \( t \). This probabilistic framework allows us to account for the inherent variability in how gestures are performed.
    \item \textbf{Emission Probabilities}: In addition to transition probabilities, emission probabilities are also established that describe how observations are generated from hidden states. Each state emits observations according to a probability distribution, often modeled using Gaussian distributions. This means that for a given state \( s_t \), the observation \( O_t \) is generated based on a distribution characterized by parameters specific to that state.
    \item \textbf{Training the HMM}: The parameters of the HMM —namely, transition and emission probabilities— are estimated using algorithms such as the Baum-Welch algorithm. This training phase is crucial for enabling the model to learn from example sequences of gestures.
    \item \textbf{Decoding Observation Sequences}: Once trained, algorithms like Viterbi are used to decode new observation sequences and to determine the most likely sequence of hidden states that corresponds to a given gesture.
\end{itemize}

Hybrid approaches integrating HMMs with CNNs or LSTMs capitalize on their strengths in spatial and temporal feature extraction, achieving state-of-the-art results in datasets with complex and continuous gestures.

\subsection{Convolutional Neural Networks (CNNs)}
Convolutional Neural Networks are often utilized in HGR due to their ability to process visual data and extract meaningful features from images in real-time \cite{alaftekin2024real}. For instance, their implementation in sign language recognition systems, such as those employing YOLOv8, has achieved recognition accuracies of up to 99.4\%, effectively bridging communication gaps for deaf-mute individuals \cite{TALAAT2024109475}. 

In the gesture recognition task using CNNs, each image is first represented as a 3D tensor. Let $\mathcal{I} = \{ I_1, I_2, \ldots, I_N \}$ be the set of input images representing different gestures, where $N$ is the total number of images. Each image $I_i$ can be represented as a 3D tensor $I_i \in \mathbb{R}^{H \times W \times C}$, where $H, W$ are the image height and width, respectively, and $C$ is the number of color channels (e.g., RGB).

The CNN model consists of multiple layers that process the input images to extract features and make predictions. The architecture typically includes:
\begin{itemize}
    \item \textbf{Convolutional Layers}: These layers apply convolution operations to extract spatial features from the input images.
    \item \textbf{Activation Functions}: Non-linear activation functions like ReLU are applied after convolutional layers to introduce non-linearity into the model.
    \item \textbf{Pooling Layers}: These layers reduce the spatial dimensions of feature maps, helping to down-sample and retain important features while reducing computation.
    \item \textbf{Fully Connected Layers}: At the end of the network, fully connected layers are used to classify the gestures based on the extracted features.
\end{itemize}

The images are fed to the CNN in batches during the training and inference phase. The formal definition of gesture recognition on a given image is described in the following.

For a given input image $I_i$, let $F_k(I_i)$ denote the feature map produced by the $k^{th}$ convolutional layer. The feature extraction can be expressed as:
\begin{equation}
F_k(I_i) = f(W_k * I_i + b_k)
\end{equation}

\noindent where $W_k$ is the filter (kernel) applied at layer $k$, $b_k$ is the bias term, $*$ denotes convolution, $f(\cdot)$ is an activation function (e.g., ReLU).

The pooling operation that follows reduces dimensionality:  
\begin{equation}
P(F_k(I_i)) = P(F_k(I_i))_{h,w}
\end{equation}
\noindent where $P(\cdot)$ represents a pooling function (e.g., max pooling or average pooling).

After passing through several convolutional and pooling layers, let $H(I_i)$ be the output from the last fully connected layer before classification:
\begin{equation}
   H(I_i) = W_f^T F_{final}(I_i) + b_f
\end{equation}

The final class prediction for gesture recognition can then be obtained using a softmax function:
\begin{equation}
   P(y | I_i) = softmax(H(I_i))
\end{equation}

The integration of CNNs with diverse datasets enhances their robustness, making them suitable for dynamic environments \cite{app11094164}. Metrics such as precision, recall, and processing time underscore their reliability, positioning CNNs as a foundational tool in the development of HGR systems.

\subsection{Long Short-Term Memory (LSTM) Networks}
When it comes to video input, the need for processing the temporal dimension of gestures raises. 
Long Short-Term Memory networks, a type of Recurrent Neural Network, are adept at capturing temporal dependencies in gesture data, particularly in video sequences \cite{RASTGOO2024123349, 8545095}. 

The LSTM model consists of memory cells that maintain information over long sequences. Each memory cell contains: i) the input gate that controls how much of the new input $x_i$ to let into the memory, ii) the forget gate that decides what information to discard from the memory, and iii) the output gate that determines what part of the memory to output.

The first step of LSTM processing is the calculation of the gates, which comprise the input gate $i_t$, the forget gate $f_t$, and the output gate $o_t$, and are computed as follows:
   
\begin{equation}
   i_t = \sigma(W_i x_t + U_i h_{t-1} + b_i)
\end{equation}

\begin{equation}
   f_t = \sigma(W_f x_t + U_f h_{t-1} + b_f)
\end{equation}
   
\begin{equation}
   o_t = \sigma(W_o x_t + U_o h_{t-1} + b_o)
\end{equation}
   
\noindent where $W_i, W_f, W_o$ are weight matrices for the input, forget, and output gates respectively, 
$U_i, U_f, U_o$ are recurrent weight matrices, $b_i, b_f, b_o$ are bias vectors, $h_{t-1}$ is the hidden state from the previous time step,
and $\sigma(\cdot)$ is the sigmoid activation function.

The next step is the update of the cell state $c_t$ using the following equation:

\begin{equation}
   c_t = f_t * c_{t-1} + i_t * \tanh(W_c x_t + U_c h_{t-1} + b_c)
\end{equation}
\noindent where $W_c, U_c, b_c$ are weights and bias for the candidate values added to the cell state.

Finally, the output of the hidden state $h_t$ for time step $t$ is computed as:
\begin{equation}
   h_t = o_t * \tanh(c_t)
\end{equation}

After processing the entire sequence of frames in a video, a fully connected layer with a softmax activation function can be used to predict the class label as follows:
\begin{equation}
   P(y | x_1, x_2, ..., x_T) = softmax(W_h h_T + b_h)
\end{equation}
\noindent where $W_h$ and $b_h$ are weights and biases for the final classification layer.

When combined with CNNs for feature extraction, LSTMs excel at analyzing sequential frames to recognize complex and continuous gestures \cite{NUNEZ201880}. Features may include joint coordinates from skeletal data or pixel values from video frames. For example, in skeleton-based approaches, 3D coordinates of key joints (e.g., wrists, elbows) are often used. Gestures may be segmented into fixed-length sequences before being fed into the LSTM. This segmentation can help capture important temporal dynamics and dependencies.
The ability of LSTMs to retain context across time steps enables nuanced gesture understanding, making them ideal for real-time applications. 

Enhanced with attention mechanisms, LSTMs can identify critical spatial and temporal regions, further improving accuracy and interpretability in user-centric environments. This is particularly useful in gesture recognition tasks, where certain frames or features may carry more significant information than others \cite{zhang2018attention, liu2021dynamic}. 
The attention mechanism can be integrated into the LSTM framework by computing the attention score $e_{t,i}$ for input $i$ at time $t$ as follows:

\begin{equation}
    e_{t,i} = v_a^T tanh(W_ah_t+U_ah_i)
\end{equation}
\noindent where $v_a$ is a learned weight vector and $W_a$ and $U_a$ are weight matrices.

The attention weights can be computed using softmax: $a_{t,i}=\frac{e_{t,i}}{\sum_j{e_{t,j}}}$, whereas the context vector $c_t$ is computed as a weighted sum of the hidden states $c_t=\sum_i{a_{t,i}h_i}$.

The final output of the LSTM with attention can be expressed as: 
\begin{equation}
y_t = W_yh_t+V_yc_t+b_y
\end{equation}
\noindent where $W_y, V_y$ and $b_y$ are parameters for the output layer.

\subsection{Graph Convolutional Networks (GCNs)}
Graph Convolutional Networks offer a unique approach to HGR by effectively modeling the spatial and structural relationships inherent in gesture data \cite{10049842, 10452001}. Unlike CNNs, which operate on grid-like data, GCNs excel with non-Euclidean structures such as graphs, capturing complex dependencies like joint relationships in skeletal data or superpixel connections in images. 

Hand gestures are often represented as graphs of skeletal joints. Each joint corresponds to a node in the graph, and edges represent physical connections between joints. For example, a hand gesture can be represented with nodes for each finger joint and edges connecting them based on anatomical structure. Features associated with each joint may include spatial coordinates, velocity, acceleration, or angles between adjacent joints. This multi-dimensional feature representation allows GCNs to learn complex patterns within gesture data. For dynamic gestures, sequences of graphs can be constructed over time. Each frame of a video can be represented as a graph, and GCNs can process these sequences to capture temporal dependencies.

The formulation of the gesture recognition task using GCNs begins with the representation of the potential gesture as a graph. $\mathcal{G} = (V, E)$ where $V = \{ v_1, v_2, \ldots, v_N \}$ is the set of nodes corresponding to key joints of the human body or hand, and $E$ is the set of edges representing the connections between these joints. Each node $v_i$ can be associated with a feature vector $x_i \in \mathbb{R}^d$, where $d$ is the dimensionality of the features (e.g., joint coordinates).

The relationships between nodes are then described using an adjacency matrix $A$, where $A_{ij} = 1$ if there is an edge between nodes $v_i$ and $v_j$ and $0$ otherwise.

The GCN processes the input graph through multiple layers to learn node representations. The output for each node can be expressed as:
\begin{equation}
   H^{(l+1)} = \sigma\left( A H^{(l)} W^{(l)} \right) 
\end{equation} 
\noindent where $H^{(l)}$ is the matrix of node features at layer $l$, $W^{(l)}$ is the weight matrix for layer $l$,
$A$ is the adjacency matrix, $\sigma(\cdot)$ is an activation function (e.g., ReLU).

After processing through several GCN layers, the final output can be computed as:
\begin{equation}
   Y_{pred} = softmax(H^{(L)} W^{(L)})
\end{equation} 
\noindent where $H^{(L)}$ is the output from the last GCN layer, and $W^{(L)}$ is the weight matrix for classification.

Innovations such as hybrid architectures combining GCNs with attention mechanisms have further enhanced their capabilities \cite{BAMANI2024108443}. These systems leverage both local and global contextual information, ensuring precise recognition across challenging datasets, including multicultural sign language translation \cite{10452793}.

\subsection{Transformer Models}
The introduction of transformer-based models has significantly advanced HGR, offering superior performance in tasks requiring the capture of long-range dependencies and intricate spatial-temporal relationships \cite{10484197, 10007834, 10233392}. 
\textcolor{black}{
Transformer-based architectures have emerged as powerful tools since they can address both static and dynamic gesture tasks. For instance, GestFormer \cite{garg2024gestformer} introduced multiscale wavelet pooling to better model spatiotemporal variations in dynamic gestures, addressing challenges like hand orientation and trajectory changes. Meanwhile, HGR-ViT \cite{tan2023hgr} adapted Vision Transformers (ViTs) for static gesture recognition, achieving robust performance on ASL and NUS datasets by encoding hand orientation and positional features. Recent advancements also emphasize efficiency: ConvMixFormer \cite{garg2024convmixformer} reduced parameter counts while maintaining generalization for dynamic gestures. 
}

Since Transformers do not inherently capture the order of sequences, positional encodings $PE(t)$ are added to the input feature vector $x_t \in \mathbb{R}^d$ of each frame thus forming the final representation for time step $t$ as $z_t = x_t + PE(t)$.

The self-attention mechanism allows the model to weigh the importance of different time steps when making predictions. The attention scores are calculated as:
\begin{equation}
   A_{ij} = \frac{\exp\left(\frac{Q_i K_j^T}{\sqrt{d_k}}\right)}{\sum_{k=1}^{T} \exp\left(\frac{Q_i K_k^T}{\sqrt{d_k}}\right)} 
\end{equation}
\noindent  where $Q$ and $K$ are the query and key matrices respectively and $d_k$ is the dimensionality of the keys.

Consequently, the output representation for each time step can be computed as:
\begin{equation}
   O_i = \sum_{j=1}^{T} A_{ij} V_j
\end{equation}
\noindent where $V$ is the value matrix corresponding to each input feature.

To generate $Q$, $K$, and $V$ (for values) representations from $z_t$ we apply learned linear transformations using weight matrices $W_Q. W_K. W_V$ so that: $Q_t=W_Qz_t$, $K_t=W_Kz_t$, $V_t=W_Vz_t$.

After processing through multiple layers of self-attention and feed-forward networks, a final classification layer is applied:
\begin{equation}
   P(y | x_1, x_2, ..., x_T) = softmax(W_O O + b_O)
\end{equation}
\noindent where $O$ is the output from the last transformer layer, $W_O$ and $b_O$ are weights and biases for the output layer.

\textcolor{black}{
The primary advantage of transformers in HGR lies in their exceptional ability to model long-range dependencies and to process sequences in parallel, leading to potentially faster training and a more holistic understanding of gestures compared to sequential models like LSTMs. They also offer flexibility for transfer learning and multimodal fusion. However, transformers are not without drawbacks. Their standard self-attention mechanism exhibits computational complexity quadratic to the input sequence length ($O(N^2)$), posing challenges for very long video sequences or high-resolution inputs. This can impact inference speed, particularly on resource-constrained devices. Furthermore, transformers typically require substantial amounts of training data, due to their weaker inductive biases compared to CNNs, and can be harder to interpret, despite the insights offered by attention maps. The necessity for positional encodings to manage their permutation-equivariant nature also adds a layer of design consideration.
}

\textcolor{black}{
When compared to established CNN+LSTM stacks, transformers offer a different paradigm for temporal modeling. While LSTMs inherently capture temporal order and excel at short-to-medium range dependencies, they can struggle with very long sequences. Transformers, in contrast, are theoretically better suited for such long-range interactions due to their global attention, but their $O(N^2)$ complexity can be more demanding than the LSTMs' $O(N)$ per-step one for extremely long sequences. CNNs remain highly effective for initial spatial feature extraction due to strong inductive biases, and many HGR transformers leverage CNN backbones. Thus, transformers might be less data-efficient than CNN+LSTM approaches on smaller datasets.
}

Innovations like hand-level tokenization reduce computational complexity while maintaining high accuracy. Multi-task transformer networks, designed for simultaneous gesture recognition and pose estimation, have further optimized performance through collaborative learning. These models excel in scenarios involving hand-object interactions or continuous sign language translation, achieving competitive results on benchmark datasets. Their ability to encode contextual features with precision highlights their transformative impact on HGR. \textcolor{black}{These developments align with findings from Hashi et al. \cite{hashi2024systematic}, whose systematic review highlights transformers’ growing dominance in vision-based HGR since 2018, citing improvements in accuracy and computational efficiency. Together, these works illustrate how transformers address diverse HGR challenges, from temporal modeling to cross-dataset generalization.}

\subsection{Key Takeaways of this section}
\textcolor{black}{
This section explored the prominent algorithms utilized in HGR, highlighting their unique strengths in modeling spatial and temporal aspects of hand gestures. The main algorithms discussed include:
\begin{itemize}
    \item \textbf{Support Vector Machines (SVMs):} Remain effective for gesture classification, often paired with handcrafted features like HOG or LEMMP, and can be extended with kernel methods for dynamic gestures.
    \item \textbf{Hidden Markov Models (HMMs):} Widely employed for modeling the sequential nature of dynamic gestures, particularly in sign language recognition, by representing gestures as sequences of hidden states.
    \item \textbf{Convolutional Neural Networks (CNNs):} A cornerstone in HGR for their ability to automatically learn and extract hierarchical spatial features directly from visual data (images/video frames).
    \item \textbf{Long Short-Term Memory (LSTM) Networks:} A type of Recurrent Neural Network (RNN) excelling at capturing temporal dependencies in sequential gesture data, frequently combined with CNNs for spatio-temporal modeling.
    \item \textbf{Graph Convolutional Networks (GCNs):} Suited for HGR tasks where data can be represented as a graph (e.g., hand skeletons), allowing for the modeling of non-Euclidean structural relationships between joints.
    \item \textbf{Transformer Models:} Have shown significant advancements in HGR, effectively capturing long-range dependencies and complex spatio-temporal patterns in both static and dynamic gestures, leading to state-of-the-art performance.
\end{itemize}
}

\section{Datasets and metrics} \label{sec:datasets}

\textcolor{black}{This section presents the main datasets used as benchmarks in HGR tasks and gives an overview of the main evaluation metrics. It} categorizes datasets based on various criteria, including input type (e.g., RGB, RGB-D, video), number of samples, resolution, number of classes (for classification datasets), and gesture type (dynamic or static).  Additionally, the categorization incorporates the primary hand gesture recognition tasks discussed in Section \ref{subsec:hand_gest_reco}. Since the same dataset can be used for multiple tasks and can be employed in multiple research works, we determine the primary task associated with each dataset, by considering the task that it is applied to in the majority of articles that use the dataset. For instance, if most articles that use a dataset, use it for Hand Gesture Classification and only a few of them use it for Sign Language Recognition, the dataset is categorized under Hand Gesture Classification.

The datasets used for classification tasks such as Hand Gesture Classification and Sign Language Recognition are presented in Table \ref{tab:datasets_classif}. In a similar manner, Table \ref{tab:datasets_estim} displays the datasets used for estimation tasks, including Hand Gesture Estimation and Hand/Body Reconstruction.

\begin{table*}[!htb]
\centering
\caption{Overview of datasets for hand gesture classification and sign language recognition.}
\label{tab:datasets_classif}
\begin{adjustbox}{width=\textwidth}
\begin{tabular}{lrllllll}
\specialrule{1.5pt}{0pt}{0pt} 
\textbf{Topic} & \textbf{Type of Gesture} & \textbf{Input} & \textbf{Dataset} & \textbf{Year} &   \textbf{\# Samples} & \textbf{Resolution} & \textbf{\# Classes}  \\
\specialrule{1.5pt}{0pt}{0pt} 
\multirow{30}{*}{\parbox{1.5cm}{\textbf{Hand Gesture \\ Classification}}} & \multirow{10}{*}{Static} & \multirow{6}{*}{RGB} & MUGD \cite{barczak2011new} & 2011 & 2524 images & Not specified & 36 \\
 &  &  & NUS II \cite{pisharady2013attention} & 2013 & 2000 images & 120x160 pixels & 10 \\
 &  &  & TSL-FS \cite{SADEGHZADEH2024200384} & 2018 & 2974 images & Not specified & 29 \\
 &  &  & NUS-I \cite{BAHUGUNA2024106203} & 2024 & 240 images & 160x120 pixels & 10 \\
 &  &  & Hagrid-14 \cite{BAHUGUNA2024106203} & 2024 & Not specified & 256x256 pixels & 14 \\ \cline{3-8}
 &  & \multirow{4}{*}{RGB+RGB-D} & \makecell[l]{Microsoft Kinect and Leap Motion \\ Dataset \cite{marin2014hand}} & 2013 & 1400 samples & Not specified & 10 \\
 &  &  & OU Hand \cite{matilainen2016ouhands} & 2022 & 3000 images & 640x480 pixels & 10 \\
 &  &  & 3MHand \cite{9903078} & 2023 & 20000 frames & Not specified & 7 \\\cline{3-8}
 &  & video & RWTH \cite{dreuw2006modeling} & 2006 & 1400 samples & Not specified & 35 \\\cline{2-8}
 & \multirow{22}{*}{Dynamic} & \multirow{2}{*}{RGB} & GRIT Robot Commands \cite{TSIRONI201776} & 2017 & 543 videos & 640x480 pixels & 9 \\
 &  & & JESTER \cite{Materzynska_2019_ICCV} & 2019 & 148092 videos & 100x100 pixels & 27 \\\cline{3-8}
 &  & \multirow{4}{*}{RGB+RGB-D} & NVIDIA Dynamic Hand Gesture \cite{molchanov2016online} & 2016 & 1532 videos & Not specified & 25 \\
 &  & & EgoGesture \cite{zhang2018egogesture} & 2018 & 24161 videos & 640x480 pixels & 83 \\
 &  & & Briareo \cite{manganaro2019hand} & 2019 & 120 frames & Not specified & 12 \\ 
 &  & & H2O \cite{kwon2021h2o} & 2024 & 571k frames & Not specified & 36 \\\cline{3-8}
 &  & RGB+RGB-D+Skeleton & SHREC’17 \cite{de2017shrec} & 2017 & 2800 sequences & 640x480 pixels & 28 \\
 &  & RGB+RGB-D+Skeleton & UTKinect-Action3D \cite{xia2012view} & 2018 & 6220 frames & 320x240 pixels & 10 \\\cline{3-8}
 &  & \makecell[l]{RGB+RGB-D \\ +Skeleton+infrared} & NTU RGB+D \cite{shahroudy2016ntu} & 2016 & 56880 samples & Not specified & 60 \\\cline{3-8}
 &  & RGB-D & Montalbano \cite{PigouODHD15} & 2014 & 13858 videos & 640x480 pixels & 20 \\\cline{3-8}
 &  & RGB-D+video+Skeleton & MSRDailyActivity3D \cite{wang2012mining} & 2012 & 320 sequences & Not specified & 16 \\\cline{3-8}
 &  & \multirow{2}{*}{RGB-D+video} & DHG-14/28 \cite{de2016skeleton} & 2016 & 2800 frames & 640x480 pixels & 28 \\
 &  &  & First-Person Hand Action \cite{chao2021dexycb} & 2018 & 100000 frames & Not specified & 45 \\\cline{3-8}
 &  & \multirow{8}{*}{video} & ASLVID \cite{athitsos2008american} & 2008 & 3800 signs & 640x480 & 3000 \\
 &  &  & MSRA-3D  \cite{li2010action} & 2010 & 557 sequences & Not specified & 20 \\
 &  &  & HMDB-51 \cite{kuehne2011hmdb} & 2011 & 6766 videos & 240x240 pixels & 51 \\
 &  &  & UCF-101 \cite{Jiang_2019_ICCV} & 2011 & 13320 videos & Not specified & 101 \\
 &  &  & ChaLearn LAP IsoGD \cite{wan2016chalearn} & 2016 & 47933 videos & Not specified & 249 \\
 &  &  & Autism spectrum disorder dataset \cite{8545095} & 2018 & 1837 videos & 1280×720 pixels & 4 \\
 &  &  & EPIC Kitchens \cite{damen2018scaling} & 2018 & 11.5M frames & 1920x1080 pixels & 149 \\
 &  &  & RKS-PERSIANSIGN \cite{rastgoo2020hand} & 2020 & 10000 videos & Not specified & 100 \\\hline
\multirow{21}{*}{\parbox{1.5cm}{\textbf{Sign Language\\ Recognition}}} & \multirow{7}{*}{Static} & \multirow{6}{*}{RGB} & American Sign Language Digits \cite{barczak2011new} & 2011 & 2245 images & Not specified & 36 \\
 &  &  & Nigerian Sign Language \cite{kang2015real} & 2015 & 25900 images & Not specified & 37 \\
 &  &  & BdSL \cite{hoque2020bdsl36} & 2020 & 26713 images & 300x500 pixels & 36 \\
 &  &  & PSL \cite{imran2021dataset} & 2021 & 1480 images & 480x640 pixels & 37 \\
 &  &  & Japanese Sign Language \cite{10452793} & 2021 & 7380 images & 400x400 pixels & 41 \\
 &  &  & Arabic Alphabet Sign Language \cite{al2023rgb} & 2023 & 7534 images & Not specified & 31 \\\cline{3-8}
 &  & RGB+RGB-D & American Sign Language \cite{pugeault2011spelling} & 2011 & 48000 samples & 128x128 pixels & 24 \\\cline{2-8}
 & \multirow{14}{*}{Dynamic} & \multirow{8}{*}{RGB} & LSA64 \cite{Ronchetti2016} & 2016 & 3200 videos & 1920x1080 pixels & 64 \\
 &  &  & Korean Sign Language \cite{yang2019korean} & 2019 & 1540 videos & Not specified & 77 \\
 &  &  & British Sign Language \cite{ABDULLAHI2024123258} & 2020 & 16497 frames & Not specified & 18 \\
 &  &  & BosphorusSign22k \cite{ozdemir2020bosphorussign22k} & 2020 & 22542 videos & Not specified & 744 \\
 &  &  & WLASL \cite{9093512} & 2020 & 34404 videos & Not specified & 2000 \\
 &  &  & Indian Sign Language \cite{sridhar2020include} & 2020 & 4287 videos & Not specified & 263 \\
 &  &  & MINDS-Libras (BRASL) \cite{rezende2021development} & 2021 & 1200 videos & 1080x1920 pixels & 20 \\
 &  &  & Chinese sign language \cite{https://doi.org/10.1049/cvi2.12240} & 2024 & 25000 videos & Not specified & 178 \\\cline{3-8}
 &  & RGB-D & Greek Sign Language \cite{ABDULLAHI2024123258} & 2012 & 4910 videos & Not specified & 20 \\\cline{3-8}
 &  & RGB-D+RGB+Skeleton & AUTSL \cite{sincan2020autsl} & 2020 & 38336 videos & 512x512 pixels & 226 \\\cline{3-8}
 &  & \multirow{4}{*}{video} & PHOENIX-2014 \cite{app14198937} & 2014 & 6841 videos & Not specified & 1295 \\
 &  &  & Microsoft American Sign Language \cite{joze2018ms} & 2018 & 25000 videos & Not specified & 1000 \\
 &  &  & RWTH-PHOENIX14T \cite{camgoz2018neural} & 2018 & 0.95 million frames & Not specified & 3953 \\
 &  &  & ALLVD \cite{de2019spatial} & 2019 & 7798 videos & Not specified & 2745 \\

\specialrule{1.5pt}{0pt}{0pt} 
\end{tabular} 
\end{adjustbox}
\end{table*}

\begin{table*}[!ht]
\centering
\caption{Overview of datasets for hand gesture estimation and hand/body reconstruction.}
\label{tab:datasets_estim}
\begin{adjustbox}{width=\textwidth}
\begin{tabular}{lrlllll}
\specialrule{1.5pt}{0pt}{0pt} 
\textbf{Topic} & \textbf{Type of Gesture} & \textbf{Input} & \textbf{Dataset} & \textbf{Year} & \textbf{\# Samples} & \textbf{Resolution} \\
\specialrule{1.5pt}{0pt}{0pt} 
\multirow{26}{*}{\textbf{Hand Gesture Estimation}} & \multirow{6}{*}{Static} & RGB & MPII+NZSL \cite{Iqbal_2018_ECCV} & 2018 & 2800 images & Not specified \\
 &  & RGB & OneHand10K \cite{wang2018mask} & 2019 & 11703 images & Not specified \\
 &  & RGB & FreiHAND \cite{Zimmermann_2019_ICCV} & 2019 & 37k frames & 224x224 pixels \\\cline{3-7}
 &  & RGB+RGB-D & Stereo Hand \cite{zhang20163d} & 2016 & 18000 RGB+18000 RGB-D & Not specified \\
 &  & RGB+RGB-D & ObMan \cite{AVOLA2022108762} & 2018 & 141550 images & 256x256 pixels \\\cline{3-7}
 &  & RGB-D & ICVL \cite{tang2014latent} & 2014 & 180K images & Not specified \\\cline{2-7}
 & \multirow{20}{*}{Dynamic} & \multirow{4}{*}{RGB} & RHP \cite{zimmermann2017learning} & 2017 & 43986 images & 320x320 pixels \\
 &  &  & BiliHands \cite{10506333} & 2024 & 11204 images & Not specified \\
 &  &  & GANerated \cite{mueller2018ganerated} & 2018 & 331k frames & 256x256 pixels \\
 &  &  & InterHand2.6M \cite{moon2020interhand2} & 2020 & 2590k frames & 512x334 pixels \\\cline{3-7}
 &  & \multirow{3}{*}{RGB+RGB-D} & HIC \cite{tzionas2016capturing} & 2016 & 29 sequences & 640x480 pixels \\
 &  & & B2RGB \cite{panteleris2017back} & 2017 & 1888 frames & Not specified \\
 &  & & SHOWMe \cite{SWAMY2024104073} & 2023 & 87540 frames & Not specified \\\cline{3-7}
 &  & \multirow{6}{*}{RGB-D} & Dexter+Object \cite{sridhar2016real} & 2016 & 3000 frames & 640x480 pixels \\
 &  & & EgoDexter \cite{mueller2017real} & 2017 & 3k frames & 640x480 pixels \\
 &  & & Hands2017 \cite{yuan20172017} & 2017 & 1.2 million frames & 640x480 pixels \\
 &  & & SynthHands \cite{mueller2017real} & 2017 & 220k frames & 640x480 pixels \\
 &  & & HO-3D \cite{hampali2020honnotate} & 2020 & 78k frames & 640x480 pixels \\
 &  & & DexYCB \cite{chao2021dexycb} & 2021 & 582k frames & 640x480 pixels \\\cline{3-7}
 &  & \multirow{7}{*}{RGB-D+video} & FORTH \cite{oikonomidis2011efficient} & 2011 & 7148 frames & Not specified \\
 &  &  & ASTAR \cite{xu2013efficient} & 2013 & 435 frames & Not specified \\
 &  &  & Dexter \cite{sridhar2013interactive} & 2013 & 3157 frames & Not specified \\
 &  &  & UCI-EGO \cite{rogez20153d} & 2014 & 3640 frames & Not specified \\
 &  &  & NYU \cite{tompson2014real} & 2014 & 8252 frames & Not specified \\
 &  &  & MSRA \cite{qian2014realtime} & 2014 & 2400 frames & Not specified \\
 &  &  & ICL \cite{tang2014latent} & 2014 & 1599 & Not specified \\\hline
\multirow{12}{*}{\textbf{Hand/Body Reconstruction}} & \multirow{4}{*}{Static} & \multirow{4}{*}{RGB} & LSP \cite{johnson2011learning} & 2010 & 2000 images & Not specified \\
 &  &  & COCO \cite{lin2014microsoft} & 2014 & 328000 images & Not specified \\
 &  &  & Expressive hands and faces \cite{Pavlakos_2019_CVPR} & 2019 & 100 images & Not specified \\
 &  &  & LSP-Extended \cite{johnson2011learning} & 2019 & 10000 images & Not specified \\\cline{2-7}
 & \multirow{8}{*}{Dynamic} & \multirow{3}{*}{RGB} & MPI-INF-3DHP \cite{mehta2017monocular} & 2017 & 1.3 million frames & Not specified \\
 &  &  & Kinetics-400 \cite{carreira2017quo} & 2017 & 400 videos & Not specified \\
 &  &  & InstaVariety \cite{kanazawa2019learning} & 2019 & 2.1 million frames & Not specified \\\cline{3-7}
 &  & RGB+3D & Human3.6M \cite{ionescu2013human3} & 2013 & 3.6 million images & 50Hz video \\\cline{3-7}
 &  & RGB+Skeleton & 3DPW \cite{von2018recovering} & 2018 & 60 videos & Not specified \\\cline{3-7}
 &  & RGB+RGB-D+video & CMU \cite{joo2015panoptic} & 2015 & 297000 frames & Not specified \\\cline{3-7}
 &  & \multirow{8}{*}{video} & PennAction \cite{zhang2013actemes} & 2013 & 2326 videos & Not specified \\
 &  &  & PoseTrack \cite{andriluka2018posetrack} & 2018 & 1337 videos & Not specified \\
\specialrule{1.5pt}{0pt}{0pt} 

\end{tabular} 
\end{adjustbox}
\end{table*}

\subsection{Dataset statistics}

It is worth noting that some datasets referenced in the selected papers predate 2018 or even 2015, despite our focus on papers published after 2018. This discrepancy can be explained by the fact that these older datasets were created before the publication of the collected papers, establishing themselves as foundational or benchmark datasets in the field of hand gesture recognition. They are frequently reused and referenced due to their relevance and reliability in providing consistent and comparable results across studies. For instance, datasets from 2011 like the American Sign Language dataset (ASL) are often cited because they played an important role in shaping the direction of research and remain widely applicable in obtaining baseline results.

\begin{figure}[htbp]
\centering
\includegraphics[width=\columnwidth]{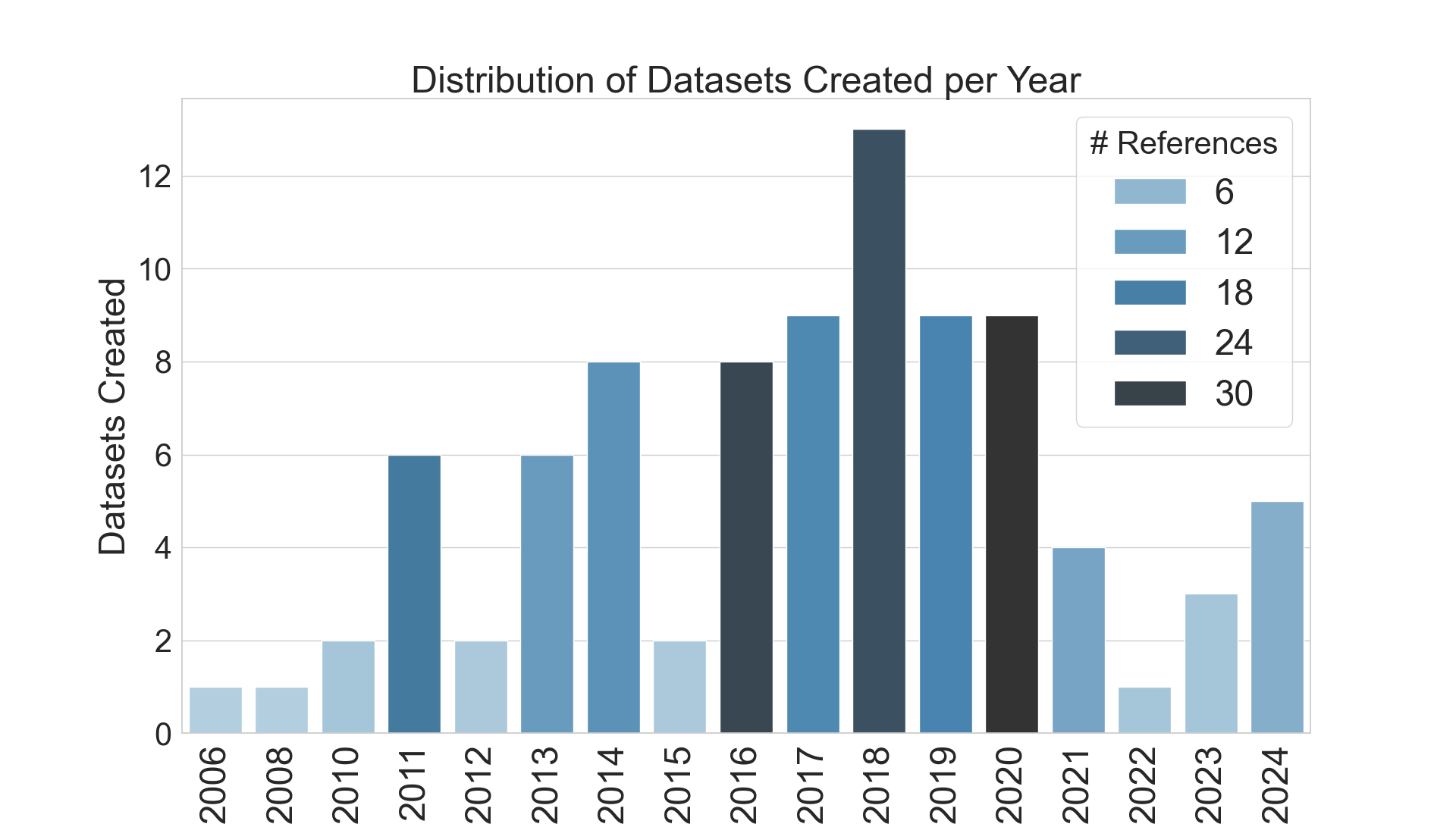}
\caption{Dataset Distribution.}
\label{fig:dataset_distr}
\end{figure}

\begin{figure}[htbp]
\centering
\includegraphics[width=\columnwidth]{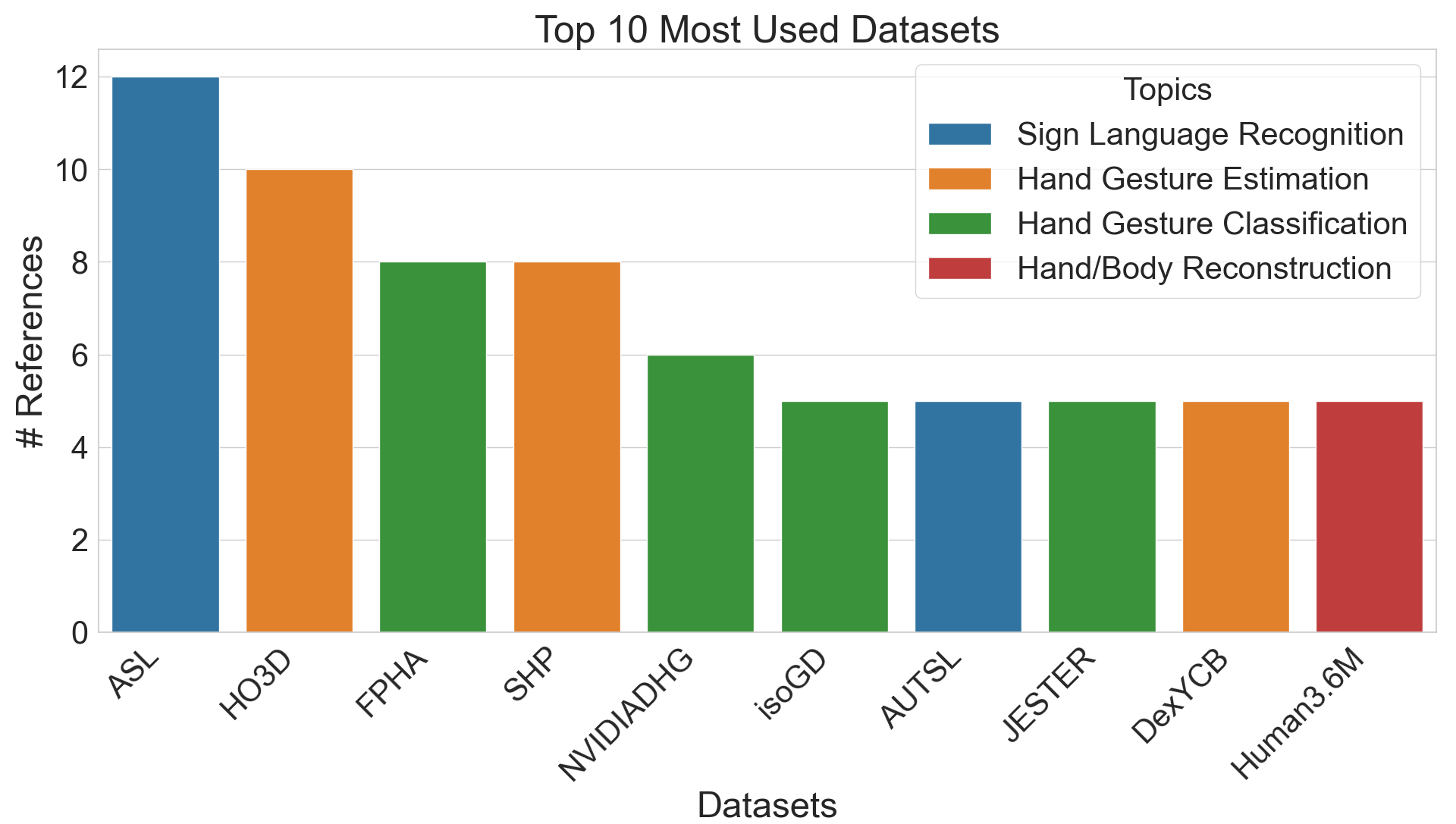}
\caption{Analysis of Most Utilized Datasets.}
\label{fig:top_dataset}
\end{figure}

Figures \ref{fig:dataset_distr} and \ref{fig:top_dataset} illustrate the distribution of datasets analyzed in this study. Specifically, Fig. \ref{fig:dataset_distr} highlights that most datasets were created in 2018. The darker hues in the plot indicate that the most frequently used datasets were created in 2020, followed by those from 2016 and 2018. Overall, datasets created between 2016 and 2020 are the most utilized, likely because they strike a balance between recency and maturity. Older datasets (e.g., 2008–2014) may be less relevant due to outdated methods (e.g., lower resolution or fewer samples), while more recent datasets (2021 and later) may not yet have been widely adopted or refined.

Fig. \ref{fig:top_dataset} presents the most frequently used datasets, categorized by their respective tasks. The dataset with the highest number of references is ASL, which has been employed in 12 articles. ASL is widely recognized in the field of sign language research, and the number of works that use it proves the interest of researchers in benchmark datasets for sign language recognition and consequently for the broader task of hand gesture classification. The HO3D dataset that follows has been employed in 10 articles for hand gesture estimation, establishing it as a key resource for this task.

\subsection{Analysis of Dataset Findings}

The datasets for hand gesture classification and sign language recognition vary significantly in terms of size, resolution, and complexity. Dynamic gesture datasets dominate, reflecting the complexity of real-world gestures, with notable examples such as the EPIC Kitchens dataset featuring over 11.5 million frames at high resolution and the Word-Level American Sign Language dataset offering 34,404 videos with 2,000 classes. Static datasets, like NUS II and Japanese Sign Language, are smaller but still crucial for specific applications. Sign language recognition datasets often include more detailed vocabulary, such as RWTH-PHOENIX14T with 3,953 classes and Microsoft American Sign Language with 1,000 classes, showcasing the focus on linguistic richness. The use of RGB-D data and skeleton features in several datasets, like NTU RGB+D and AUTSL, highlights the importance of multimodal inputs for enhancing classification performance.

The datasets for hand gesture estimation and hand/body reconstruction present a wide variety of input types, resolutions, and sample sizes, offering rich resources for advancing research in these domains. For hand gesture estimation, datasets such as InterHand2.6M (2.59 million frames, RGB) and Hands2017 (1.2 million frames, RGB-D) stand out due to their extensive size, making them ideal for training deep learning models. Many datasets, like SynthHands (220k frames) and GANerated (331k frames), use synthetic data, highlighting the role of simulated environments in complementing real-world data.

For hand/body reconstruction, datasets like Human3.6M (3.6 million frames, RGB+3D) and MPI-INF-3DHP (1.3 million frames, RGB) provide large-scale annotations that are crucial for 3D pose estimation. The inclusion of static datasets, such as LSP and Expressive Hands and Faces Dataset, enriches the diversity by supporting precise annotation of hand and body details. Dynamic datasets, such as PoseTrack and PennAction, focus on video-based reconstruction, facilitating temporal modeling for human motion analysis. The breadth of these datasets underscores their pivotal role in developing robust, generalizable models across hand gesture estimation and body reconstruction tasks.

\subsection{Evaluation Metrics in HGR} \label{subsec:eval_metrics}
\textcolor{black}{
Hand gesture recognition (HGR) systems are evaluated using a combination of task-specific and general-purpose metrics. For classification tasks, (classification) \textbf{accuracy} remains the most widely reported metric for static gestures, defined as the ratio of correctly classified gestures to the total number of samples:
}

\begin{equation}
Accuracy = \frac{TP + TN}{TP + TN + FP + FN}  
\end{equation}
\noindent \textcolor{black}{where $TP$ (true positives) and $TN$ (true negatives) represent correct classifications, while $FP$ (false positives) and $FN$ (false negatives) denote errors. While intuitive, accuracy can be misleading in imbalanced datasets (e.g., rare gestures in sign language). Also, \textbf{temporal alignment metrics} such as the \textbf{Levenshtein distance} (edit distance) are often used to measure sequence similarity. The Edit Distance counts the minimum required number of operations (insertions, deletions, substitutions) to match predicted and ground-truth sequences.
This metric is critical for continuous gesture recognition but computationally expensive.}

\textcolor{black}{Task-specific metrics vary depending on the task and application. For example, real-time HGR systems prioritize \textbf{latency} and \textbf{computational efficiency} and employ \textbf{Frame processing rate} (FPS) to measure real-time viability.}

\begin{equation}
FPS = \frac{\text{Number of processed frames}}{\text{Processing time (seconds)}}
\end{equation}

\textcolor{black}{For temporal models, \textbf{mean per-joint position error (MPJPE)} evaluates 3D hand pose estimation:}

\begin{equation}
\text{MPJPE} = \frac{1}{N} \sum_{i=1}^{N} \| \mathbf{P}_i - \mathbf{\hat{P}}_i \|_2  
\end{equation}

\noindent \textcolor{black}{where $\mathbf{P}_i$ and $\mathbf{\hat{P}}_i$ are ground-truth and predicted joint positions. While MPJPE is precise, it requires expensive motion capture data. \textbf{Intersection-over-Union (IoU)} is popular for spatial segmentation tasks but struggles with fine-grained finger movements.}

\textcolor{black}{A limitation of most commonly used metrics is that they fail to holistically capture real-world usability. For example, high accuracy on benchmark datasets does not guarantee robustness to occlusion or lighting variations which are common in real-world data. Similarly, FPS measurements often ignore energy consumption, a critical factor for wearable HGR systems. }

\textcolor{black}{From the early work on the Gesture Recognition Performance Score (GPRS) \cite{pisharady2014gesture}, which combined 14 different indices to evaluate the performance of an HGR algorithm, to the recent Energy-Delay Product (EDP) that combines energy consumption with latency \cite{ceolini2020hand} there exist various composite metrics for evaluating the performance and their complexity and factors depend on the task.}

\subsection{Performance on Key Datasets}

\textcolor{black}{
To provide a snapshot of the state-of-the-art, Table~\ref{tab:key_dataset_performance} summarizes the best-reported performance metrics on several of the most frequently utilized datasets, according to our research.
The choice of metric, namely Accuracy for datasets primarily used in classification tasks, and Mean Per-Joint Position Error (MPJPE) for those focused on estimation tasks, aligns with common evaluation practices, discussed in Section~\ref{subsec:eval_metrics}. \textcolor{black}{The table consists of four columns: Task groups datasets by their primary application (Classification or Estimation), Dataset identifies the benchmark dataset name, Best Accuracy reports the highest achieved accuracy percentage for classification datasets, Best MPJPE presents the lowest mean per-joint position error for estimation datasets, and Method references the publication of the methodology achieving the respective best performance.}
}

\begin{table}[h!]
\caption{\textcolor{black}{State-of-the-Art Performance on Key Benchmark Datasets.}}
\label{tab:key_dataset_performance}
\centering
\begin{adjustbox}{width=\columnwidth,center}
\begin{tabular}{@{}lcccc@{}}
\toprule
\textcolor{black}{\textbf{Task}} & \textbf{Dataset} & \textbf{Best Accuracy (\%)} & \textbf{Best MPJPE} & \textcolor{black}{\textbf{Method}} \\
\midrule
\multirow{7}{*}{\textcolor{black}{\textbf{Classification}}} & ASL & 100 & --- & \cite{SHARMA2021115657} \\
 & FPHA & 94.43 & --- & \cite{10582035} \\
 & NVIDIADHG & 88.4 & --- & \cite{Kopuklu_2018_CVPR_Workshops} \\
 & isoGD & 68.15 & --- & \cite{chen_li_fang_xin_lu_miao_2022} \\
 & AUTSL & 98.53 & --- & \cite{9523142} \\
 & JESTER & 96.7 & --- & \cite{Jiang_2019_ICCV} \\
 & \textcolor{black}{WLASL} & \textcolor{black}{84.65} & --- & \textcolor{black}{\cite{electronics13071229}} \\
\midrule
\multirow{5}{*}{\textcolor{black}{\textbf{Estimation}}} & HO3D & --- & 1.1 & \cite{10050006} \\
 & SHP & --- & 6.61 & \cite{Cai_2019_ICCV} \\
 & DexYCB & --- & 7.54 & \cite{liu_ren_gao_wang_sun_qi_zhuang_liao_2024} \\
 & Human3.6M & --- & 48.78 & \cite{Cai_2019_ICCV} \\
 & \textcolor{black}{FreiHAND} & --- & \textcolor{black}{1.18} & \textcolor{black}{\cite{10050006}} \\
\bottomrule
\end{tabular}
\end{adjustbox}
\end{table}

\textcolor{black}{
The results presented in Table~\ref{tab:key_dataset_performance} offer several valuable insights into the current capabilities and challenges within HGR. For instance, the leading approaches for ASL \cite{SHARMA2021115657}, AUTSL \cite{9523142}, and JESTER \cite{Jiang_2019_ICCV} report accuracies of 100\%, 98.53\%, and 96.7\%, respectively. This demonstrates a high level of proficiency in recognizing isolated signs or gestures, particularly when datasets provide clear, relatively unoccluded views, or focus on a constrained set of classes. The top accuracies reported for NVIDIADHG \cite{Kopuklu_2018_CVPR_Workshops} (88.4\%) and WLASL \cite{electronics13071229} (84.65\%), while strong, show that recognizing dynamic gestures in more diverse settings remains an open problem. This difficulty is also highlighted in isoGD, where the leading method's accuracy of 68.15\% \cite{chen_li_fang_xin_lu_miao_2022} underscores the ongoing struggle with large and dynamic vocabularies, with high intra-class variation.
}

\textcolor{black}{
\textcolor{black}{However, challenges persist when dealing with more complex and diverse datasets. The top-performing methods on NVIDIADHG \cite{Kopuklu_2018_CVPR_Workshops} and WLASL \cite{electronics13071229} achieve accuracies of 88.4\% and 84.65\%, respectively. While strong, these results indicate that recognizing dynamic gestures in more naturalistic settings remains an open problem. This difficulty is further underscored by the 68.15\% accuracy achieved by the leading method on isoGD \cite{chen_li_fang_xin_lu_miao_2022}, highlighting the persistent challenges posed by larger vocabularies and greater intra-class variation.}
}

\textcolor{black}{
\textcolor{black}{For estimation tasks, where lower MPJPE signifies better method performance, we observe a significant trend towards methods that generalize well across different capture conditions. An example is the methodology from \cite{10050006}, which achieves the best-reported performance on two distinct and widely-used benchmarks: HO3D (1.1 MPJPE) and FreiHAND (1.18 MPJPE). Since HO3D focuses on hand-object interaction and FreiHAND is known for its diversity in hand poses, the fact that a single method excels on both suggests its underlying architecture is robust to variations in scene context and hand appearance, demonstrating a strong capability for learning generalizable features.}
}

\textcolor{black}{
This pattern of cross-dataset generalization is further reinforced by the work of \cite{Cai_2019_ICCV}, which achieves the best reported MPJPE on both SHP (6.61) and Human3.6M (48.78). While the Human3.6M dataset primarily focuses on full-body pose, the hand component is notoriously challenging due to its smaller relative size and frequent occlusions. The strong performance of \cite{Cai_2019_ICCV} across these two distinct estimation datasets —one focused on stereo hand pose (SHP) and the other on broader human pose (Human3.6M)— suggests that the underlying methodology possesses a degree of generalizability and robustness applicable to diverse 3D pose estimation scenarios. This is a commendable achievement, as methods often specialize heavily on specific dataset characteristics. The relatively higher MPJPE for Human3.6M compared to dedicated hand datasets is expected, given the full-body context and scale differences, but the comparative success of \cite{Cai_2019_ICCV} is significant.
}

\textcolor{black}{
It is crucial, however, to interpret these figures with caution. As discussed previously, direct numerical comparison across studies can be confounded by variations in experimental setups, data splits, pre-processing steps, and even minor differences in metric implementation if not strictly standardized. Nevertheless, these benchmark performances serve as important indicators of progress and highlight areas where particular methods excel or where significant challenges remain.
}

\textcolor{black}{
\subsection{Key Takeaways of this section}
Section VI presented an overview of the benchmark datasets and evaluation metrics commonly used in HGR research. The essential points regarding datasets and their evaluation are:
\begin{itemize}
    \item \textbf{Dataset Landscape:} A variety of datasets exist, categorized by input type (RGB, RGB-D, video), task (classification, estimation), sample size, and gesture type. Foundational datasets, even older ones, are still used as benchmarks.
    \item \textbf{Key Dataset Characteristics:}
        \begin{itemize}
            \item \textit{Classification:} Dynamic gesture datasets are often large and complex. RGB-D and skeleton features are commonly used.
            \item \textit{Estimation/Reconstruction:} Require extensive annotated data; synthetic datasets often complement real-world data.
        \end{itemize}
        \item \textbf{Prominent Datasets and Performance Insights:} Datasets like ASL (American Sign Language) for classification and HO3D for hand gesture estimation are frequently referenced, with high performance reported (e.g., 100\% accuracy on ASL, 1.1 MPJPE on HO3D in select studies).
        \item \textbf{Evaluation Metrics:}
        \begin{itemize}
            \item \textit{Classification:} Accuracy (static), Levenshtein distance (dynamic).
            \item \textit{Real-time:} Frame Processing Rate (FPS).
            \item \textit{Estimation:} Mean Per-Joint Position Error (MPJPE) for 3D pose, Intersection-over-Union (IoU) for segmentation.
        \end{itemize}
    \item \textbf{Limitations of Current Metrics:} Often fail to holistically capture real-world usability, robustness to common variations (occlusion, lighting), or energy efficiency.
\end{itemize}
}

\section{Current research challenges and future trends} \label{sec:challenges}

\subsection{Challenges in Hand Gesture Classification}
Hand gesture classification faces several challenges when applied in real-world scenarios. The variability in data, including different camera views, diverse execution styles of the same action (e.g., using one or both hands), and varying action durations raise significant challenges that make temporal analysis and gesture segmentation particularly demanding. Other challenges comprise the occlusions of the hand by surrounding objects, which further complicates detection and tracking, low resolution, and the background and illumination variations that exacerbate the challenge of robust classification. Accurate annotation of occluded or ambiguous keypoints across multi-camera setups demands advanced techniques, such as triangulation and re-projection, adding layers of complexity to training data preparation.

Another critical aspect is the sensitivity to motion scales. Gesture classification can depend on short-term motion details for actions like "clapping" or long-term trends for movements such as "waving up." Effective handling of this variation requires sophisticated models that balance fast and slow temporal streams. Lastly, achieving real-time recognition with low latency while maintaining high accuracy is a persistent challenge, especially for early-stage gesture detection. The high data volumes that are inherent to temporal video streams increase computational demands, and the fusion of multimodal input which is becoming a necessity often requires separate networks for each modality, leading to significant model complexity. In terms of multimodal fusion, another challenge is the correct balance between modalities, which do not perform equally well in recognizing actions.

\subsection{Future Trends in Hand Gesture Classification}
\textcolor{black}{
Emerging approaches in static and dynamic hand gesture classification leverage advanced deep learning techniques and innovative fusion strategies to address domain-specific challenges. \textbf{Hybrid models} combining Convolutional Neural Networks (CNNs) for spatial feature extraction with Recurrent Neural Networks (RNNs) or Long Short-Term Memory (LSTM) networks for temporal modeling excel at capturing short-term spatiotemporal dependencies. However, they face challenges in real-time deployment due to computational overhead and struggle with variable-length gesture sequences in continuous recognition scenarios. To mitigate this, \textbf{3D CNNs} paired with parallel processing architectures (e.g., multi-branch designs for temporal resolutions) improve robustness to occlusions and lighting variations but require extensive computational resources, limiting their use in edge devices.
}

\textcolor{black}{
\textbf{Multimodal fusion strategies} integrating RGB, depth, and optical flow data enhance performance under complex backgrounds by combining complementary information at data, feature, or decision levels. However, these methods introduce synchronization challenges between modalities and increased model complexity, risking overfitting on small datasets. For skeleton-based recognition, \textbf{Spatial-Temporal Graph Convolutional Networks} (ST-GCNs) effectively model joint kinematics but are highly sensitive to noisy skeleton detection, particularly in low-resolution or occluded scenarios.
}

\textcolor{black}{
Transformers, such as \textbf{Vision Transformers} (ViT) \cite{BAMANI2024108443}, address limitations of CNNs in capturing global dependencies, achieving superior performance on static gestures with high intra-class variability (e.g., sign language). However, their reliance on large-scale datasets and computational intensity restricts their applicability to resource-constrained environments. For dynamic gestures, \textbf{Connectionist Temporal Classification (CTC) loss} \cite{graves2014towards} enables alignment-free training for sequence prediction but struggles with gestures involving abrupt motion transitions or overlapping actions.
}

\textcolor{black}{
Recent efforts to streamline complexity include transfer learning for domain adaptation (e.g., cross-dataset generalization) and feature reduction techniques (e.g., PCA) to handle high-dimensional multimodal data. While these reduce computational costs, they risk losing discriminative features critical for fine-grained gestures. Notably, studies like \cite{9525136} demonstrate that random frame sampling can match frame-wise feature extraction in accuracy for video-based HGR, offering a trade-off between efficiency and performance.
}

\subsection{Challenges in Sign Language Recognition}
Sign language recognition faces numerous challenges, primarily arising from the complexity of sign languages and the technical restrictions that affect all computer vision tasks. Vision-based approaches must contend with changing environmental conditions, cluttered backgrounds, varying illumination, and diverse image resolutions. Any occlusions of the hands further complicate sign identification, making it difficult to extract clear and consistent features. These challenges underscore the need for robust preprocessing and feature extraction techniques, which can help mitigate issues like noise and inconsistencies in input data.

Another major hurdle lies in data diversity and availability. Sign languages vary across regions, with unique vocabularies and grammatical structures, necessitating large-scale, annotated datasets that cover a wide range of signs and dialects. Many existing datasets are either too small or lack diversity, which limits generalization across different sign languages. Continuous sign language recognition (cSLR) is a computationally intensive task that requires simultaneous processing of spatial and temporal features. The integration of 3D-CNNs or Transformer models is a promising approach, which however demands significant computational resources, thus challenging the implementation of lightweight or real-time systems. Addressing these challenges requires advancements in model efficiency, and increased focus on data augmentation and synthesis techniques.

\subsection{Future Trends in Sign Language Recognition}

Recent trends in Sign Language Recognition comprise the use of advanced AI and deep learning methodologies to overcome existing challenges. Convolutional Neural Networks (CNNs) and Recurrent Neural Networks (more specifically LSTM networks) have been instrumental in achieving high accuracy for static and dynamic gesture recognition. Studies such as those utilizing faster R-CNNs and 3D-CNNs have showcased near-perfect recognition rates on structured datasets, with further improvements anticipated through the expansion of training data. Additionally, end-to-end learning frameworks, including the Transformer model, are gaining attraction for their ability to integrate spatial and temporal features, offering state-of-the-art results in continuous sign language recognition.

Emerging approaches also emphasize integrating pose estimation and wearable technologies to enhance sign identification. Systems such as MyoSign, which leverage wearables for precise hand and body motion tracking, represent a shift towards multimodal recognition systems that combine video and depth images with sensor data for improved robustness, signaling the trend for multimodal data fusion. Lightweight models like YOLOv3 are also being explored for real-time applications in challenging environments. Furthermore, researchers are exploring generative adversarial networks (GANs) and neural machine translation for automatic sign language production, bridging the gap between spoken and sign languages. These trends reflect a growing emphasis on scalability, multimodal integration, and creation of user-friendly applications that support seamless communication for individuals with hearing or speech impairments.

\subsection{Challenges in Hand Gesture Estimation}
Hand gesture estimation faces several challenges, which are connected to the complexity of human hand movements and the visual properties of hand images. One of the most prominent challenges is the high degree of freedom (DOF) of the human hand, which allows for complex and highly variable gestures involving numerous degrees of motion for each finger and joint. This flexibility leads to ambiguity in hand appearance, especially when fingers overlap or are occluded by objects. Self-occlusion and external object occlusion further complicate the task, as the system needs to accurately identify joint positions and hand gestures despite missing parts of the hand in the captured image. The variations in hand shapes, sizes, and textures introduce additional layers of difficulty, apart from the varying background and the changes in illumination, which all hinder accurate hand pose estimation. Traditional computer vision techniques often struggle to effectively represent and process the intricate details of hand poses, especially in dynamic scenarios where gestures are continuous and evolve over time.

The presence of noise in data collected from low-quality cameras can also degrade the performance of gesture recognition systems. Despite advancements in deep learning, the performance of hand pose estimation models still suffers from issues like overfitting when training on small or unbalanced datasets, especially when dealing with extreme or rare hand shapes. The need to generalize across a wide variety of hand postures while accounting for these imperfections remains a critical problem. Furthermore, the temporal dynamics of hand gestures present additional challenges, as many recognition systems primarily focus on spatial features, overlooking the importance of temporal changes. Gesture recognition systems need to capture sequential relationships and dynamic features effectively, requiring the integration of time-series data processing capabilities, such as temporal convolutions. Handling the temporal dimension, while also addressing spatial complexities, is a significant hurdle in creating robust, real-time hand gesture recognition systems.

\subsection{Future Trends in Hand Gesture Estimation}
One of the key trends in the field of Hand Gesture Estimation is the use of convolutional neural networks (CNNs) and other deep architectures to automatically learn features from large datasets of hand images. These techniques allow to further train more accurate and robust models that can handle complex hand poses. By leveraging large, diverse datasets, CNNs can adapt to variations in hand shapes, poses, and environmental conditions, improving the overall accuracy of gesture recognition and hand pose estimation. Additionally, the integration of advanced neural network architectures, such as recurrent neural networks (RNNs) and 3D convolutional neural networks (3D CNNs), has shown promise in effectively handling dynamic gestures. These models capture both spatial and temporal features, making them highly suitable for time-series gesture recognition tasks. For instance, the development of dynamic gesture recognition systems that employ optimized Convolutional 3D architectures allows for more efficient processing of both local and global hand features, leading to improved recognition performance in real-time applications.

Another trend that has gained attraction is the joint estimation of both hand pose and gesture recognition in a unified framework. By combining the two tasks, models can more efficiently utilize shared features, improving the accuracy and robustness of both tasks. Multi-task learning approaches have become popular in this domain, where a single model is trained to perform multiple related tasks simultaneously. This trend is evident in the development of hybrid models that not only estimate hand poses, but also predict hand gestures using end-to-end deep learning pipelines. Such models often incorporate advanced techniques like feature extraction modules, where CNNs extract key features from input images and utilize them for both pose estimation and gesture recognition. Additionally, the use of depth cameras, RGB-D sensors, and other multi-modal inputs is gaining popularity, as these sensors provide richer information for pose estimation compared to traditional RGB images alone. These advancements allow for more accurate and real-time hand gesture recognition, even in challenging conditions like partial occlusions or cluttered environments.

\subsection{Challenges in Hand Reconstruction}
A significant challenge in 3D hand reconstruction is the variability in hand configurations and interactions with different environments, especially when the task relies solely on monocular data. Unlike multi-view images or depth maps, monocular images often provide less comprehensive information, making it difficult to capture the full 3D structure of the hand accurately. The presence of various hand poses, deformations, and interactions with objects only compounds this issue, as the models must generalize well across diverse hand-object interactions while maintaining real-time performance. As monocular input is more common in practical applications such as augmented reality and human-computer interaction, addressing these challenges is crucial to enabling seamless 3D hand tracking in real-world scenarios.

Another challenge is the accurate and consistent reconstruction of 3D hand shapes across video sequences. Hand movements are inherently dynamic, and to maintain accuracy and realism, the reconstruction model must effectively account for motion, deformation, and transitions between different poses. The complexity increases further when considering that many existing datasets suffer from limitations such as insufficient variability in hand poses, camera angles, and object interactions. These issues make it difficult for models to learn robust representations that can generalize to real-world scenarios, where hand poses can be unpredictable and highly diverse. Moreover, the reconstruction process often relies on noisy or incomplete input data, which further complicates the task of producing accurate 3D reconstructions that adhere to biomechanical constraints.

\subsection{Future Trends in Hand Reconstruction}

Recent trends in hand/body reconstruction have shifted towards self-supervised learning methods, which aim to eliminate the need for manual annotations of training data. These methods typically rely on large collections of unlabeled images or video sequences, combined with a 2D keypoint estimator, to learn 3D hand reconstructions. Self-supervised models, such as S2HAND and S2HAND(V), have shown promise in achieving accurate and consistent reconstructions by learning from video sequences with temporal consistency constraints. This approach significantly reduces the need for costly and labor-intensive data annotation, making it more scalable and applicable to real-world applications. Additionally, the integration of biomechanical models, which impose physical constraints on the reconstructed hands, has enhanced the realism and plausibility of the output, ensuring that reconstructed hand poses adhere to human anatomical limits.

The use of novel representations such as the grasping field (GF) has also gained attraction in the field. GF captures the critical interactions between hand and object by modeling the joint distribution of hand and object shape in a unified framework. This approach improves the accuracy of hand-object interaction modeling by enabling the synthesis of realistic human grasps. Furthermore, advancements in texture and shape consistency regularization are improving the realism of hand reconstructions, encouraging models to produce more coherent textures and shapes across frames. The combination of self-supervision, biomechanical constraints, and advanced modeling techniques such as the GF has led to significant improvements in both the accuracy and practicality of 3D hand and body reconstruction, pushing the boundaries of what is possible in applications like virtual reality, robotics, and human-computer interaction.

\subsection{Challenges on the practical deployment of HGR solutions}
\textcolor{black}{
One of the primary hurdles in deploying Hand Gesture Recognition (HGR) systems in real-world scenarios is achieving robust generalization. HGR models often face challenges when encountering variations in hand shapes, sizes, and orientations across different users. Environmental factors such as inconsistent lighting conditions and cluttered backgrounds, particularly for vision-based systems, can significantly degrade performance. Furthermore, occlusions, where the hand is partially or fully hidden by other objects or even parts of the user's body, present a considerable obstacle to accurate recognition. Ensuring that an HGR system can maintain a high level of accuracy and reliability across diverse and uncontrolled real-world settings remains a key challenge for practical deployment.
}

\textcolor{black}{
Another critical aspect of practical HGR deployment is managing latency to ensure real-time performance. Many applications, such as virtual reality interaction or controlling robotic devices, demand immediate feedback, requiring minimal delay between the user's gesture and the system's response \cite{benitez2021improving}. The computational complexity of sophisticated recognition algorithms, especially those based on deep learning, can introduce significant latency, making them less suitable for time-sensitive applications. Striking a balance between achieving high recognition accuracy and maintaining low latency is a persistent challenge. Moreover, the need to integrate HGR solutions into edge devices with limited processing power further complicates the task of achieving real-time responsiveness \cite{fertl2025hand}. 
}

\textcolor{black}{
Finally, the practical deployment of HGR systems is often constrained by computational resources and the need for scalability. Many target platforms, such as mobile phones, embedded systems, or wearable devices, have limited processing power, memory, and battery life. This necessitates the development of efficient and lightweight HGR models that can operate effectively under these constraints. Furthermore, as the number of recognizable gestures increases or the system needs to support a larger user base, maintaining performance and efficiency becomes a significant challenge. Power consumption is also a crucial consideration, especially for battery-powered devices, as high computational demands can lead to rapid battery drain, impacting the usability of the HGR solution.
}

\subsection{Edge Computing and Resource-Constrained Deployment} \label{subsec:edge_ai}

\textcolor{black}{The increasing demand for interactive and immersive experiences has spurred significant interest in deploying HGR systems directly on edge devices, such as smartphones, AR/VR headsets, and smart wearables. Edge deployment offers several advantages, including reduced latency by processing data locally, enhanced privacy as sensitive visual data may not need to leave the device, and improved reliability with offline functionality. However, these resource-constrained platforms present substantial challenges due to their limited computational power, memory footprint, and stringent battery life requirements, which are often at odds with the complexity of state-of-the-art deep learning models for HGR.
}

\textcolor{black}{Addressing these constraints necessitates a multi-faceted approach. Research efforts actively explore \textbf{lightweight model architectures} specifically designed for efficiency (e.g., MobileNets \cite{howard2017mobilenetsefficientconvolutionalneural}, SqueezeNets \cite{iandola2016squeezenetalexnetlevelaccuracy50x}, or custom-designed compact networks). Additionally, \textbf{model compression techniques} are crucial. These include \textit{pruning}, which removes less important weights or connections to reduce model size and computation; \textit{quantization}, which converts model parameters from high-precision floating-point numbers to lower-precision representations (e.g., 8-bit integers), significantly reducing model size and often speeding up inference, especially on compatible hardware \cite{jacob2017quantizationtrainingneuralnetworks}; and \textit{knowledge distillation}, where a smaller "student" model is trained to mimic the behavior of a larger, more accurate "teacher" model \cite{hinton2015distillingknowledgeneuralnetwork}. Concurrently, advancements in \textbf{hardware acceleration}, with dedicated NPUs (Neural Processing Units) or DSPs (Digital Signal Processors) in modern SoCs (System on Chips), are vital for achieving real-time performance. The synergy between efficient algorithmic design and hardware capabilities, often termed \textbf{algorithm-hardware co-design}, is paramount for the successful deployment of HGR on the edge, enabling sophisticated interactions in everyday devices, while carefully balancing the trade-offs between accuracy, latency, and power consumption.
}

\subsection{Explainability and Interpretability in HGR (XAI)}
\textcolor{black}{As HGR systems, particularly those based on deep learning, become more complex and achieve higher performance, their "black-box" nature poses significant challenges. Understanding \textit{why} a model makes a particular gesture-based decision is crucial for debugging, building user trust, providing meaningful feedback, ensuring fairness (e.g., identifying biases against certain user groups or hand shapes), and deploying HGR in safety-critical applications (e.g., human-robot interaction, medical assistance). Explainable AI (XAI) \cite{rodis2024multimodalexplainableartificialintelligence} aims to provide insights into the decision-making process of these models.}

\textcolor{black}{
Several XAI techniques can be adapted for HGR:
\begin{itemize}
    \item \textbf{Gradient-based Methods (e.g., Grad-CAM, Grad-CAM++):} These methods use the gradients flowing into the final convolutional layer of a CNN to produce a coarse localization map, highlighting important regions in the input image that contribute to a specific class prediction \cite{Selvaraju_2019}. In HGR, Grad-CAM can visualize which parts of the hand (e.g., specific fingers, palm orientation) or surrounding context are salient for classifying a static gesture from an RGB or depth image. Extensions exist for video data, allowing visualization of spatio-temporal saliency in dynamic gestures.
    \item \textbf{Perturbation-based Methods (e.g., LIME):} LIME (Local Interpretable Model-agnostic Explanations) \cite{ribeiro2016whyitrustyou} explains individual predictions by learning a simpler, interpretable model locally around the instance being explained. In HGR, this could mean identifying superpixels in an image or key joints in a skeleton whose presence or absence significantly alters the gesture prediction.
    \item \textbf{SHAP (SHapley Additive exPlanations):} Based on Shapley values from cooperative game theory, SHAP assigns an importance value to each input feature for a particular prediction, indicating its contribution to pushing the prediction away from a baseline \cite{lundberg2017unifiedapproachinterpretingmodel}. In HGR, SHAP can be used to understand the importance of individual skeletal joints' coordinates, orientations, or even higher-level engineered features for a gesture classification or estimation task.
    \item \textbf{Attention Mechanisms:} For transformer-based HGR models, the self-attention or cross-attention weights themselves can offer a degree of interpretability, by showing which parts of the input sequence (e.g., frames in a video, or different modalities) the model focuses on when making a decision \cite{vaswani2023attentionneed}. Visualizing these attention maps can help identify key frames or critical interactions. However, careful insights must be extracted, since raw attention weights do not always directly correlate with feature importance for the final prediction, especially in deep multi-head attention architectures.
\end{itemize}
}


\textcolor{black}{Future research in XAI for HGR will likely focus on developing HGR-specific interpretability techniques, creating methods that can explain multimodal and spatio-temporal reasoning more effectively, exploring interactive explanations where users can query the model, and generating human-understandable textual or symbolic explanations for gesture-based decisions. As HGR systems become more integrated into daily life, the demand for transparent and trustworthy AI will only grow, making XAI an important component of future HGR development.}

\subsection{Bias, Inclusivity, and Privacy Considerations}
\textcolor{black}{Beyond technical performance, the responsible development and deployment of HGR systems necessitate careful consideration of ethical challenges, including dataset bias, inclusivity, and user privacy. Failure to address these can lead to systems that are unfair, inaccessible, or violate user trust.}

\subsubsection{Dataset Bias and Inclusivity}

\textcolor{black}{Deep learning models are known to inherit and potentially amplify biases present in their training data. HGR datasets are often susceptible to various forms of bias, leading to significant performance disparities across different user groups:
\begin{itemize}
    \item \textbf{Demographic Bias:} Many publicly available HGR datasets predominantly feature subjects from specific demographic groups (e.g., certain ethnicities like Caucasian, age groups like young adults, or predominantly male participants). Models trained on such data may perform poorly for under-represented ethnicities, older adults, or other groups \cite{10.1145/3457607}.
    \item \textbf{Handedness Bias:} The majority of the population is right-handed, and datasets often reflect this, potentially containing insufficient examples of left-handed gestures. This can lead to reduced accuracy for left-handed users, especially if the model implicitly learns features specific to right-hand dominance or orientation.
    \item \textbf{Cultural Bias:} Gestures, including sign languages, are culturally specific. A dataset collected in one region might not capture the nuances or variations of gestures used elsewhere. Relying solely on such datasets limits the global applicability and fairness of HGR systems, particularly for sign language recognition where distinct national or regional sign languages exist \cite{papad_hgr}.
    \item \textbf{Disability and Physical Variation Bias:} Datasets may lack representation of users with physical disabilities, hand tremors, limb differences, or variations in hand structure, potentially rendering HGR systems inaccessible or inaccurate for these individuals.
\end{itemize}
Ensuring inclusivity requires proactive efforts in dataset curation, including targeted collection from diverse populations across ethnicity, age, gender, handedness, cultural backgrounds, and physical abilities. Furthermore, algorithmic fairness techniques (e.g., re-weighting loss functions, adversarial debiasing, fairness constraints) and rigorous auditing of model performance across different subgroups are essential steps towards building equitable HGR systems.}

\subsubsection{Privacy Implications of Video-Based HGR}
\textcolor{black}{
Visual input, especially video, inherently carries rich information that extends beyond the intended hand gesture, raising significant privacy concerns:
\begin{itemize}
    \item \textbf{Identification and Environmental Exposure:} Cameras inevitably capture the user's surroundings, potentially revealing personal information about their location, home environment, or activities. Faces might inadvertently be captured, leading to user identification.
    \item \textbf{Inference of Sensitive Information:} Hand movements can potentially reveal sensitive attributes. For instance, tremors might indicate health conditions, agitated gestures could suggest emotional states, and specific interaction patterns might reveal habits or preferences.
    \item \textbf{Biometric Data Concerns:} Hand shape, size, and movement dynamics can potentially serve as biometric identifiers. Continuous capture and analysis of this data could enable tracking and profiling of individuals.
    \item \textbf{Surveillance Potential:} The ubiquitous deployment of cameras for HGR (e.g., in smart homes, vehicles, or public kiosks) creates potential for unintended surveillance or data misuse if not properly managed.
\end{itemize}
Mitigating these privacy risks requires a combination of technical and policy-based solutions. \textbf{On-device processing (Edge AI)} is a crucial strategy, minimizing the need to transmit sensitive visual data to the cloud (as discussed in Section \ref{subsec:edge_ai}). \textbf{Privacy-preserving techniques} such as background blurring/subtraction, hand region segmentation (discarding the rest of the frame), federated learning (training models on decentralized data without sharing raw inputs), and differential privacy can further reduce risks. Data minimization principles—collecting only the data strictly necessary for the task—should be adopted. Crucially, \textbf{transparency} about data collection practices and obtaining \textbf{explicit user consent} are fundamental ethical requirements. Adherence to data protection regulations like GDPR is mandatory, guiding responsible data handling, storage, and usage policies. Addressing privacy concerns proactively is vital for fostering user trust and enabling the widespread acceptance of video-based HGR technology.
}

\subsection{Key Takeaways of this section}
\textcolor{black}{
This section highlighted the ongoing challenges in the field of Hand Gesture Recognition and discussed emerging future trends aimed at addressing these issues. Key challenges and directions include:
\begin{itemize}
    \item \textbf{Hand Gesture Classification Challenges:} Include data variability (views, execution styles, duration), occlusions, low resolution, background clutter, and achieving low-latency real-time recognition with high accuracy, especially for multimodal inputs.
    \item \textbf{Sign Language Recognition (SLR) Challenges:} Complexity of sign languages, environmental variations, hand occlusions, data diversity for different sign languages/dialects, and computational demands of Continuous SLR (cSLR).
    \item \textbf{Hand Gesture Estimation Challenges:} High degrees of freedom (DOF) of the hand, self-occlusion and object occlusion, variations in hand appearance, and capturing temporal dynamics.
    \item \textbf{3D Hand Reconstruction Challenges:} Variability with monocular data, ensuring temporal consistency in video, and limitations of existing datasets in pose/interaction diversity.
    \item \textbf{Practical Deployment Challenges:} Achieving robust generalization across users and environments, managing latency for real-time interaction, and operating within resource constraints (computational power, memory, battery life) of edge devices.
    \item \textbf{Future Trends:}
        \begin{itemize}
            \item \textit{Models:} Hybrid models (CNN-RNN), 3D CNNs, Transformers, self-supervised learning.
            \item \textit{Techniques:} Multimodal fusion, pose estimation integration, lightweight architectures, model compression for edge deployment (pruning, quantization).
            \item \textit{Ethical AI:} Development of Explainable AI (XAI) techniques (e.g., Grad-CAM, LIME, SHAP) and addressing dataset bias and user privacy.
        \end{itemize}
\end{itemize}
}

\section{Conclusions} \label{sec:conclusions}

In this survey, we provided a comprehensive review of recent advancements in hand gesture recognition (HGR) from visual input, highlighting the methodologies, datasets, and applications that define the current state of the field. By organizing the literature based on input data types, recognition methods, and task-specific challenges, we established a clear framework for understanding the diverse approaches employed in HGR research. Our analysis revealed significant progress in areas such as robust classification techniques, advancements in 3D hand pose estimation, and the development of benchmark datasets tailored to specific application domains like virtual reality, sign language recognition, and robotics. Additionally, we emphasized the importance of integrating multimodal data and leveraging deep learning architectures to address increasingly complex real-world scenarios.

Despite these advancements, several challenges remain, including improving the robustness of HGR systems in uncontrolled environments, handling occlusions, ensuring scalability and generalization across diverse users, and achieving computational efficiency for real-time applications. 
\textcolor{black}{
A critical limitation of current HGR research—and by extension, this survey—is the lack of standardized benchmarks for fair numerical comparisons across methods. As highlighted in prior sections, heterogeneity in evaluation protocols (e.g., dataset splits, preprocessing, and metric reporting) and inconsistent code availability make direct performance comparisons impractical without large-scale reproducibility efforts.
}
Addressing these issues will require interdisciplinary collaboration and continued innovation in algorithm design, dataset creation, and hardware optimization. 
\textcolor{black}{
To this end, we advocate for community-driven initiatives to establish unified evaluation frameworks and open-source benchmarks, which will enable rigorous, apples-to-apples comparisons of accuracy, efficiency, and robustness.
}

This survey aims to serve as a valuable resource for researchers and practitioners by synthesizing recent trends, identifying gaps in the literature, and outlining future directions for developing more accurate, efficient, and accessible hand gesture recognition systems.
\textcolor{black}{
As part of our ongoing work, we plan to address the benchmarking gap by curating a reproducibility study with standardized metrics and datasets, fostering transparency and reproducibility in HGR research.
}





\bibliographystyle{plain}
\bibliography{sample-base.bib}

\begin{IEEEbiography}[{\includegraphics[width=1in,height=1.25in,clip,keepaspectratio]{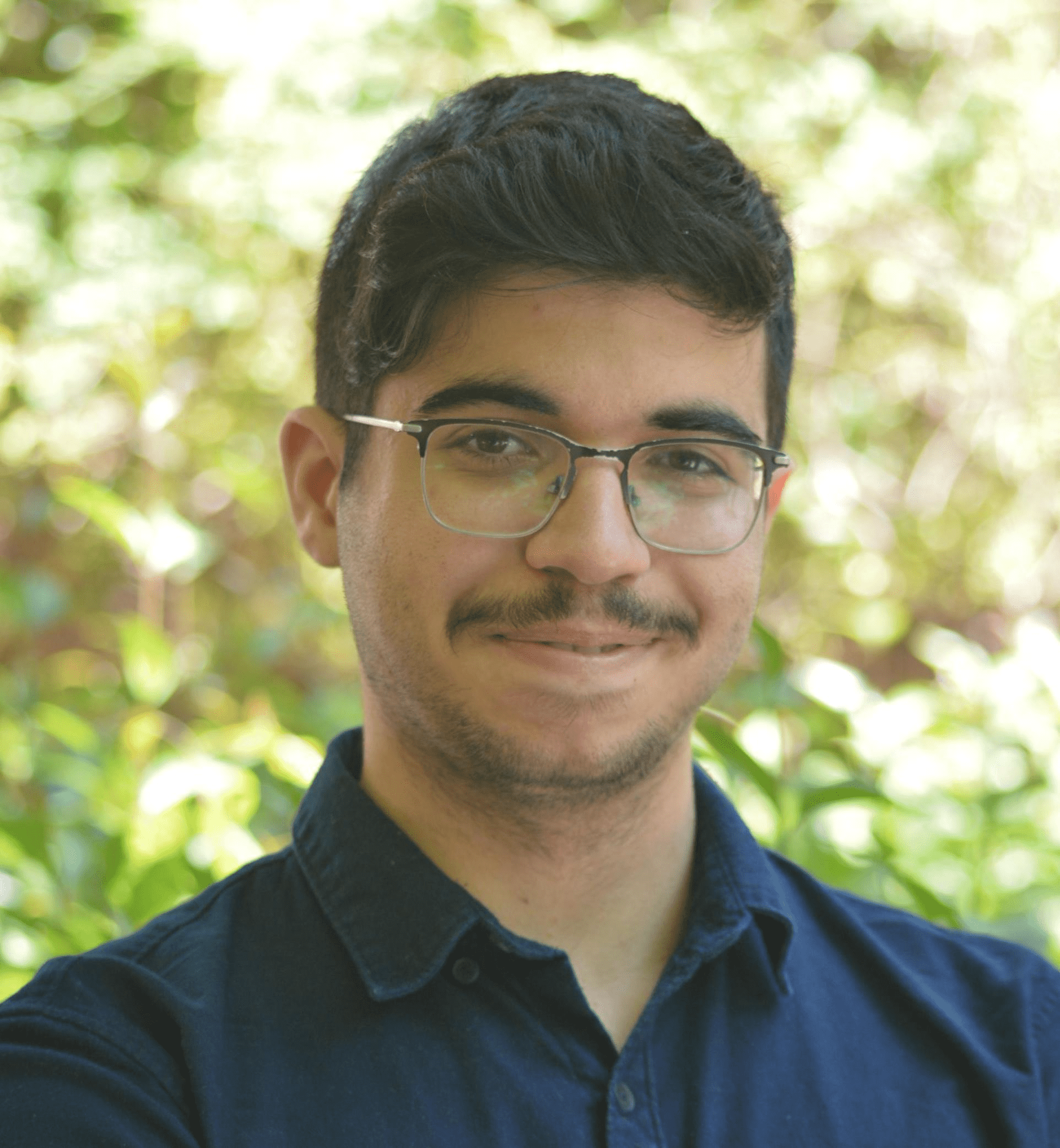}}]{Manousos Linardakis} is a B.Sc. graduate of the Department of Informatics and Telematics, Harokopio University of Athens, where he holds the all-time highest grade in the department's history as of this writing. He is currently pursuing an M.Sc. in Data Science and Machine Learning at the School of Electrical and Computer Engineering, National Technical University of Athens. He has conducted research in multi-robot exploration and computer vision, with practical experience in managing laboratory systems, implementing DevOps practices, and developing decision support software. He also participated in the Google Summer of Code as an open-source developer. His research interests include data science, machine learning and robotics with a focus on integrating theoretical advancements into practical, real-world applications.
\end{IEEEbiography}

\begin{IEEEbiography}[{\includegraphics[width=1in,height=1.25in,clip,keepaspectratio]{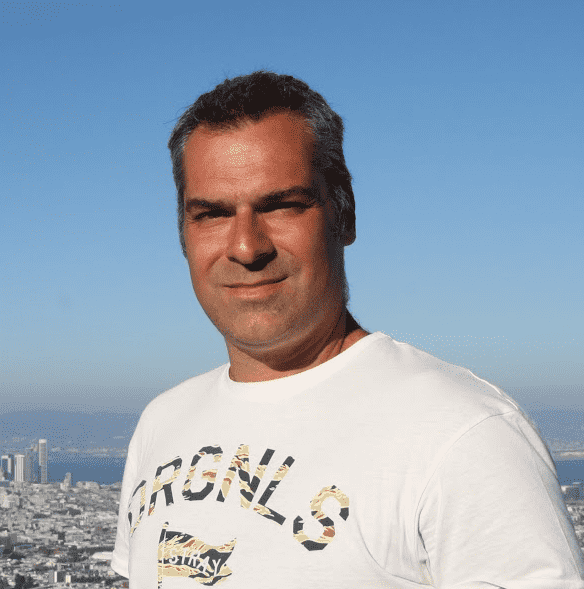}}]{Iraklis Varlamis} (Member, IEEE) received an M.Sc. degree in information systems engineering from the University of Manchester Institute of Science and Technology, Manchester, U.K., and a Ph.D. degree from the Athens University of Economics and Business, Athens, Greece. He is currently a Professor of data management at the Department of Informatics and Telematics, Harokopio University of Athens (HUA), Kallithea, Greece. He has more than 250 articles published in international journals and conferences and more than 5000 citations on his work. He holds a patent from the Greek Patent Office for a system that thematically groups web documents using content and links. His research interests include data mining and social network analytics to recommender systems for social media and real-world applications. He is the Scientific Coordinator for HUA in several EU (H2020, ECSEL, REC) and Qatar (QNFR) projects as well as in national projects.
\end{IEEEbiography}

\begin{IEEEbiography}[{\includegraphics[width=1in,height=1.25in,clip,keepaspectratio]{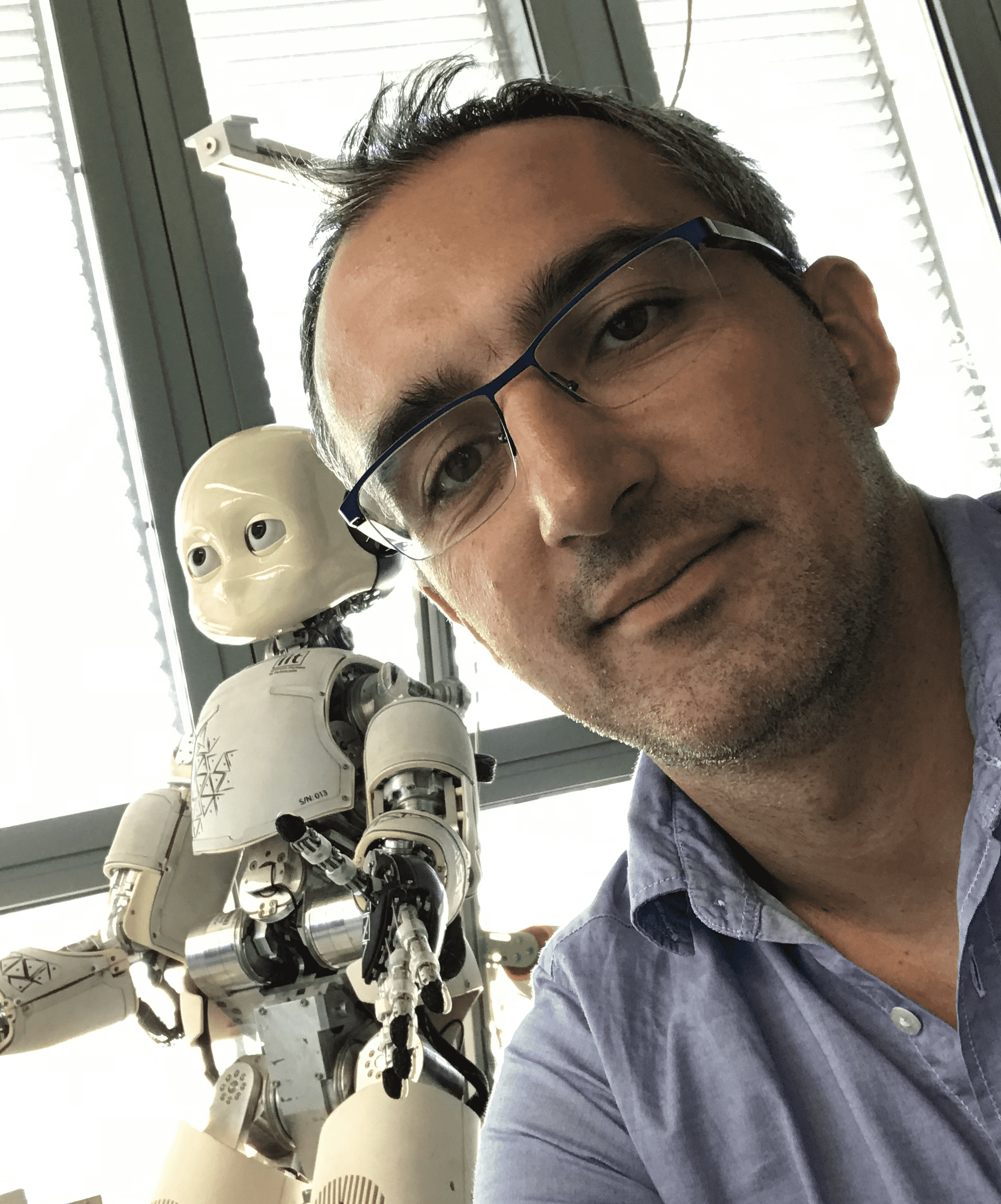}}]{Georgios Th. Papadopoulos} (M) is an Assistant Professor in the area of Computer Graphics and Computational Vision at the Department of Informatics and Telematics of the Harokopio University of Athens in Greece. He received the Diploma and Ph.D. degrees in electrical and computer engineering from the Aristotle University of Thessaloniki (AUTH), Thessaloniki, Greece. He has worked as a Post-doctoral Researcher at the Foundation For Research And Technology Hellas / Institute of Computer Science (FORTH/ICS) and the Centre for Research and Technology Hellas / Information Technologies Institute (CERTH/ITI). He has published over 70 peer-reviewed research articles in international journals and conference proceedings. His research interests include computer vision, artificial intelligence, machine/deep learning, human action recognition, human-computer interaction and explainable artificial intelligence. Dr. Papadopoulos is a member of the IEEE and the Technical Chamber of Greece.

\end{IEEEbiography}

\end{document}